%% file: main.tex
\documentclass{article}

\PassOptionsToPackage{sort, numbers}{natbib}

\usepackage[final]{neurips_2022}

\usepackage[utf8]{inputenc} 
\usepackage[T1]{fontenc}    
\usepackage{hyperref}       
\usepackage{url}            
\usepackage{booktabs}       
\usepackage{amsfonts}       
\usepackage{nicefrac}       
\usepackage{microtype}      
\usepackage{xcolor}         
\usepackage{enumitem}

\usepackage{graphicx}
\usepackage{subfigure}
\usepackage{bm}
\usepackage{bbm}
\usepackage{algorithm}
\usepackage{algorithmic}
\usepackage{amsmath}
\usepackage{amssymb}
\usepackage{mathtools}
\usepackage{amsthm}
\usepackage{todonotes}
\usepackage[capitalize,noabbrev]{cleveref}

\usepackage{wrapfig}
\usepackage{caption}

\input{equations}

\title{Missing Data Imputation and Acquisition with Deep \\ Hierarchical Models and Hamiltonian Monte Carlo}

\author{%
  Ignacio Peis \\
  Universidad Carlos III de Madrid\\
  Madrid, Spain\\
  \texttt{ipeis@tsc.uc3m.es} \\
  \And
  Chao Ma \\
  Microsoft Research\\
  University of Cambridge\\
  Cambridge, UK\\
  \texttt{cm905@cam.ac.uk} \\
  \And
  José Miguel Hernández-Lobato\\
  University of Cambridge\\
  Cambridge, UK\\
  \texttt{jmh233@cam.ac.uk} \\
}

\begin{document}

\maketitle

\begin{abstract}
  Variational Autoencoders (VAEs) have recently been highly successful at imputing and acquiring heterogeneous missing data. However, within this specific application domain, existing VAE methods are restricted by using only one layer of latent variables and strictly Gaussian posterior approximations. To address these limitations, we present HH-VAEM, a Hierarchical VAE model for mixed-type incomplete data that uses Hamiltonian Monte Carlo with automatic hyper-parameter tuning for improved approximate inference.  Our experiments show that HH-VAEM outperforms existing baselines in the tasks of missing data imputation and supervised learning
  with missing features.
  Finally, we also present a sampling-based approach for efficiently computing the information gain when missing features are to be acquired with HH-VAEM. Our experiments
  show that this sampling-based approach is superior to alternatives based on Gaussian approximations.
\end{abstract}

\input{introduction}

\input{related_work}

\input{model}
\input{sampling_method}

\input{experiments}

\input{conclusion}

\begin{ack}

Ignacio Peis acknowledges support from Spanish government Ministerio de Ciencia, Innovaci\'on y Universidades under grants FPU18/00516, RTI2018-099655-B-I00, EST21/00467 and PID2021-123182OB-I00 and from Comunidad de Madrid under grant Y2018/TCS-4705 PRACTICO-CM. Jos\'e Miguel Hern\'andez-Lobato acknowledges support from Boltzbit.

\end{ack}

\bibliography{references}
\bibliographystyle{abbrv}


\appendix

\input{appendix}

\end{document}

%% file: equations.tex
\newcommand{\xo}{\bm{x}_O} 
\newcommand{\yo}{\bm{y}_O} 
\newcommand{\xu}{\bm{x}_U} 
\newcommand{\zo}{\bm{z}_O} 

\newcommand{\yu}{\bm{y}_U} 
\newcommand{\x}{\bm{x}} 
\newcommand{\y}{\bm{y}} 
\newcommand{\z}{\bm{z}} 
\newcommand{\h}{\bm{h}} 

\newcommand{\beps}{\boldsymbol{\epsilon}} 

\newcommand{\p}{p_{\theta}} 
\newcommand{\KL}{D_{\text{KL}}} 

\newcommand{\loss}{\mathcal{L}} 

%% file: introduction.tex
\section{Introduction}
\label{sec:intro}

Many real-world unsupervised learning tasks require dealing with complicated datasets with mixed types (real, positive-valued, continuous, or discrete) and missing values. For this purpose, variational autoencoders \citep{kingma2013auto, rezende2014stochastic, kingma2019introduction} stand out in the recent literature as robust generative models that efficiently handle high-dimensional data. However, in their naive configuration, every data dimension is assumed to have similar statistical properties (i.e., homogeneity), and all dimensions are considered to be completely observed. Both assumptions won't hold in many real-world scenarios. Recent works have adapted VAEs to handle incomplete \citep{collier2020vaes, ma2018eddi, garnelo2018conditional, mattei2019miwae} and mixed-type data \citep{nazabal2020handling, ma2020vaem, gong2021variational}, and demonstrated improved performance in downstream tasks such as missing data imputation and active information acquisition.
Despite these advances, existing approaches are far from optimal as they are based on restrictive design choices: 1), only one layer of latent variables are considered; 2),
Gaussian posterior approximations are usually adopted. These will lead to limited flexibility and additional bias, especially under real-world settings with complex mixed-type incomplete data.

In the literature, the issue of model flexibility and inference bias are often addressed separately. For example, approximate inference bias can be reduced by using Monte Carlo sampling \citep{salimans2015markov, thin2021monte}. More specifically, Hamiltonian Monte Carlo (HMC) \citep{duane1987hybrid, betancourt2015hamiltonian} stands out among MCMC methods in machine learning
due to its superior efficiency for exploring the target density. In the context of VAE, HMC has also been combined with stochastic variational inference \citep{caterini2018hamiltonian} for improving the training of VAEs. On the other hand, the flexibility of VAEs can be improved by considering hierarchical VAEs with
multiple layers of hidden variables \citep{sonderby2016ladder, maaloe2019biva, vahdat2020nvae, child2020very}. By using a hierarchical structure in the latent space, they enforce the information to flow from high-level representations to more specific observable factors, imitating the way information is often organized in the real world. 

However, the issue of modeling flexibility and approximate inference bias are often heavily intertwined, and addressing them simultaneously in a \emph{joint} manner is highly non-trivial. The hierarchical organization of the latent variables creates complicated posterior dependencies that are not straightforward to deal with and require special consideration. To improve Gaussian approximate inference, most  works opt by defining shared paths between the recognition and generative networks.
While this makes hierarchical VAEs practical, the bias introduced by the Gaussian approximations is still present.
To the best of our knowledge, none of the aforementioned hierarchical VAEs has been previously combined with Monte Carlo algorithms for improving over standard Gaussian approximate inference.

To overcome such limitations, we focus on training new hierarchical VAE models 
for heterogeneous mixed-type data with
HMC. Our models can be used for missing data imputation and for supervised learning with missing data.
We also present a sampling-based  framework that allows our models to perform accurate sequential active information acquisition.

Our main contributions are as follows:
\begin{itemize}
    \item We present HH-VAEM, a deep hierarchical model for handling mixed-type incomplete data that uses HMC with automatic hyper-parameter tuning for outperforming amortized variational inference by generating low bias samples from the true posterior. 
    \item We propose a sampling-based strategy for missing feature acquisition that benefits from the improved inference of HH-VAEM. 
    By using histograms to estimate the mutual information, this strategy achieves lower bias than other Gaussian-based alternatives.
    \item We exhaustively evaluate HH-VAEM in the tasks of 1) missing data imputation, 2) supervised learning with missing data and 3)
    information acquisition
    with our sampling-based strategy. In all cases we report significant gains with respect to baselines.
\end{itemize}

%% file: related_work.tex
\section{Related work}

\subsection{VAEs for mixed-type incomplete data}
Variational Autoencoders \citep{kingma2013auto, rezende2014stochastic} are deep generative models that make use of encoder and decoder networks for mapping data into a latent Gaussian distribution, and reconstructing the latent codes into the original observational space, respectively. The parameters of these networks are trained using amortized Variational Inference \citep{zhang2018advances, cremer2018inference} optimizing a lower bound (ELBO) on the log evidence: $
    \mathbb{E}_{q_{\z} (\z | \x)} \log \frac{p_{\theta}(\x, \z)}{q_{\psi}(\z | \x)} $, 
where the generative model $p_\theta(\x, \z)$ can be expressed in terms of the likelihood $\p(\x|\z)$ and the prior $p(\z)$. In mixed-type data, the vector $\x$ is composed by data from different types: real, positive real, categorical, binary, etc. A naive approach is to consider a factorized decoder using different likelihood contributions $\p(\x|\z) = \prod_d\p(x_d|\z)$ \citep{nazabal2020handling, barrejon2021medical}. Nonetheless, the problem of handling unbalanced likelihoods leads to the domination of some dimensions during the optimization process. In \citep{ma2020vaem}, authors propose a solution using a set of \emph{marginal} VAEs that encode each feature into a Gaussian uni-dimensional space, and a \emph{dependency} VAE that captures the inter-dimensional dependencies more effectively using balanced Gaussian likelihoods.

By marginalizing each dimension of the decoder, incomplete data can be easily handled by dividing the vector $\x$ into the observed $\xo$ and unobserved $\xu$ parts. This methodology is completely valid when using the missing-at-random (MAR) assumption \citep{little2019statistical}, i.e. assuming the  missing mechanism is independent of the missing values. In this our work, we use the same assumption. As proposed in \citep{nazabal2020handling} and \citep{mattei2019miwae}, the ELBO objective is transformed into a lower bound on the observed data, and the unobserved data is replaced with zeros. 

\subsection{Hierarchical VAEs}

Hierarchical models have been successfully employed in deep generative modeling, \citep{bengio2009learning, salakhutdinov2009deep, salakhutdinov2015learning}. In VAEs, defining a hierarchical latent space for VAEs can be straightforward. Nevertheless, potential pitfalls require special attention. Concretely, if the decoder is powerful enough, the model tends to uniquely use the shallowest layers, ignoring the deepest ones and falling into the well-known problem of \emph{posterior collapse} \citep{wang2020posterior, razavi2019preventing, maaloe2019biva}. In the last few years, several works have investigated possible hierarchical structures for VAEs. In \citep{sonderby2016ladder}, a bottom-up deterministic path is used along with a top-down inference path that shares parameters with the top-down structure of the generative model. 
In \citep{maaloe2019biva}, the authors use a bidirectional stochastic inference path.  More recently, \citep{vahdat2020nvae} or \citep{child2020very} have adapted these architectures to complex datasets and high quality images. Possibly motivated by the residual connections in \citep{kingma2016improved}, all these works use a shared path between recognition and generative models that helps in tying the divergences between approximations and priors in the ELBO.

\subsection{Hamiltonian Monte Carlo}

HMC \citep{duane1987hybrid, neal2011mcmc, betancourt2017conceptual} is a particularly effective MCMC algorithm for sampling from a target distribution $p(\z) = \frac{1}{\mathcal{Z}} p^*(\z)$ where $\mathcal{Z}$ is the normalization constant and $\z$ is a $d$-dimensional vector. By augmenting this model to $p(\bm{r},\z)  = \mathcal{N}(\bm{r}; \bm{0},\bm{M}) p(\z)$, and denoting $\bm{r}$ as the \emph{momentum} variable with diagonal covariance matrix $\bm{M}$, with the same dimensionality as $\z$, HMC 
samples are obtained from the distribution by
simulating the time-evolution of a fictitious physical system. 

The algorithm starts by firstly sampling $\z$ and $\bm{r}$ from an initial proposal and the momentum distribution, respectively.  Chains with length $T$ are built by recurrently proposing and accepting new states. To propose a new state, the Hamiltonian dynamics are simulated using a symplectic integrator, Leapfrog being the most common choice. The following updates are repeated for $l=1:LF$ steps:
\begin{equation}\label{eq:leapfrog}
    \begin{aligned}
        & \bm{r}_{l+\frac{1}{2}} = \bm{r}_l + \frac{1}{2} \bm{\phi} \, \odot \nabla_{z_l} \log p^*(\z_l)\,, \\ 
        & \z_{l+1} = \z_l + \bm{r}_{l+\frac{1}{2}} \, \odot \bm{\phi} \, \odot \frac{1}{\bm{M}}\,, \\
        & \bm{r}_{l+1} = \bm{r}_{l+\frac{1}{2}} + \frac{1}{2} \bm{\phi} \, \odot \nabla_{z_{l+1}} \log p^*(\z_{l+1}) ,
    \end{aligned}
\end{equation}
where $\odot$ refers to the Hadamard product, and $\phi$ is the \emph{step size} hyperparameter. Although it is typically defined as a scalar for simplicity, a $d$-dimensional vector can be considered to apply different step sizes per dimension, or further, as considered in this work, a $T\times d$ matrix to apply different steps per each proposal of the chain. The new proposal $(\z', \bm{r}')$ is accepted with probability $min \left[ 1, \exp (-H(\z', \bm{r}') + H(\z, \bm{r}) ) \right]$, where
\begin{equation}
    H(\z, \bm{r}) = -\log p^*(\z) + \frac{1}{2} \bm{r}^T \bm{M}^{-1} \bm{r} .
\end{equation}
For the consecutive $T$ proposals, a new momentum $\bm{r}$ is resampled and the updates of eq. \eqref{eq:leapfrog} are repeated for $LF$ steps to update the state if $(\z', \bm{r}')$ is accepted.  

\subsection{Active Feature Acquisition}
Among all the Active Learning techniques, Active Feature Acquisition \citep{melville2004active, saar2009active, thahir2012efficient, huang2018active} is of special interest in cost-sensitive applications for modeling a trade-off between the improvement of predictions and the cost of acquiring new data at the feature level. Several works in the recent literature have studied methods for performing the task of sequentially acquiring high-value information by selecting features that maximize an information theoretical reward function and enhance the accuracy of the predictions. This task is denoted by SAIA (Sequential Active Information Acquisition). In \citep{ma2018eddi}, an efficient method is proposed  for approximating a non-tractable reward by using the encoder of a VAE that handles missing data. In \citep{ma2020vaem} they extend this method for handling mixed-type data. Both works 
estimate the reward by
relying on Gaussian approximations given by the encoder networks.

%% file: model.tex
\section{Hamiltonian hierarchical VAE for mixed-type incomplete data {(HH-VAEM)}}

\begin{figure}[t]
	\centering
	\subfigure[The HH-VAEM model]{\includegraphics[height=0.4\linewidth]{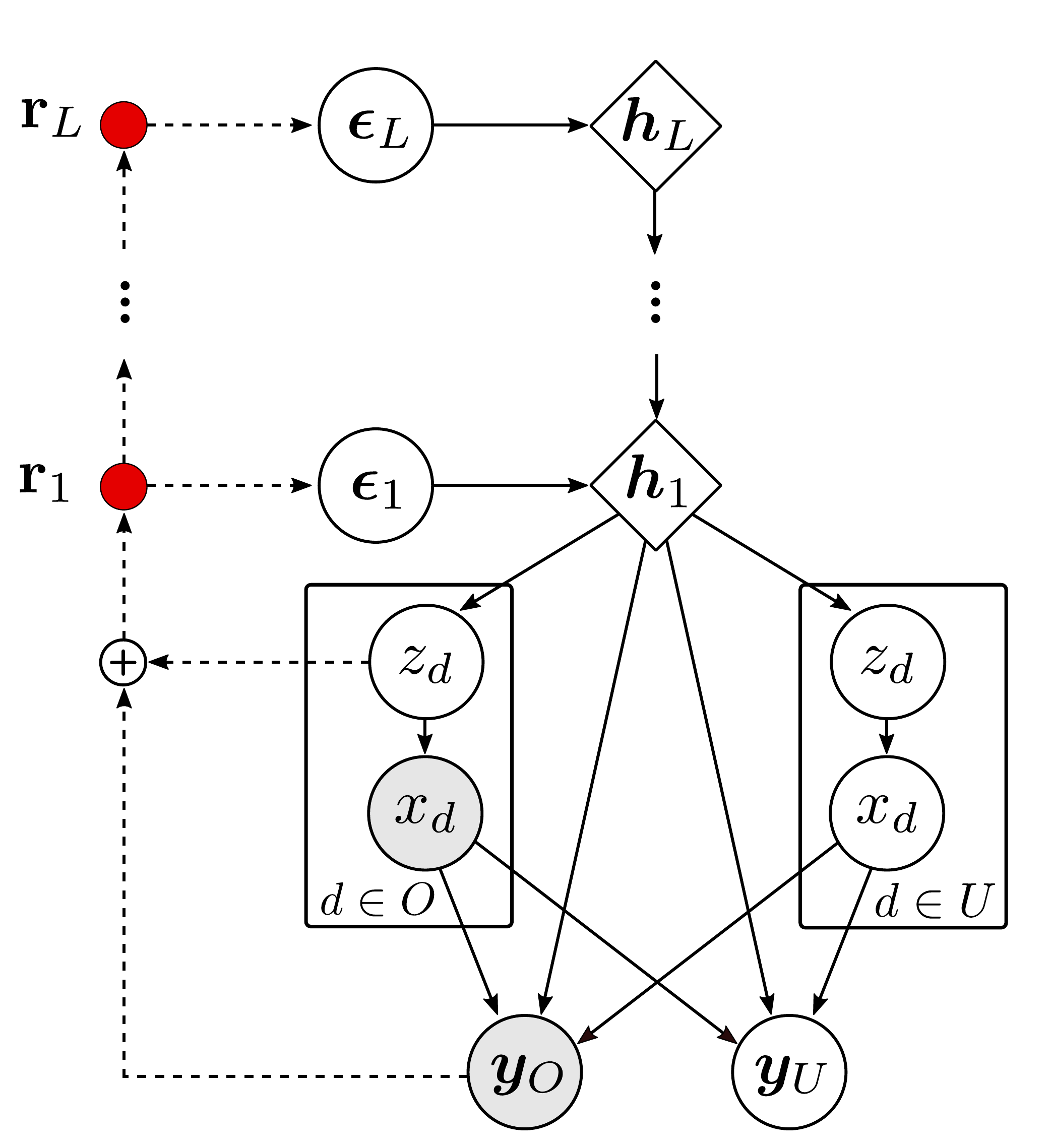}}
	\hspace{2cm}
	\subfigure[HMC sampling]{\includegraphics[height=0.4\linewidth]{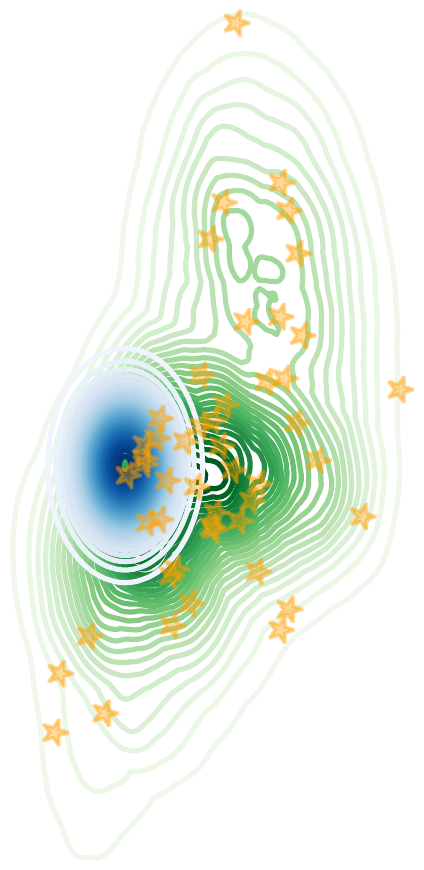}}
	\caption{The HH-VAEM model (a). Illustrative example (b): samples $\beps^{(T)}$ obtained with HMC (orange) following the true posterior $p(\beps|\xo, \yo)$ (green) using the Gaussian distribution given by the encoder $q^{(0)}(\beps|\xo, \yo)$ (blue) as the initial proposal, with latent dimension $M=2$.}
	\label{fig:model_hmc_samples}
\end{figure}

The HH-VAEM model (Figure \ref{fig:model_hmc_samples} (a)) is a Hierarchical VAE  for mixed-type, incomplete data that incorporates HMC with automatic hyper-parameter optimization for sampling from the posterior of the latent variables. In a first stage, the mixed-type data is encoded into marginal Gaussian
posterior distributions as given by univariate VAEs fitted to each data dimension. In a second stage, a hierarchical structure captures the dependencies among the standarized, homogeneous dimensions with well-balanced Gaussian likelihoods. The model is trained using samples from the posterior of the hierarchical latent variables by means of HMC, whilst the HMC hyper-parameters are automatically tuned. A more detailed description is provided in the following subsections.

\subsection{Notation}

The model generates both data $\x\in \mathbb{R}^D$ and output $\y\in \mathbb{R}^P$, where each of these variables is divided into observed parts $\xo$, $\yo$ and unobserved parts $\xu$, $\yu$. Each dimension of $\x$ is denoted by $x_d$.
A training set is composed of $N$ observations as tuples $(\xo^{(n)}, \yo^{(n)})$. For ease of notation, we omit the observation index $n$, and the objectives are presented for a single observation. The \emph{dependency} latent space is composed of $L$ latent variables $\left[ \beps_1, ..., \beps_L \right]$, with $\beps_l \in \mathbb{R}^{m_l}$. The dimension of the joint latent distribution is $\sum_l m_l = M$.
 In the marginal VAEs, the  latent variables $z_d$ are unidimensional. 

\subsection{Handling heterogeneous incomplete data}

Following the strategy proposed by \citep{ma2020vaem}, we perform a two-staged approach for handling heterogeneous data. The marginal distribution of each feature  $p_{\theta_d}(x_d)$ is modeled by a one-dimensional VAE. First, the $D$ marginal VAEs are trained independently by maximizing the marginal ELBO over observed points
\begin{equation}\label{eq:marg_elbo}
    \loss_d(x_d; \{ \theta_d, \gamma_d \} ) = 
    \mathbbm{1}(x_d \in \xo) 
    \mathbb{E}_{q_{\gamma_d} (z_d | x_d)} \log \frac{p_{\theta_d}(x_d, z_d)}{q_{\gamma_d}(z_d | x_d)} ,
\end{equation}
where $\mathbbm{1}(x_d \in \xo)$ is an indicator function that activates the ELBO when the feature is observed. Under the missing-at-random assumption, which is the one considered in this work, Equation \eqref{eq:marg_elbo} leads to a lower bound on observed data likelihoods. Second, a \emph{dependency} VAE encodes a vector $\z$ with the concatenated samples from the marginal posteriors $q_{\gamma_d}(z_d|x_d)$ into the global latent variable $\h$, using zero-filling for the unobserved variables. By using this approach, $\z$ is now homogeneous and can then be easily modeled using a standard Gaussian decoder. By contrast, other works \citep{nazabal2020handling, eduardo2020robust} directly operate with different decoding likelihoods per data type. This approach often leads to having very different magnitudes in the ELBO and may reduce learning efficiency. The ELBO for the second stage dependency VAE is
\begin{equation}\label{eq:dep_elbo}
\begin{gathered}
    \loss(\xo, \yo; \{ \theta, \psi \}) = 
    \mathbb{E}_{q_{\psi}} \left[ \log \frac{p_{\theta}(\zo, \yo, \beps)}{q_{\psi}(\beps | \zo, \xo, \yo)}\right]
\end{gathered}
\end{equation}
where $\bm{\epsilon}=\{\bm{\epsilon}_1, ...,  \bm{\epsilon}_L\}$ is a set of reparameterized hierarchical latent variables. Further details on the design of the hierarchical dependency VAE are provided below.

\subsection{Predictive enhancement}

The combination of generative and discriminative models is an effective well-studied strategy for dealing with predictive models under missing data \citep{tresp1993training, ghahramani1995learning}. In \citep{ghahramani1995learning}, they model $p(\x)$ for imputing missing data using a Gaussian Mixture Model. In a deep learning context, recent supervised VAE models have revisited this combination \citep{smieja2018processing, ipsen2020deal} or used factorisations of type $p(\z)p(\x|\z)p(\y|\z)$ \citep{joy2021capturing} to learn meaningful representations. In  \citep{li2019generative} the authors propose a deep generative model with factorisation $p(\z)p(\x|\z)p(\y|\x,\z)$ for detecting adversarial attacks.

With the aim at reinforcing the prediction of the variable of interest, we turn into a supervised model by including a separate predictor for $p_{\theta_y}(\y | \hat{\x}, \h)$, apart from the decoder $p_{\theta_z}(\z |\h)$. The vector $\hat{\bm{x}} = \left( x_i \in \bm{x}_O, \; \hat{x}_j \in \bm{x}_U \right) $ includes the observed part and imputation of the missing variables $\hat{x}_j$ by decoding the latent $\h$ into $\z$ using $p(\bm{z | \bm{h}_1})$, and each dimension $z_j \in \bm{z}$ into $\hat{x}_j$ using $p(x_j|z_j)$. The predictor parameters $\theta_y$ are optimized along with the decoder parameters $\theta_z$.

\subsection{Hierarchical reparameterized latent space}\label{sec:hier_reparam}

A hierarchical structure over the latent space $\h = \{\h_1, ..., \h_L \}$ enriches the prior assumptions and permits a flexible generation of data in a more natural fashion. 
Nevertheless, as stated in \citep{betancourt2017conceptual, betancourt2015hamiltonian}, HMC can be pathological when used for sampling from hierarchical densities, where the magnitude of autoregressive variations increase with the depth. For approximating the Hamiltonian dynamics, inside each Leapfrog integrator step, gradients $\nabla_{\bm{h}_{1:L}} \log p^*(\bm{h}_{1:L})$ are required. Due to the strong curvature regions, huge norm of high-order derivatives are backpropagated and might eventually explode, ending in overflow issues (Figure 35 in \citep{betancourt2015hamiltonian}).
    
If we were to run our HMC method over the hierarchical variables without any reparameterization (Figure \ref{fig:reparam}a), by the time the states reached the aforementioned problematic regions, the integrator would diverge and we would experience the aforementioned overflow problems. By rejecting these problematic states, chains would get stuck close to the proposal and the hierarchical density would not be properly explored (concluding that HMC would not improve the Gaussian proposal). To give an example,  in \citep{vahdat2020nvae}, the AR path $p(\bm{h}_l|\bm{h}_{<l})$ would provoke
instabilities inside the HMC integrator due to huge gradients
in $\nabla_{\bm{h}_{1:L}} \log p^*(\bm{h}_{1:L})$. 
    
We successfully solved this issue by introducing a hierarchical reparameterization technique. The representation at each layer is reparameterized from variable $\beps_l$ with standard Gaussian prior $p(\beps_l)$
\begin{equation}
    \h_l = f_{\mu_l}(\h_{l+1}) + f_{\sigma_l}(\h_{l+1}) \cdot \beps_l ,
\end{equation}
where the functions $f_{\mu_l}(\h_{l+1})$ and $f_{\sigma_l}(\h_{l+1})$ are applied by NNs with parameters $\theta_l = \{\theta_{\mu_l}, \theta_{\sigma_l} \}$. The result is equivalent as learning the mean and covariance of autoregressive variables (see Figure \ref{fig:reparam} for illustrative details). However, thanks to this trick, we relax the dependencies among the latent variables, resulting in a smoother joint posterior density $p(\beps | \xo, \yo)$. Performing the inference over $\beps = \{ \beps_1, ..., \beps_L \}$ leads to a better posed basis for running our HMC optimization, detailed in section \ref{sec:hmc_hier}, and avoids the necessity of employing more advanced HMC samplers like \citep{girolami2011riemann, betancourt2015hamiltonian}. We include further details on the pathological behavior and demonstration of the effectiveness of our solution in Section \ref{sec:app_reparam} of the Supplementary. Provided the promising results we obtain in Section \ref{sec:experiments}, we propose our reparameterization trick as a novel contribution for solving the pathological behavior of HMC combined with hierarchical VAEs.
    
For the sake of simplicity, we name the generative parameters as $\theta=\{ \theta_z, \theta_y, \theta_1, ..., \theta_L \}$. 
The dependency ELBO under this hierarchical reparameterized model becomes 
\begin{equation}\label{eq:dep_hier_elbo}
\begin{gathered}
    \loss_{VI}(\xo, \yo; \{ \theta, \psi \}) = 
    \mathbb{E}_{q_{\psi}} \left[ \log \frac{p_{\theta}(\zo, \yo, \beps)}{q_{\psi}(\beps | \zo, \xo, \yo)}\right] = \\
    \mathbb{E}_{q_{\psi}} \left[ \log p_{\theta}(\zo | \h_1) +  \log p_\theta(\yo | \hat{\x}, \h_1)  \right] 
    -  \sum_{l=1}^{L} \KL \left( q_\psi (\beps_l | \xo, \yo ) || p(\beps_l) \right) .
\end{gathered}
\end{equation}
We name $\bm{r}_l$ the hidden representation at each layer, and defining $\bm{r}_0=\{\xo, \yo\}$, we employ NNs with parameters $\psi_{r_l}$ for computing $\bm{r}_l = f_r(\bm{r}_{l-1})$. These vectors are mapped into the parameters of the variational posterior $q_{\psi_l}(\bm{\epsilon_l}|\xo, \yo)$, using NNs for computing the mean as $g_{\mu_l}(\bm{r}_l)$ and the covariance as $g_{\sigma_l}(\bm{r}_l)$, with parameters $\psi_{\mu_l}$ and $\psi_{\sigma_l}$. With compactness purposes, we will denote the encoder parameters as $\psi=\{ \psi_1, ..., \psi_L\}$, with $\psi_l=\{ \psi_{r_l}, \psi_{\mu_l}, \psi_{\sigma_l} \}$. 

\subsection{HMC over the hierarchical density}\label{sec:hmc_hier}

In recent works, HMC has been combined with deep generative models for improving the inference of the latent variables by obtaining better samples from the posterior \citep{hoffman2017learning, caterini2018hamiltonian}.
In this work, we propose to transcend these previous approaches and build a generalized method for sampling from complicated, hierarchical latent structures composed by several layers. Inspired by \citep{campbell2021gradient} and their method for sampling from the posterior within a vanilla VAE framework while tuning the HMC hyperparameters, we follow a procedure for training the dependency model where i) during a pre-training stage, the encoder and decoder are optimized using standard VI and the ELBO from equation \eqref{eq:dep_hier_elbo}, and ii) using the pre-trained encoder for starting from a good proposal \citep{hoffman2017learning}, HMC samples are obtained to follow the true posterior and jointly optimize the generative model and the HMC hyperparameters. In Figure \ref{fig:model_hmc_samples} (b) we include an illustrative example.

We denote by $q_{\phi}^{(T)} (\beps | \zo, \xo, \yo)$ the implicit distribution for the posterior after $T$ HMC steps. The hyper-parameters of HMC are named $\phi$, a $T\times d$ matrix containing the step sizes for each dimension at each step of the chain. 
Within this perspective, the hyper-parameters can be optimized using variational inference by maximizing the ELBO
\begin{equation}\label{eq:hmc_elbo}
\begin{gathered}
    \mathbb{E}_{q_{\phi}^{(T)}(\beps)} [  \log p(\zo, \yo, \beps)  ] + H[ q_{\phi}^{(T)}(\beps | \xo, \yo)] ,
\end{gathered}
\end{equation}
where the first part is the HMC objective, and can be easily estimated via Monte Carlo
\begin{equation}\label{eq:hmc_objective}
\begin{gathered}
    \loss_{HMC}(\zo, \yo; \{ \theta, \psi, \phi\}) =
      \mathbb{E}_{q_{\phi}^{(T)} (\beps)} [ \log p_{\theta}(\zo | \h_1) +  \log p_\theta(\yo | \hat{\x}, \h_1)   + 
      \sum_{l=1}^{L} p( \beps_l^{(T)}  ) ] .
\end{gathered}
\end{equation}
Nevertheless, the entropy term $H[ q_{\phi}^{(T)}(\beps | \xo, \yo)]$ in \cref{eq:hmc_elbo} is not tractable since we are not able to explicitly evaluate the distribution $q_{\phi}^{(T)}(\beps | \xo, \yo)$. Although optimizing the first term might result in a well-posed algorithm, this would bring consequences that must be considered. Namely, without a proper regularization term, and in case the initial proposal $q_{\phi}^{(T)}(\beps | \xo, \yo)$ is concentrated in high density regions, the chains would barely move from the initial state and only these regions with high density would be explored (see Section \ref{sec:app_hmc} of the Supplementary for illustrative details). To cope with this problem, we define an inflation parameter $\bm{s}$ to increase the variance of the proposal $q_{\phi} (\beps | \zo, \xo, \yo)$ given by the Gaussian encoder, ending in the proposal $ \prod_{l=1}^L \mathcal{N} (g_{\mu_l}(\bm{r}_l), \; s_l \cdot g_{\sigma_l}(\bm{r}_l) )$. Whilst in \citep{campbell2021gradient} the authors define this parameter as a scalar factor applied to all the latent dimensions, in our work, we extend this to apply a different inflation factor at each latent level $\bm{s}=\{ s_1, ..., s_L\}$. 
In order to tune these inflations we ensure a wider coverage of the density by minimizing the Sliced Kernelized Stein Discrepancy (SKSD) \citep{gong2020sliced} 
\begin{equation}
	\loss_{SKSD} (\xo, \y_O; \bm{s}) = 
	\text{SKSD} \left( q_{\phi}^{(T)} (\beps | \zo, \xo, \yo; \bm{s}),  \,   p(\beps | \zo, \xo, \yo) \right) ,
\end{equation}

which fits perfectly our requirements, since only requires samples from HMC and gradients $\nabla_{\epsilon} \log p(\zo, \yo, \beps)$ for measuring a discrepancy between the implicit and the true posterior. Further, the SKSD performs better than other discrepancies like \citep{liu2016kernelized} in high dimensional latent spaces.

We provide in Section \ref{sec:app_hmc} of the Supplementary a toy demonstration on the efficacy of the HMC optimization, and in Section \ref{sec:app_hmc_convergence} a demonstration of the optimization convergence.


\subsection{HH-VAEM optimization}\label{sec:opt}

The optimization of the HH-VAEM is divided into three stages. In a first stage, we train one independent \emph{marginal} VAE per dimension. In a second stage, the \emph{dependency} VAE is trained, using as inputs the concatenation of the encoded dimensions $\z=\{z_1, ..., z_D \}$
and the target $\y$. Finally, in a third stage, the HMC hyperparameters, the decoder and the predictor are tuned using the HMC objective, the inflation parameter is trained using the SKSD discrepancy, and the encoder is trained used the ELBO. The pseudocode for HH-VAEM training is shown in Algorithm \ref{alg:training}.

\subsection{Computational cost}\label{sec:computational_cost}

In a VAE, the data is fed to the encoder with aim at obtaining the variational parameters for sampling from the Gaussian approximated posterior. In our method, the data is similarly encoded to obtain the initial Gaussian proposal $q^{(0)}$, and the samples from this distribution are updated for $T$ cycles to obtain the implicit $q^{(T)}$ using HMC. Within each of these iterations, $L$ leapfrog steps \eqref{eq:leapfrog} are executed. For each of these steps, the computation of the gradients $\nabla_{\beps_l} \log p(\z, \y, \beps_l)$ and $\nabla_{\beps_{l+1}} \log p(\z, \y, \beps_{l+1})$ is required. To obtain these gradients, we need to \textit{i)} compute the parameters of the likelihood $ p(\z, \y | \beps)$ that are given by the decoder ($p(\z | \h_1)$) and predictor ($p(\y | \hat{\x}, \h_1)$), \textit{ii)} evaluate the likelihood and \textit{iii)} perform the automatic differentiation. Thus, for running our method, an additional cost from both decoding and performing differentiation a total of $2TL$ times is introduced.

\begin{wrapfigure}[20]{R}{0.6\textwidth}
    \vspace{-0.8cm}%
    \begin{minipage}{0.6\textwidth}
        \begin{algorithm}[H]
           \caption{Training algorithm for HH-VAEM}
           \label{alg:training}
            \begin{algorithmic}
               \STATE {\bfseries Input:} data $ \left( \xo^{(1:N)}, \yo^{(1:N)} \right)$,  steps: $T_d$, $T_{VI}$, $T_{HMC}$
               \STATE {\bfseries Parameters:} $\gamma$, $\theta$, $\psi$, $\phi$, $s$
               \STATE {\textsc{Stage 1: marginal VAEs}}
               \FOR{$d=1$ {\bfseries to} $D$}
               \STATE Initialize marginal VAE $\{ \theta_d,  \gamma_d\}$
               \FOR{$t=1$ {\bfseries to} $T_d$}
               \STATE $\gamma_d^{t+1}, \theta_d^{t+1} \leftarrow\text{Adam}_{\gamma_d^t, \theta_d^t}(\loss_d)$ 
               \ENDFOR
               \ENDFOR
               \STATE {\textsc{Stage 2: dependency VAE}}
               \FOR{$t=1$ {\bfseries to} $T_{VAE}$}
               \STATE $\theta^{t+1}, \psi^{t+1}\leftarrow\text{Adam}_{\theta^t, \psi^t}(\loss_{VI})$ 
               
               \ENDFOR
               \STATE {\textsc{Stage 3: Jointly optimizing VAE + HMC}}
               \FOR{$t=1$ {\bfseries to} $T_{HMC}$}
               \STATE $\psi^{t+1}\leftarrow\text{Adam}_{\psi^t}(\loss_{VI})$ 
               \STATE $\theta^{t+1}, \phi^{t+1} \leftarrow\text{Adam}_{\theta^t, \phi^t}(\loss_{HMC})$ 
               \STATE $s^{t+1}\leftarrow\text{Adam}_{s^t}(\loss_{SKSD})$ 
               \ENDFOR
            \end{algorithmic}
        \end{algorithm}
        \vspace{-0.5cm}
    \end{minipage}
\end{wrapfigure}

By jointly optimizing the HMC hyperparameters we are able to achieve faster convergence with smaller lengths. To reduce the computational cost, we optimize the hyperparameters in a final training stage, since convergence is rapidly achieved (as demonstrated empirically in Section \ref{sec:app_hmc_convergence} of the Supplementary). At test, samples from the Gaussian $q^{(0)}$ (faster and cheaper), or from HMC $q^{(T)}$ (slower and better) can be used to fit  computational constraints.

%% file: sampling_method.tex
\section{Sampling-based Active Learning} \label{sec:active}



Considering that the input data are tuples of observed and missing features  $\{ \xo, \xu\}$, our Active Learning framework follows \citep{ma2018eddi, ma2020vaem} and  determines which feature $\x_i \in \xu$ will enhance the prediction of the target $\y$ the most for a particular $\xo$. Concretely, in a Sequential Active Information Acquisition (SAIA) scenario, this decision is taken sequentially to optimally increase  knowledge and accurately predict $\y$. From a information theoretical perspective, this task can be performed recurrently by maximizing a reward function $R$ at each step $d$.
This reward might represent abstract quantities of interest like the cost or benefit of acquiring $\x_i$ (depending on the sign). In Bayesian experimental analysis, $R$ is the expected gain of information \citep{lindley1956measure}. 
Following \citep{bernardo1979expected}, we can define it as
\begin{equation}\label{eq:reward}
    R(i, \xo) = \mathbb{E}_{p(x_i | \xo)} \, D_{\text{KL}} \left( p(\y | x_i, \xo) \\ p(\y | \xo) \right) ,
\end{equation} 
where $i$ is the index of each unobserved feature. Intuitively, this quantity can be interpreted as the expected change in the predictive distribution when $\x_i$ is observed. The reward needs to be estimated via Monte Carlo by sampling from $p(x_i | \xo)$. With a robust generative model that handles missing data like HH-VAEM, these samples are easily obtained: first, using HMC, we sample $\beps^{(T)}$ from  $p(\beps | \xo)$. Second, we decode these samples to obtain $x_i$ from $p(\z | \h_1)$ and $p(x_i | z_d)$. Nonetheless, the reward defined in \eqref{eq:reward} is intractable since both $p(\y | x_i)$ and $p(\y | x_i, \xo)$ are intractable: computing them requires to integrate out the latent variables.
This motivates the authors of \citep{ma2018eddi, ma2020vaem} to present a transformation of the reward for being computed in the latent space using the encoder network. Although they prove that this transformation effectively provides a good estimation in several datasets, we demonstrate that for low dimensional targets (commonly one or two dimensions), an approximation using histograms is more effective. Concretely, the reward in \eqref{eq:reward} can be rewritten as
\begin{equation}\label{eq:mi}
\begin{gathered}
\KL \left[ p( \y, x_i |  \xo ) || p(\y | \xo) p(x_i | \xo) \right] = \mathcal{I} (\y ; x_i  \; | \xo) ,
\end{gathered}
\end{equation}
While a set of advanced non-parametric estimators of the mutual information are available \citep{kraskov2004estimating, ross2014mutual}, many are not easilly adapted for parallelization. We demonstrate that the simplest one, 
\begin{equation}
	\hat{I}(\y ; \x_i  \; | \xo) \approx \sum_{ij} p(i, j) \log \frac{p(i, j)}{p_x(i) p_y(j)} ,
\end{equation}
based on binning the $x_i$ and $\y$ domains, is effective and easy to parallelize. In the equation, $p_x(i) = n_x(i)/N$, $p_y(j) = n_y(j)/N$ and $p(i, j) = n(i, j)/N$ are the relative frequencies that approximate $p(x)$, $p(y)$ and $p(x,y)$ for each bin. Thus, $n_x(i)$, $n_y(j)$ and $n(i, j)$ are the number of samples inside each interval. The number of bins defines the width of uniformly distributed intervals over $x_d$ and $\y$ supports. 
Since this estimator is sampling-based, under certain conditions (namely,  if all densities exist as proper functions), Equation (13) indeed converges to $\mathcal{I} (\bm{y} ; x_i  \; | \bm{x}_O)$ if we first let the number of samples $ N \rightarrow \infty$ \citep{kraskov2004estimating}.

%% file: experiments.tex
\begin{figure}[t]
\begin{minipage}{\textwidth}
\setlength{\tabcolsep}{3pt}
		\begin{minipage}{\textwidth}
        \setlength{\tabcolsep}{3pt}
            \centering
            \resizebox{\linewidth}{!}{
        	\begin{tabular}{@{}rrrrrrrrrrr@{}}
                \toprule
                \multicolumn{1}{r}{} &
                  \multicolumn{1}{r}{Bank} &
                  \multicolumn{1}{r}{Insurance} &
                  \multicolumn{1}{r}{Avocado} &
                  \multicolumn{1}{r}{Naval} &
                  \multicolumn{1}{r}{Yatch} &
                  \multicolumn{1}{r}{Diabetes} &
                  \multicolumn{1}{r}{Concrete} &
                  \multicolumn{1}{r}{Wine} &
                  \multicolumn{1}{r}{Energy} &
                  \multicolumn{1}{r}{Boston} \\ \midrule
                VAEM &
                  $2.84 \pm 0.07$ &
                  $1.81 \pm 0.03$ &
                  $1.89 \pm 0.01$ &
                  $0.55 \pm 0.05$ &
                  $3.15 \pm 0.28$ &
                  $2.78 \pm 0.16$ &
                  $2.45 \pm 0.26$ &
                  $3.01 \pm 0.61$ &
                  $2.09 \pm 0.10$ &
                  $2.01 \pm 0.23$ \\
                MIWAEM &
                  $2.74 \pm 0.05$ &
                  $1.88 \pm 0.04$ &
                  $1.92 \pm 0.04$ &
                  $0.57 \pm 0.03$ &
                  $2.66 \pm 0.11$ &
                  $2.55 \pm 0.09$ &
                  $2.34 \pm 0.51$ &
                  $2.76 \pm 0.48$ &
                  $2.06 \pm 0.14$ &
                  $1.94 \pm 0.23$ \\
                H-VAEM &
                  $2.82 \pm 0.06$ &
                  $1.80 \pm 0.04$ &
                  $1.89 \pm 0.01$ &
                  $0.48 \pm 0.06$ &
                  $3.06 \pm 0.31$ &
                  $2.74 \pm 0.09$ &
                  $2.42 \pm 0.21$ &
                  $2.85 \pm 0.56$ &
                  $1.72 \pm 0.11$ &
                  $1.89 \pm 0.24$ \\
                HMC-VAEM &
                  $2.69 \pm 0.05$ &
                  $1.77 \pm 0.06$ &
                  $1.89 \pm 0.02$ &
                  $0.49 \pm 0.07$ &
                  $\bm{2.21 \pm 0.24}$ &
                  $2.72 \pm 0.20$ &
                  $2.28 \pm 0.29$ &
                  $2.83 \pm 0.46$ &
                  $1.73 \pm 0.05$ &
                  $1.83 \pm 0.16$ \\
                \textbf{HH-VAEM} &
                  $\bm{2.63 \pm 0.04}$ &
                  $\bm{1.75 \pm 0.03}$ &
                  $\bm{1.88 \pm 0.05}$ &
                  $\bm{0.40 \pm 0.05}$ &
                  $2.47 \pm 0.27$ &
                  $\bm{2.54 \pm 0.13}$ &
                  $\bm{2.28 \pm 0.09}$ &
                  $\bm{1.90 \pm 0.17}$ &
                  $\bm{1.71 \pm 0.04}$ &
                  $\bm{1.83 \pm 0.11}$ \\ \bottomrule
            \end{tabular}
            }
			\captionof{table}{Test NLL of the unobserved features for our model and baselines.}
            \label{tab:ll_xu}
            \vspace{0.3cm}
		\end{minipage}
		
		\begin{minipage}{\textwidth}
		    \setlength{\tabcolsep}{3pt}
            \centering
            \resizebox{\linewidth}{!}{
        	\begin{tabular}{@{}rrrrrrrrrrr@{}}
                \toprule
                \multicolumn{1}{r}{} &
                  \multicolumn{1}{r}{Bank} &
                  \multicolumn{1}{r}{Insurance} &
                  \multicolumn{1}{r}{Avocado} &
                  \multicolumn{1}{r}{Naval} &
                  \multicolumn{1}{r}{Yatch} &
                  \multicolumn{1}{r}{Diabetes} &
                  \multicolumn{1}{r}{Concrete} &
                  \multicolumn{1}{r}{Wine} &
                  \multicolumn{1}{r}{Energy} &
                  \multicolumn{1}{r}{Boston} \\ \midrule
                VAEM &
                  $0.56 \pm 0.06$ &
                  $1.20 \pm 0.03$ &
                  $1.18 \pm 0.02$ &
                  $2.69 \pm 0.01$ &
                  $0.61 \pm 0.02$ &
                  $1.59 \pm 0.19$ &
                  $1.07 \pm 0.09$ &
                  $0.28 \pm 0.09$ &
                  $0.61 \pm 0.14$ &
                  $0.85 \pm 0.21$ \\
                MIWAEM &
                  $0.51 \pm 0.03$ &
                  $1.15 \pm 0.03$ &
                  $1.15 \pm 0.03$ &
                  $2.70 \pm 0.01$ &
                  $0.60 \pm 0.03$ &
                  $\bm{1.36 \pm 0.10}$ &
                  $0.95 \pm 0.22$ &
                  $0.28 \pm 0.13$ &
                  $0.54 \pm 0.12$ &
                  $0.80 \pm 0.21$ \\
                H-VAEM &
                  $0.50 \pm 0.03$ &
                  $1.06 \pm 0.02$ &
                  $1.18 \pm 0.02$ &
                  $2.68 \pm 0.01$ &
                  $0.60 \pm 0.02$ &
                  $1.71 \pm 0.14$ &
                  $1.02 \pm 0.09$ &
                  $0.26 \pm 0.11$ &
                  $0.46 \pm 0.14$ &
                  $0.90 \pm 0.22$ \\  
                HMC-VAEM &
                  $0.52 \pm 0.02$ &
                  $1.00 \pm 0.03$ &
                  $1.12 \pm 0.03$ &
                  $2.71 \pm 0.01$ &
                  $\bm{0.52 \pm 0.15}$ &
                  $1.55 \pm 0.29$ &
                  $0.95 \pm 0.26$ &
                  $0.28 \pm 0.09$ &
                  $0.41 \pm 0.07$ &
                  $0.71 \pm 0.13$ \\
                \textbf{HH-VAEM} &
                  $\bm{0.49 \pm 0.03}$ &
                  $\bm{0.93 \pm 0.06}$ &
                  $\bm{1.10 \pm 0.01}$ &
                  $\bm{2.62 \pm 0.01}$ &
                  $0.56 \pm 0.02$ &
                  $1.38 \pm 0.18$ &
                  $\bm{0.95 \pm 0.08}$ &
                  $\bm{0.20 \pm 0.04}$ &
                  $\bm{0.32 \pm 0.05}$ &
                  $\bm{0.55 \pm 0.04}$ \\ \bottomrule
            \end{tabular}
            }
            \captionof{table}{Test NLL of the predicted target for our model and baselines.}
            \label{tab:ll_y}
		\end{minipage}
		
\end{minipage}
\vspace{-0.3cm}
\end{figure}

\section{Experiments}\label{sec:experiments}

The evaluation of the HH-VAEM model is organized into three quantitative experiments and one qualitative experiment. The ablation study includes the validation of our proposed HMC-based, hierarchical model with respect to the Gaussian and one-layered alternatives. Namely, the comparison is performed with the following baseline models:
\begin{itemize}\itemsep1pt
    \item \emph{VAEM}: 1 layer, Gaussian-based VAEM \citep{ma2020vaem} (without including the Partial VAE).
    \item \emph{MIWAEM}: 1 layer, Gaussian-based, importance weighted IWAEM (VAEM + IWAE \citep{mattei2019miwae}). 
    \item \emph{H-VAEM}: 2 layers, Gaussian-based VAEM.
    \item \emph{HMC-VAEM}: 1 layer, HMC-based (with our optimization method) VAEM.
\end{itemize}

\begin{minipage}{\textwidth}
		\begin{minipage}{0.25\textwidth}
			\centering
			{\includegraphics[height=0.55\linewidth]{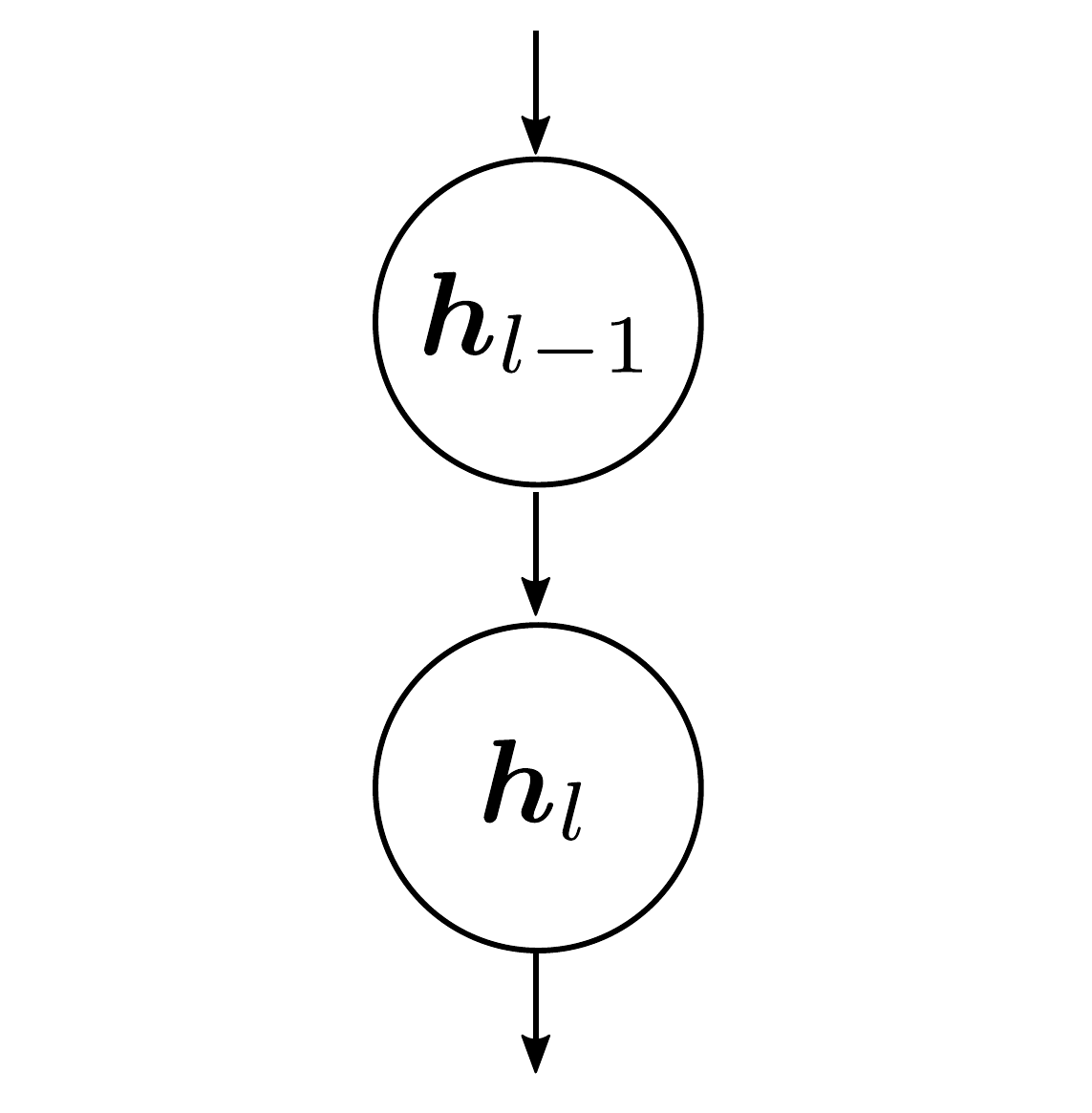}}
			\captionsetup{labelformat=empty}
			\captionof{figure}{(a) Autoregressive}
			{\includegraphics[height=0.55\linewidth]{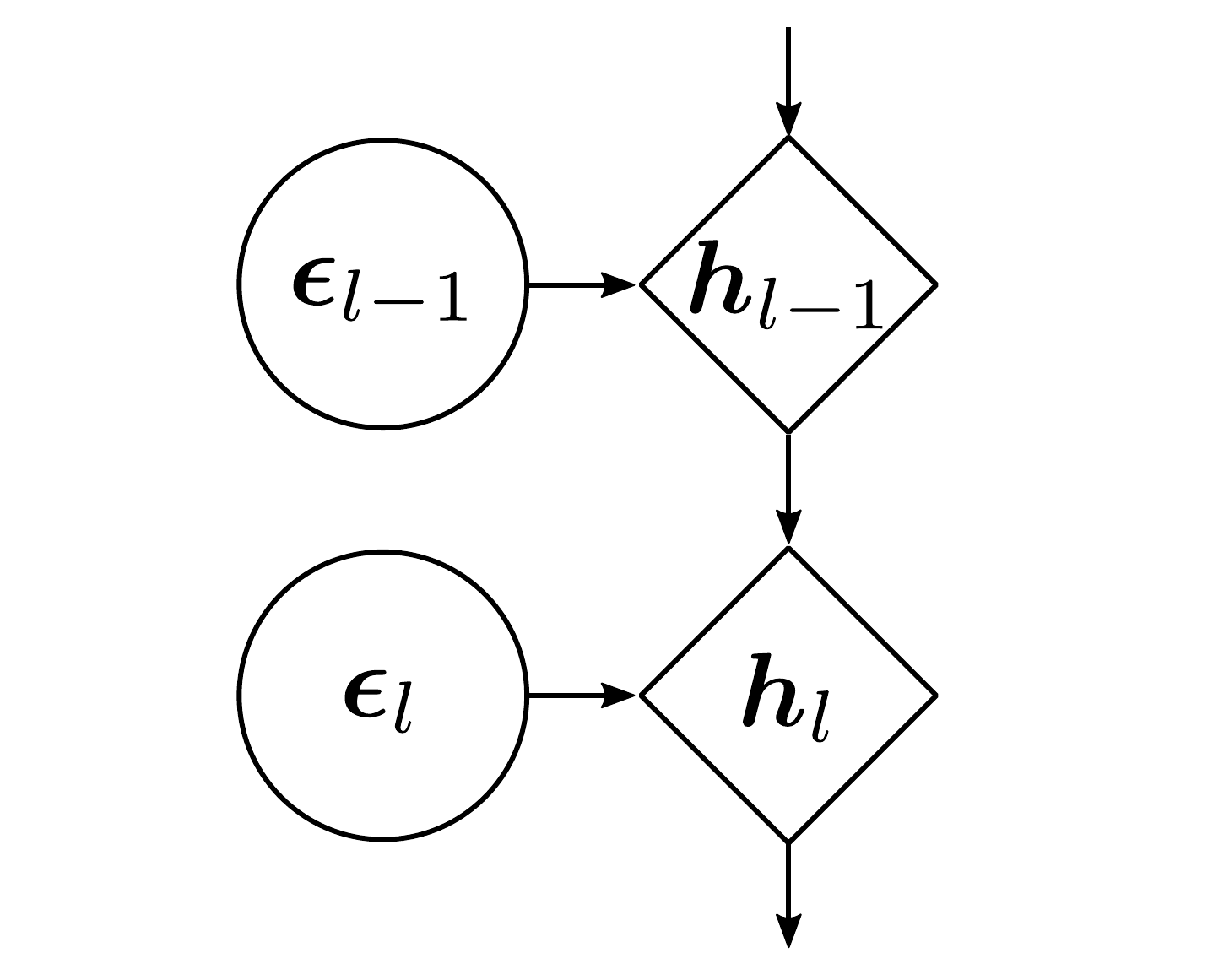}}
			\captionsetup{labelformat=empty}
			\captionof{figure}{(b) Reparameterization}
			\captionsetup{labelformat=simple}
			\caption{AR (a) vs reparameterized (b). 
			} \label{fig:reparam}
			
		\end{minipage}
		\begin{minipage}{0.72\textwidth}

			\centering
			\resizebox{\linewidth}{!}{
				\begin{tabular}{@{}lrrrrr@{}}
					\toprule
					& VAE      & MIWAE        & H-VAE  & HMC-VAE            & \textbf{HH-VAE}               \\ \midrule
					MNIST    & $ 0.124 \pm 0.001 $ & $ 0.121 \pm 0.001 $  &    $ 0.119 \pm 0.001$  & $ 0.101 \pm 0.004$ & $\bm{0.094 \pm 0.003}$ \\
					F-MNIST & $ 0.162 \pm 0.002$  &  $ 0.160 \pm 0.002$  &   $ 0.156 \pm 0.002$ & $ 0.150 \pm 0.002$ & $\bm{0.144 \pm 0.002}$ \\
					\bottomrule
				\end{tabular}
			}
			\vspace{-0.2cm}
			\captionof{table}{Test NLL of the unobserved features for the MNIST datasets.}
			\label{tab:mnist_ll_xu}
			\vspace{0.3cm}
			\resizebox{\linewidth}{!}{
				\begin{tabular}{@{}lrrrrr@{}}
					\toprule
					& VAE      & MIWAE        & H-VAE  & HMC-VAE               & \textbf{HH-VAE}               \\ \midrule
					MNIST    & $ 0.153 \pm 0.009 $ & $ 0.151 \pm 0.007 $  &    $ 0.146 \pm 0.006$  & $ 0.067 \pm 0.007$ & $\bm{0.056 \pm 0.019}$ \\
					F-MNIST & $ 0.501 \pm 0.012$  &  $ 0.496 \pm 0.008$  &    $ 0.494 \pm 0.007$ & $ 0.357 \pm 0.060$ & $\bm{0.337 \pm 0.069}$ \\ \bottomrule
				\end{tabular}
			}
			\vspace{-0.2cm}
			\captionof{table}{Test NLL of the predicted target for the MNIST datasets.}
			\label{tab:mnist_ll_y}
			\vspace{0.3cm}
			\resizebox{\linewidth}{!}{
				\begin{tabular}{@{}lrrrrr@{}}
					\toprule
					& VAE      & MIWAE        & H-VAE  & HMC-VAE               & \textbf{HH-VAE}               \\ \midrule
					MNIST    & $ 0.953 \pm 0.004 $ &  $ 0.953 \pm 0.003 $  &   $ 0.953 \pm 0.003$  & $ 0.978 \pm 0.003 $ & $\bm{0.981 \pm 0.005}$ \\
					F-MNIST & $ 0.824 \pm 0.005$  &  $ 0.824 \pm 0.004$   &   $ 0.824 \pm 0.004$ & $ 0.869 \pm 0.015$ & $\bm{0.876 \pm 0.017}$ \\ \bottomrule
				\end{tabular}
			}
			\vspace{-0.2cm}
			\captionof{table}{Test accuracy of the predicted digits for the MNIST datasets.}
			\label{tab:mnist_accuracy}
		\end{minipage}
\end{minipage}

\vspace{0.2cm}

For all the models, we manually introduce missing data in the training set by randomly setting per data point a feature as missing with a probability sampled uniformly in the interval $\left[ 0.01, 0.99\right]$ within each batch. Both the input data $\x$ and the target $\y$ can be missing. For the test set, a fixed probability of 0.5 leads to about half of the input data being observed, whilst the target is completely unobserved.

For the quantitative experiments, a total of 10 UCI datasets \citep{dua2017uci} that include mixed-type data are employed for the evaluation. We include both MNIST \citep{lecun1998mnist} and Fashion-MNIST \citep{xiao2017fashion} datasets for evaluating our model in higher dimensional observational and latent spaces and bigger architectures (3 layered convolutional nets for encoder/decoder). 
For the qualitative results, we  evaluate our model in the image inpainting task on MNIST and CelebA \citep{liu2015faceattributes}. For the three image datasets, the marginal VAEs are not included and the dependency VAE is fed directly with the Bernoulli-distributed pixels. We name this model HH-VAE, and similarly, the baselines are renamed as VAE, MIWAE, HMC-VAE and H-VAE. Extended experiments and validations are provided in the Supplementary. The source code for reproducing our work is available at \url{https://github.com/ipeis/HH-VAEM}.


\subsection{Mixed type conditional data imputation}\label{sec:exp11}
In order to evaluate the performance of the model in terms of data imputation, we opt by computing the negative log likelihood of the unobserved features. We make use of the Monte Carlo approximation
\begin{equation} \label{eq:imputation_likelihood}
    \log p(\xu | \xo) \approx  \log \mathbb{E}_{\beps \sim q^{(T)}(\beps | \xo) } \left[  p(\xu | \beps) \right] \approx   \log \frac{1}{k} \sum_i^k p(\xu | \beps_i)  ,
\end{equation}
which is averaged over features in order to compare the imputation performance with the baselines. 
Additionally, we include in Section \ref{sec:app_heterog} similar results averaging each of the considered likelihoods.
Results on the 10 UCI datasets and the MNIST datasets are included in Tables \ref{tab:ll_xu} and \ref{tab:mnist_ll_xu}, showing that for most of the datasets, incremental improvement is obtained: VAEM $<$ H-VAEM $<$ HMC-VAEM $<$ HH-VAEM. Extended results with the imputation error, included in Section \ref{sec:app_error}, corroborate this.

\subsection{Target prediction}\label{sec:exp12}

For this experiment, we compute the negative log likelihood of the target under the predictive distribution using the same procedure as in the previous section.
Results included in Tables \ref{tab:ll_y} and \ref{tab:mnist_ll_y} show the same incremental improvement in the prediction task. 

\begin{figure*}[t]
	\centering
	\includegraphics[width=0.8\linewidth]{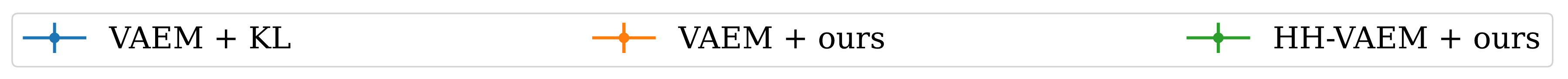} \\
	\subfigure[Avocado]{\includegraphics[height=0.2\linewidth]{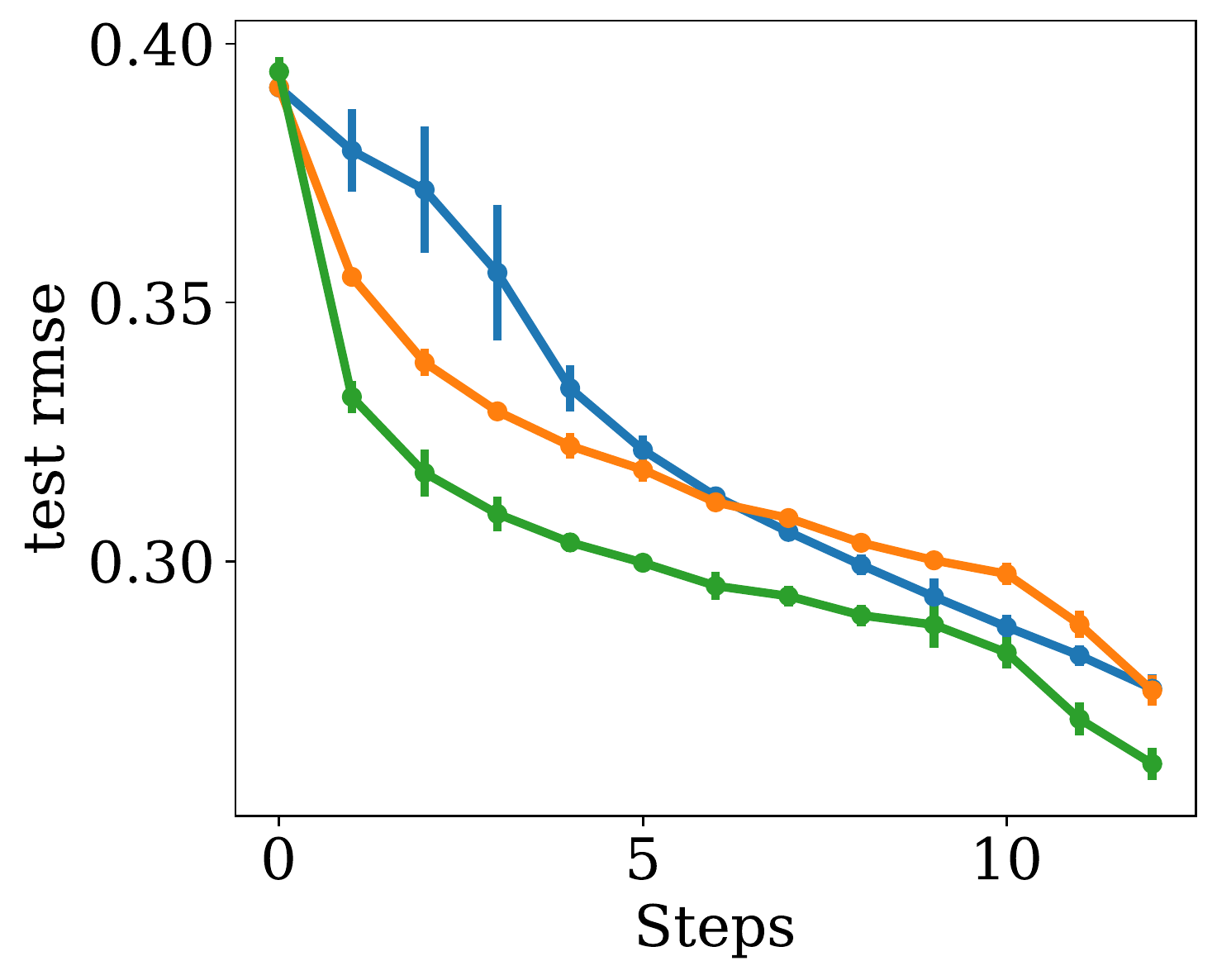}}
	\subfigure[Yatch]{\includegraphics[height=0.2\linewidth]{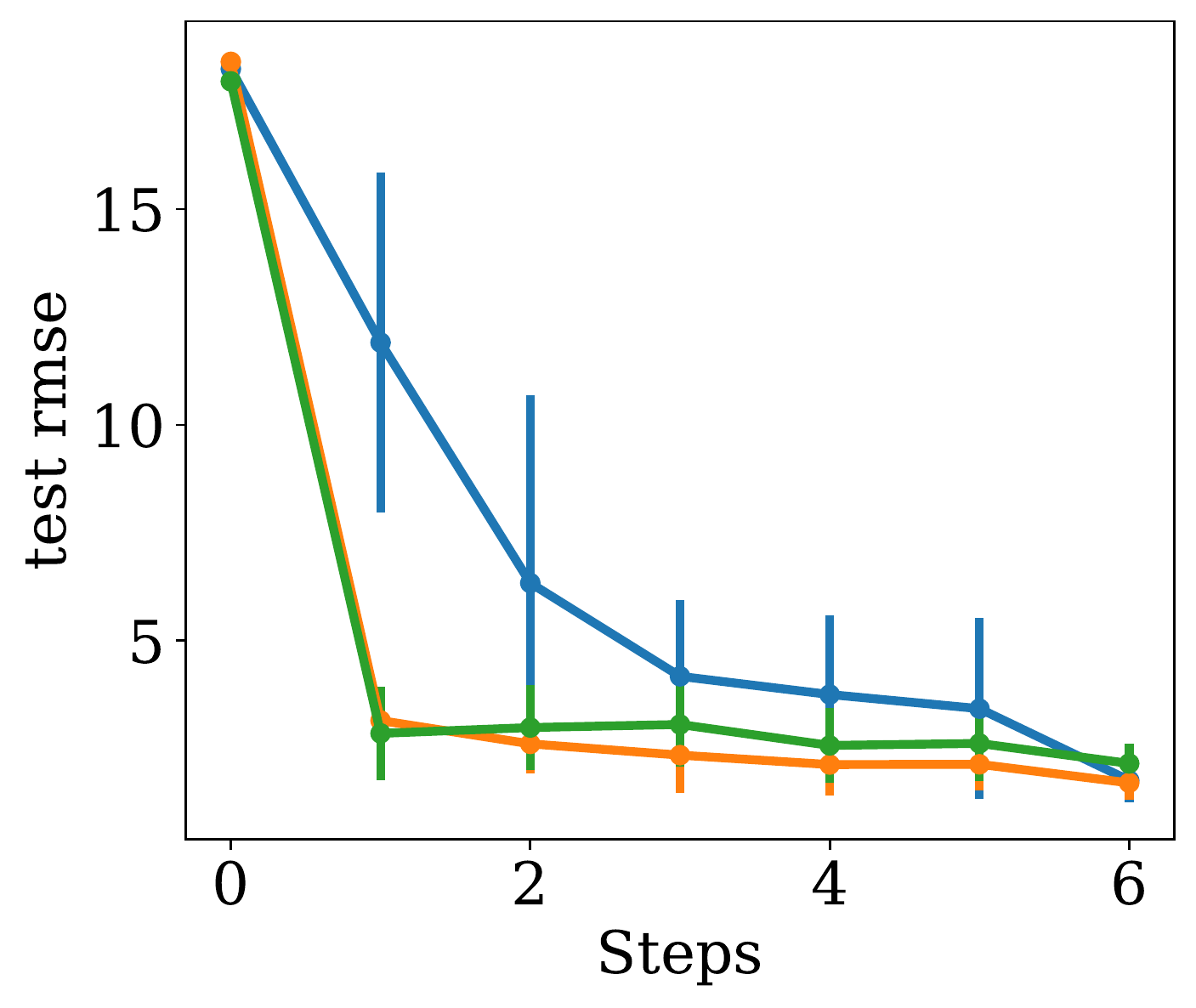}}
	\subfigure[Boston]{\includegraphics[height=0.2\linewidth]{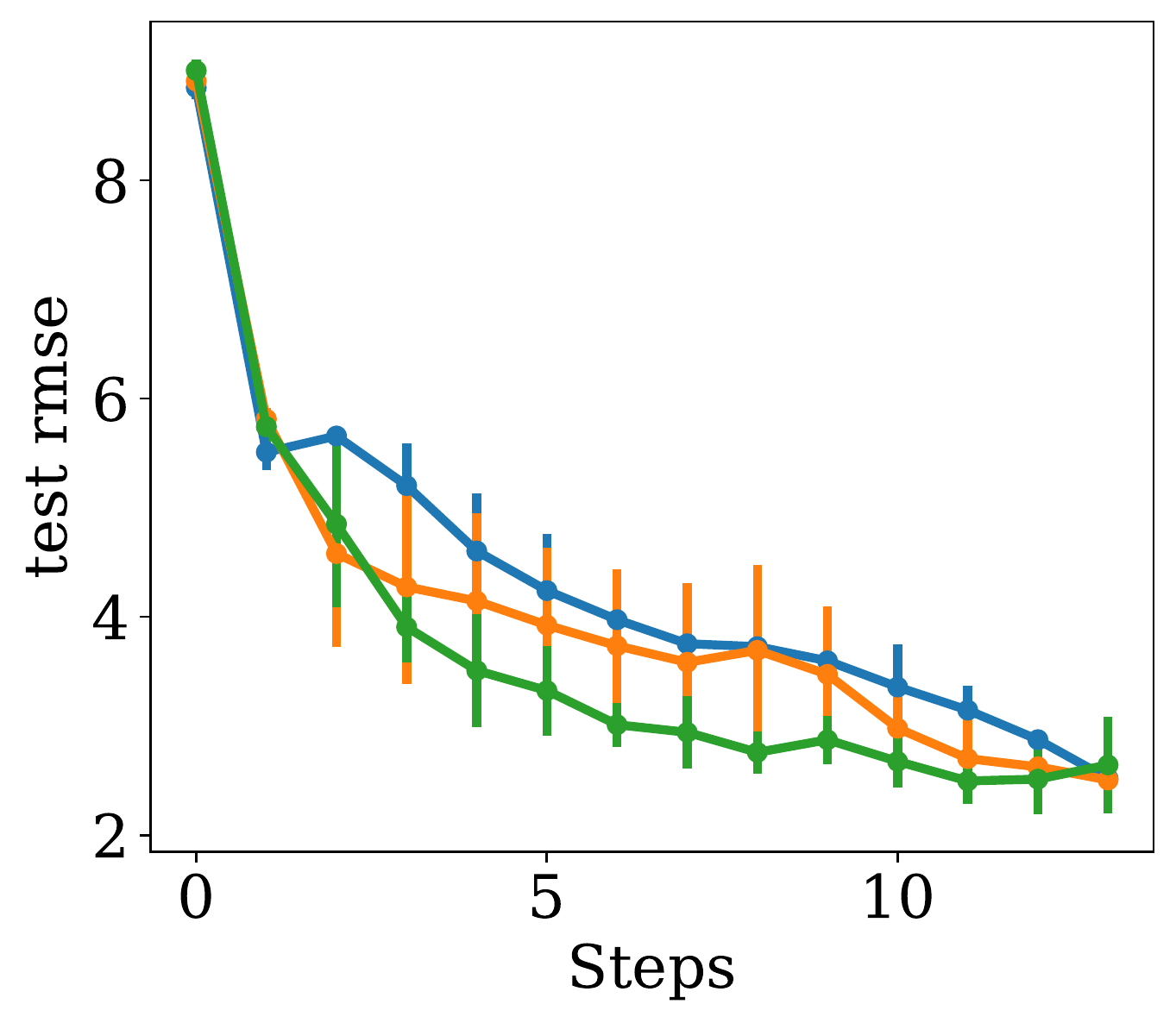}}
	\subfigure[Energy]{\includegraphics[height=0.2\linewidth]{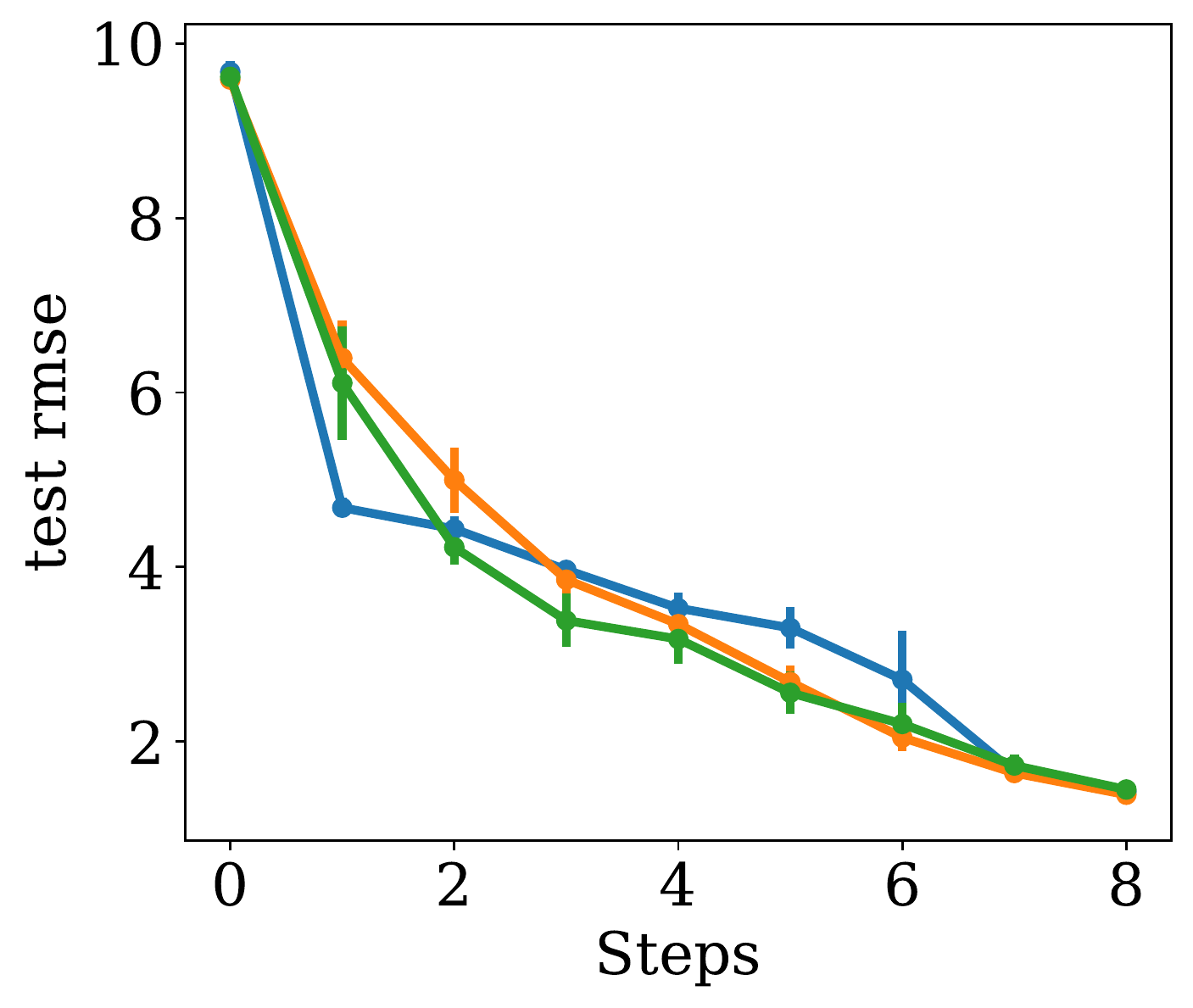}}\\
	\vspace{-0.05cm}
	\subfigure[Wine]{\includegraphics[height=0.19\linewidth]{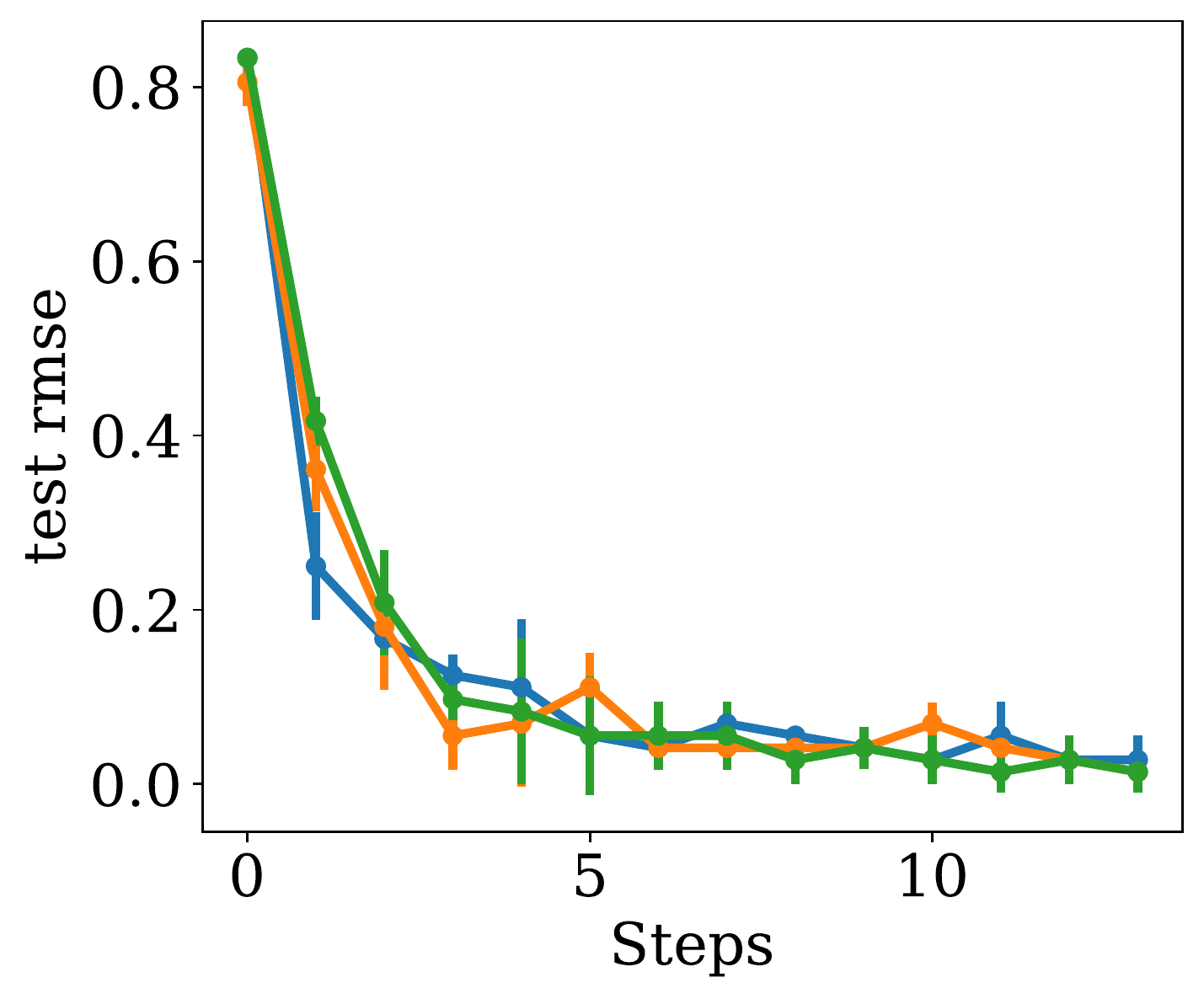}}
	\subfigure[Naval]{\includegraphics[height=0.19\linewidth]{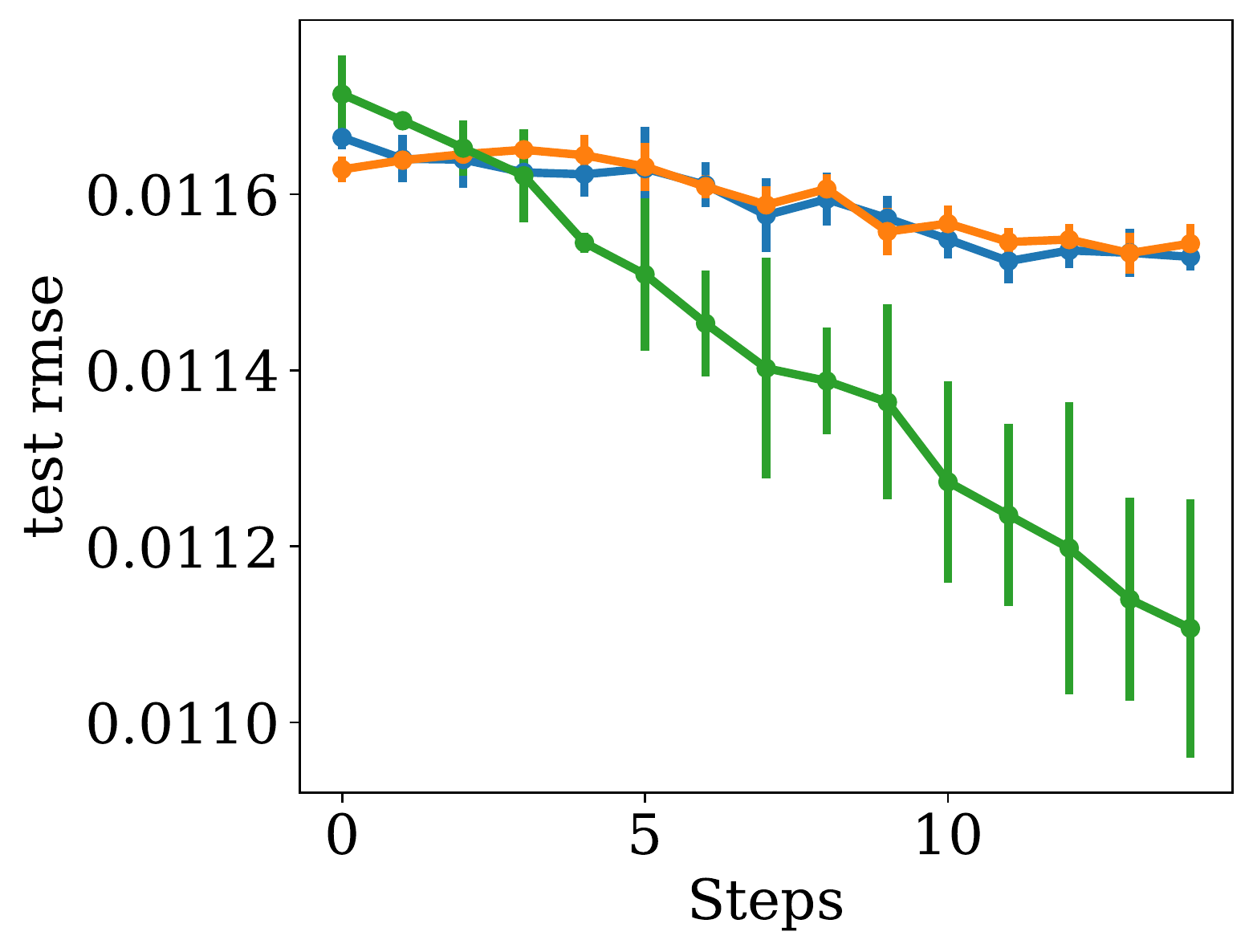}}
	\subfigure[Concrete]{\includegraphics[height=0.19\linewidth]{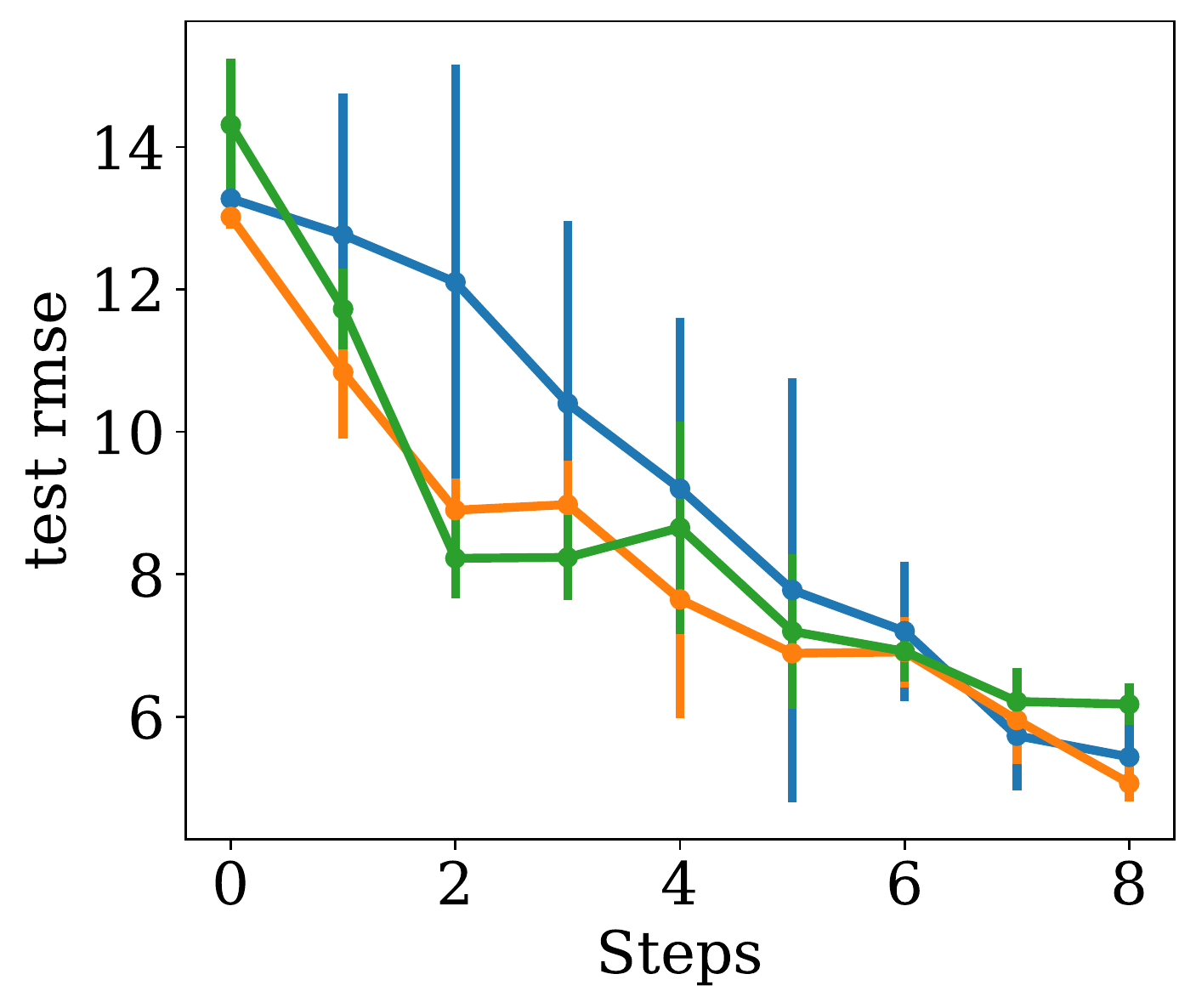}}
	\subfigure[Diabetes]{\includegraphics[height=0.19\linewidth]{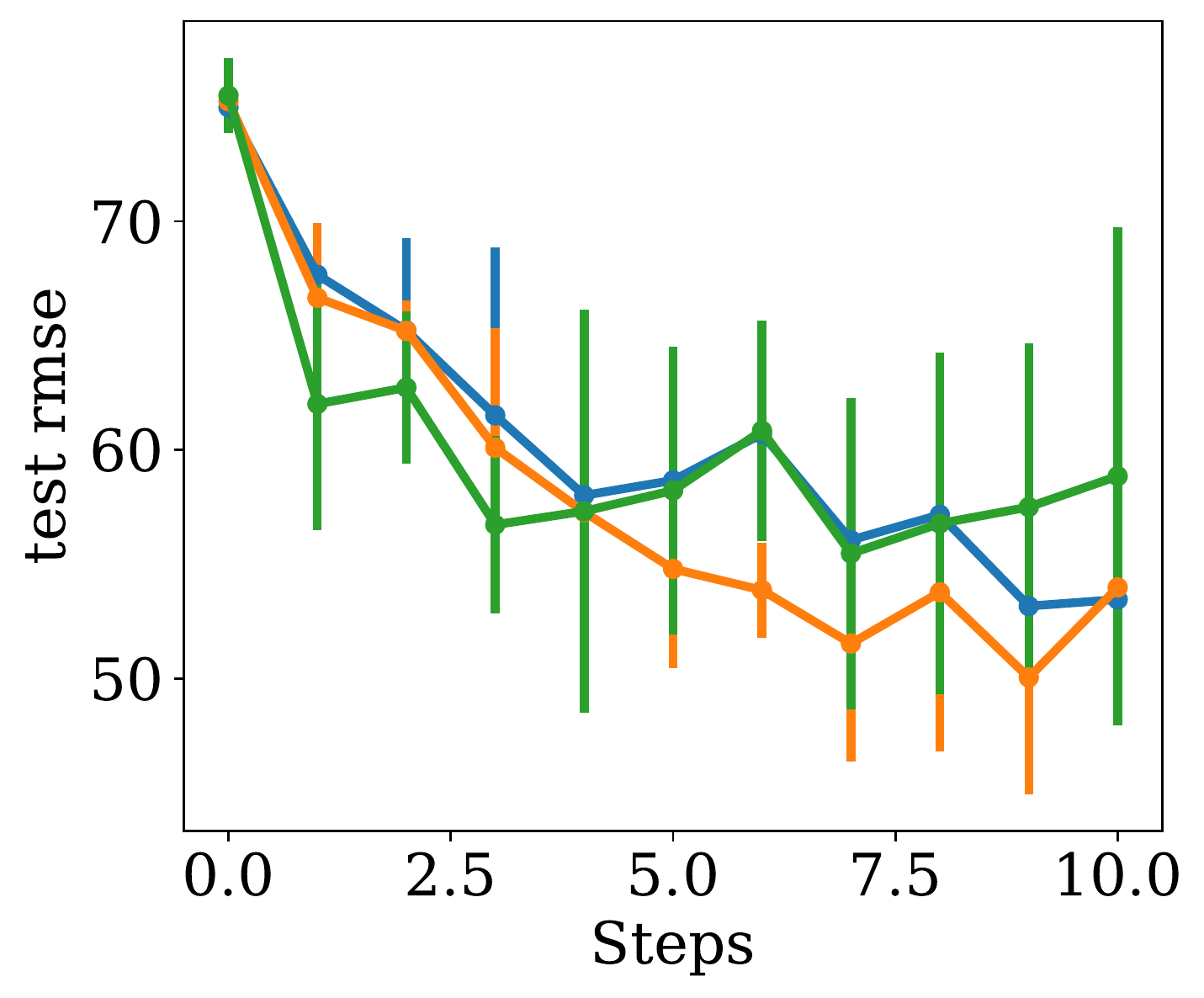}}
\caption{SAIA curves. Horizontal axis shows 
	number of discovered features. Vertical axis is RMSE.} \label{fig:al_curves}
	\vspace{-0.5cm}
\end{figure*}

\subsection{Sequential active information acquisition (SAIA)}\label{sec:saia}

In this experiment, our HH-VAEM model and our acquisition method are evaluated in a SAIA task. Starting by predicting from completely unobserved inputs, at each step, the missing feature that maximizes the reward is acquired. Figure \ref{fig:al_curves} shows the error curves for the UCI datasets. Blue lines correspond to the Gaussian-based reward proposed by \citep{ma2020vaem}. Orange lines are our sampling-based reward in a VAEM framework.
Green lines correspond to HH-VAEM with our reward. In most of the cases, our model and acquisition method 
obtains lower errors and faster discovery of information.

\subsection{Conditional image inpainting}\label{sec:image}

\begin{figure}[t]
	\centering
	\subfigure[MNIST]{\includegraphics[width=\linewidth]{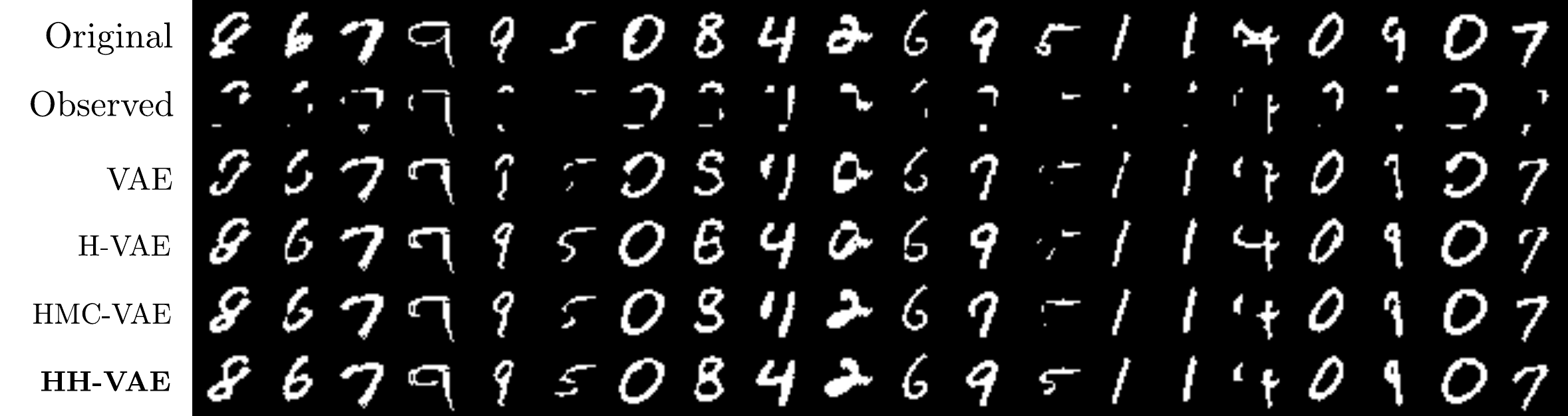}}
	\subfigure[CelebA]{\includegraphics[width=\linewidth]{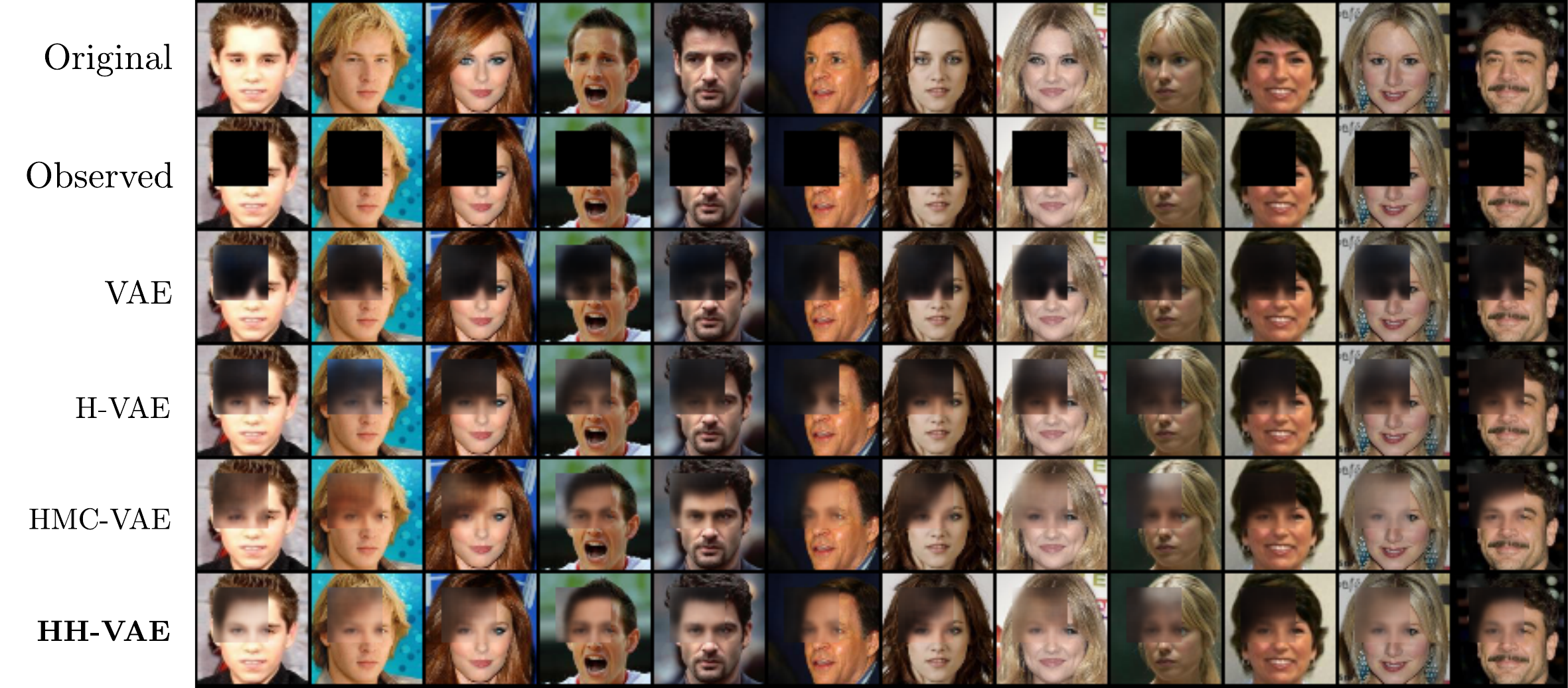}}
	\caption{Image conditional inpainting on MNIST (a) and CelebA (b). First row: original images from the test set. Second row: input to the model, $\xo$, with a black square missing mask $\xu$ manually introduced. Third, fourth, fifth and sixth rows: imputed $\hat{\x}_U$ for VAE, H-VAE, HMC-VAE and HH-VAE, our model, where $\hat{x}_U$ is decoded from samples of the approximate posterior (Gaussian for VAE and H-VAE, or HMC-based for HMC-VAE and HH-VAE).}
	\label{fig:inpainting}
	\vspace{-0.2cm}
\end{figure}

We include in this experiment qualitative results when comparing our model with the baselines on the image conditional inpainting task. 
We include results for MNIST and CelebA in Figure \ref{fig:inpainting} (a) and (b), respectively, that show the superiority of our method. 
First, the hierarchical Gaussian model (fourth row) considerably improves the one-layered Gaussian alternative. Second, the HMC-based methods (two last rows) vastly improve the Gaussian methods. Third, in some specific cases, an extra improvement is added by HH-VAE with respect to the one-layered HMC-based model (columns 2, 4, 8, 9, 11, 12 and 13 in Figure \ref{fig:inpainting} (a) or columns 1, 2, 5 and 12 in (b)).

%% file: conclusion.tex
\section{Conclusion}

We presented HH-VAEM, to our knowledge, the first hierarchical VAE for mixed-type incomplete data that uses HMC with automatic hyper-parameter tuning for improved inference. We provide both quantitative and qualitative experiments that demonstrate its superiority with respect the baselines in the tasks of missing data imputation and supervised learning, placing HH-VAEM as a robust model for real-world datasets. Further, we have developed a novel sampling-based technique for dynamic feature selection that outperforms the Gaussian-based alternatives and results in an efficient method for active learning in deep generative models.

%% file: appendix.tex
\newpage
\section{Experimental details}

\subsection{Experimental setup} \label{sec:exp_setup}
The networks for the encoder of the model with the MNIST datasets are 2 layered Deep CNNs with \{16, 32, 32\} output channels, kernel size 5, stride 2 and padding 2. For the experiments with CelebA, we use 5 layered Deep CNNs with \{32, 32, 64, 64, 512\} output channels, kernel size 4, stride 2 and padding 1, followed by batch norm layers. They are followed by MLPs with 512 hidden units for obtaining the variational parameters for each layer. The decoder that obtains $p_{\theta}(\x | \h_1)$ is the symmetric CNN.
All the NNs employed in the models trained with UCI datasets are one single layer MLPs with 256 hidden units. The noise variance for Gaussian likelihoods is set up to $0.1$. 

We employ learning rates of $1\times 10^{-3}$ for the models with MLP networks and $2\times 10^{-4}$ for the convolutional models. For the inflation parameter $\bm{s}$, we increase to $1\times 10^{-2}$ for a faster convergence. A batch size of $100$ is used for all the models except for Yatch and Wine dataset, where we use $20$ samples per batch. The number of training steps is $2\times 10^{4}$ for Boston, Energy, Wine, Yatch, Concrete, Diabetes and Yatch, and $5\times 10^{4}$ for Naval, Avocado, Bank, and Insurance. For MNIST and Fashion-MNIST, we have $1\times 10^{5}$ training steps. For CelebA, we use $1,5\times 10^{5}$ training steps. For the marginal VAEs stage, we employ $1\times 10^{3}$ training steps. The dimension of the latent variables is $\left[d_1=10 , d_2=5\right]$ for Boston, Energy, Wine, Naval Avocado, Bank and Insurance $\left[d_1=4 , d_2=2\right]$ for Concrete, Yatch and Diabetes, $\left[d_1= 20 , d_2=10\right]$ for MNIST and Fashion-MNIST, and $\left[d_1= 32 , d_2=16\right]$ for CelebA. 

We use $LF=5$ Leapfrog steps in all cases, chains of $T=10$ for Boston, Energy, Wine, Naval Avocado, Bank, Insurance, MNIST, Fashion-MNIST and CelebA, and $T=5$ for Concrete, Yatch and Diabetes. The SKSD function is estimated using $30$ HMC samples.

For the MNIST and CelebA datasets, the use of Nvidia P100 GPU with Pascal architecture sped up the training with the CNN-based architecture. For the UCI datasets, due to the use of small networks, the differences when using CPU or GPU are almost imperceptible.

\subsection{Datasets information}
Apart from the MNIST, Fashion-MNIST and CelebA datasets, a total of 10 UCI datasets have been employed in this work including: Bank Marketing, Insurance Company Benchmark, Avocado sales, Naval Propulsion Plants, Yatch Hydrodynamics, Diabetes,  Boston Housing,  Wines,  Energy efficiency and Bank Marketing.

\subsection{Balancing the KLs}
Following \citep{vahdat2020nvae}, we define a short initial warming stage (10\% of the total training steps) during the optimization where the KLs for the different layers are balanced according to their magnitude and the corresponding latent dimension, preventing the model for falling into posterior collapse by ignoring deepest layers. A factor is applied to each KL, following
\begin{equation}
    \gamma_l = \frac{ d_l \;\mathbb{E}_{x\sim B}\left[ \text{KL}(q(\bm{\epsilon}_l | \bm{x}) || p(\bm{\epsilon}) ) \right] }{ \sum_{i=1}^{L} d_i \;\mathbb{E}_{x\sim B}\left[ \text{KL}(q(\bm{\epsilon}_i | \bm{x}) || p(\bm{\epsilon}) ) \right]}.
\end{equation}
The factors penalises the fact that a layer might be ignored by making the KL smaller when its magnitude is small compared to the rest layers.

\section{Hamiltonian Monte Carlo with automatic optimization}

\subsection{Efficacy of training HMC} \label{sec:app_hmc}
\begin{figure*}[htbp]
	\centering
	\subfigure[Before training (wave)]{\includegraphics[width=0.32\linewidth]{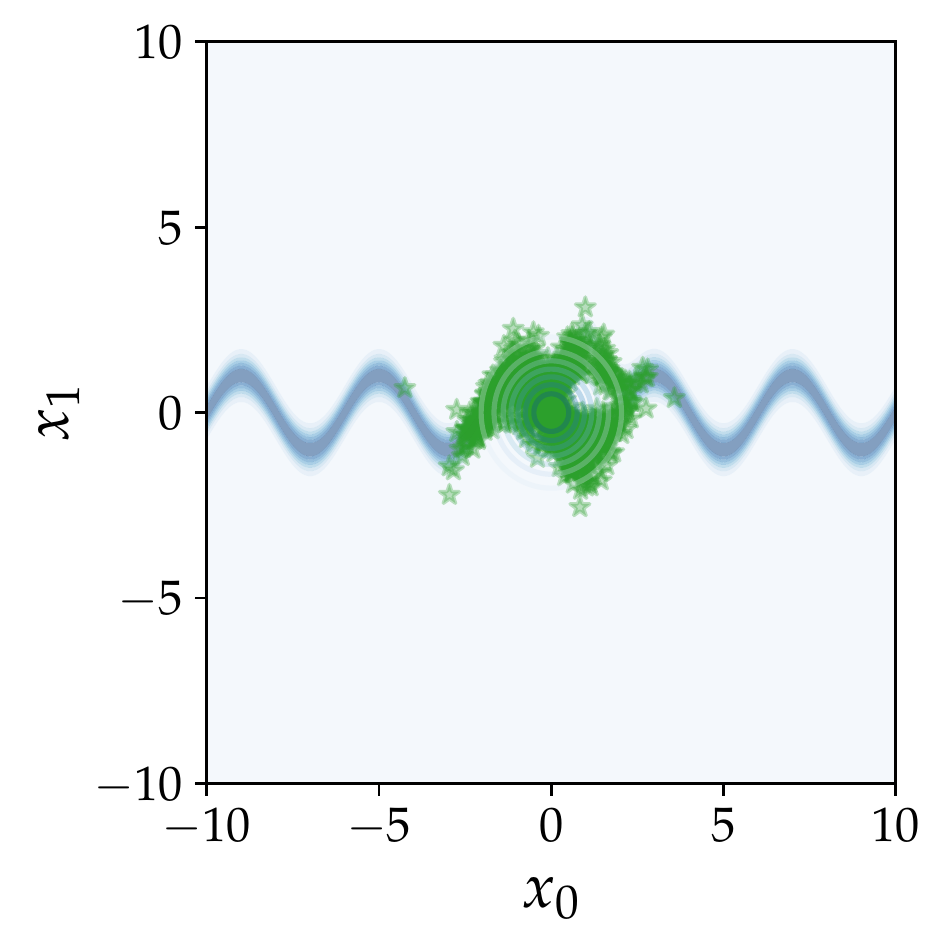}}
	\subfigure[After training (wave)]{\includegraphics[width=0.32\linewidth]{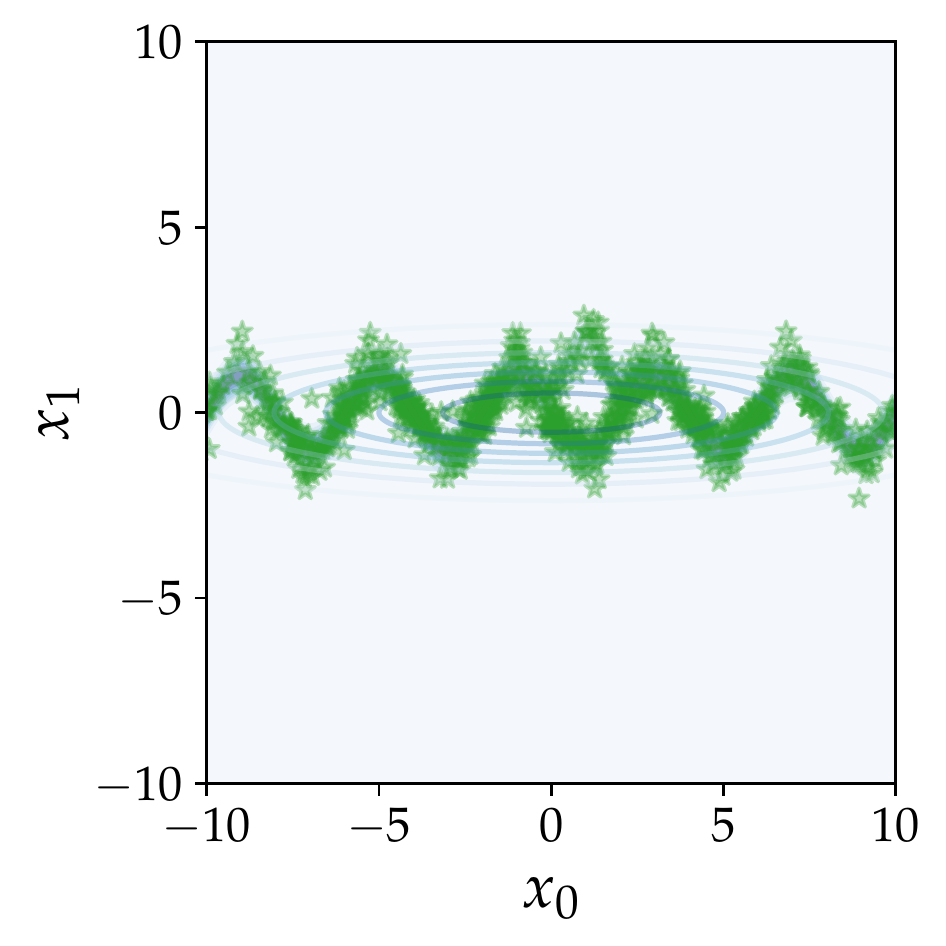}}
	\subfigure[Objectives, inflation (wave) ]{\includegraphics[width=0.32\linewidth]{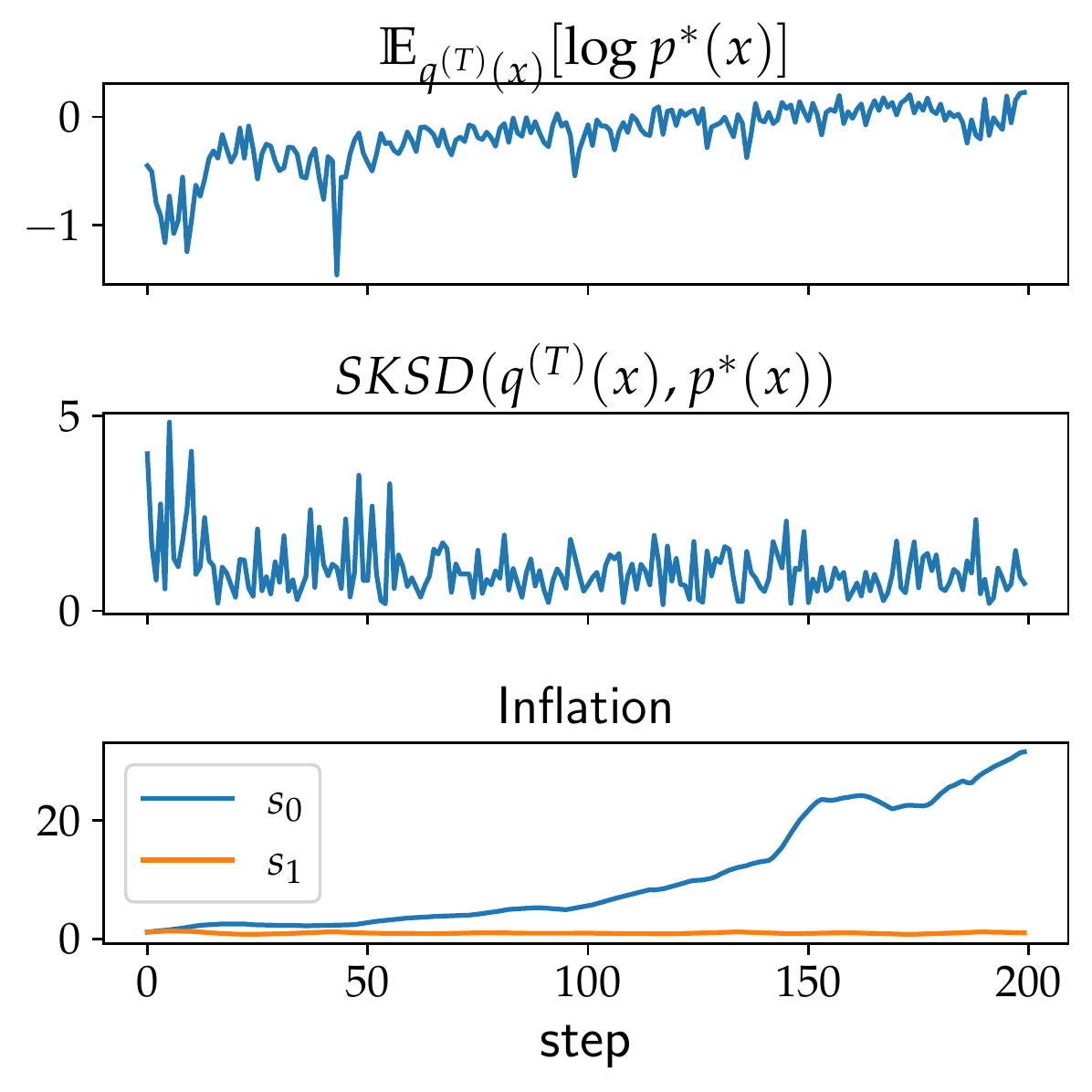}}
	\subfigure[Before training (dual moon)]{\includegraphics[width=0.32\linewidth]{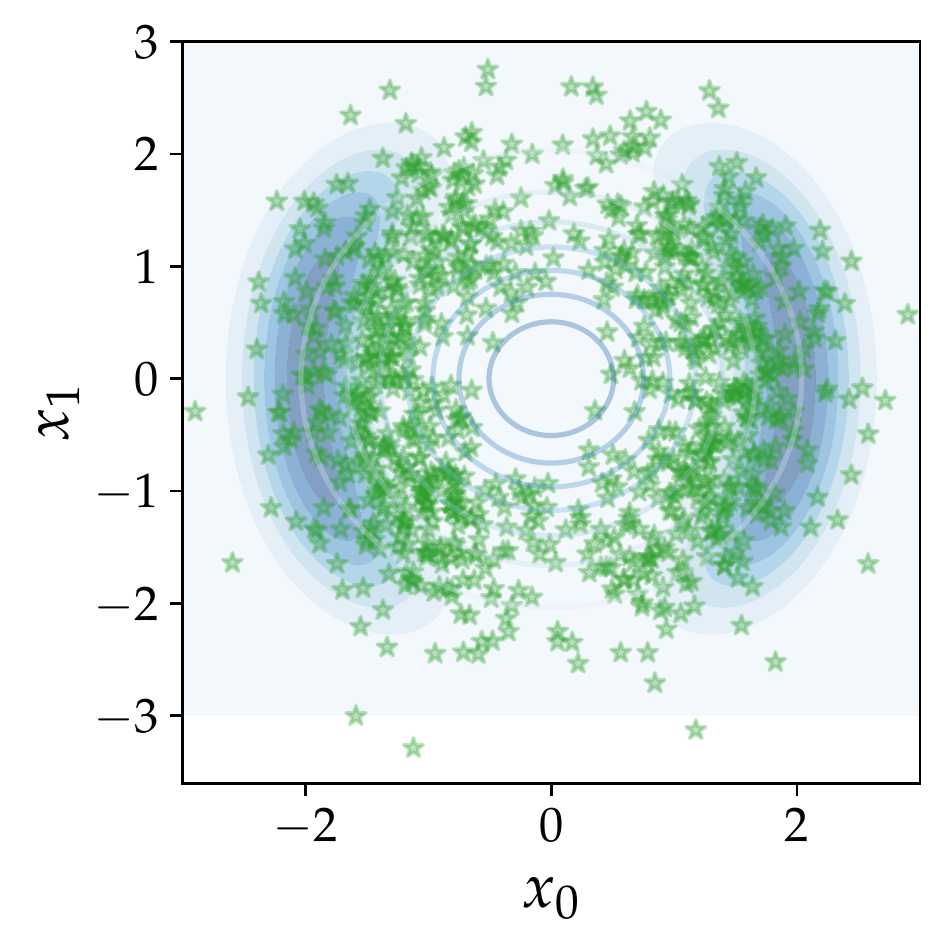}}
	\subfigure[After training (dual moon)]{\includegraphics[width=0.32\linewidth]{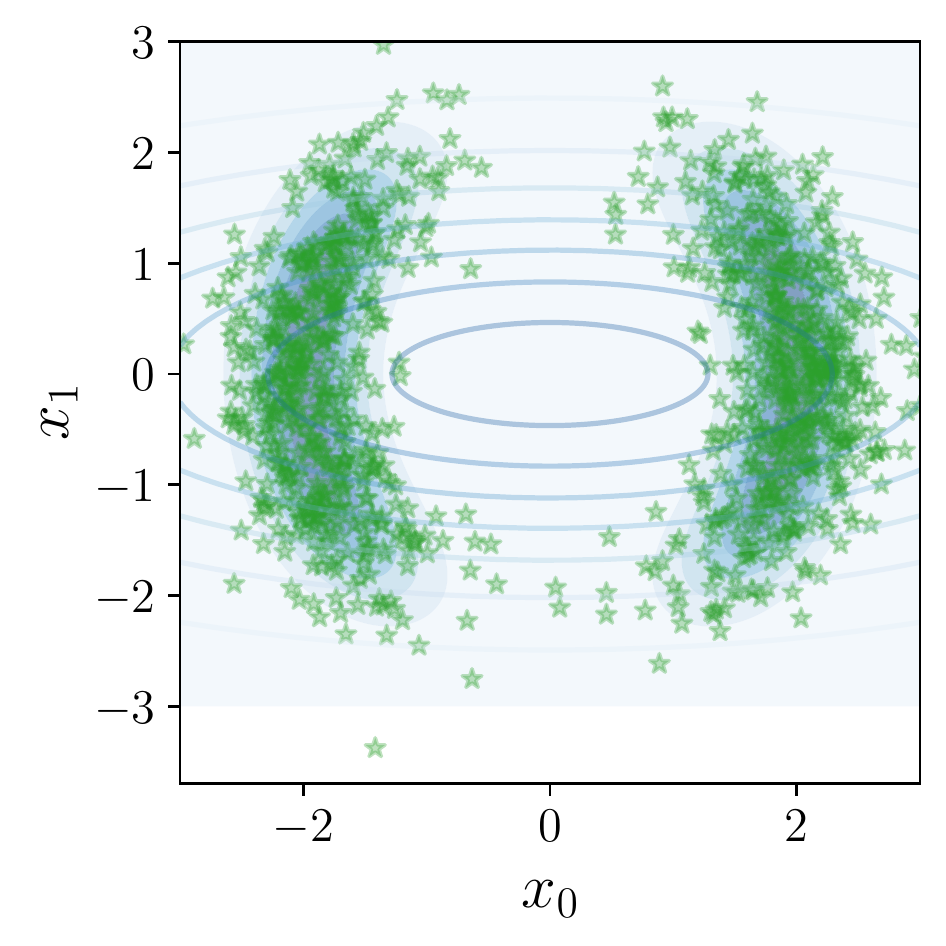}}
	\subfigure[Objectives, inflation (dual moon) ]{\includegraphics[width=0.32\linewidth]{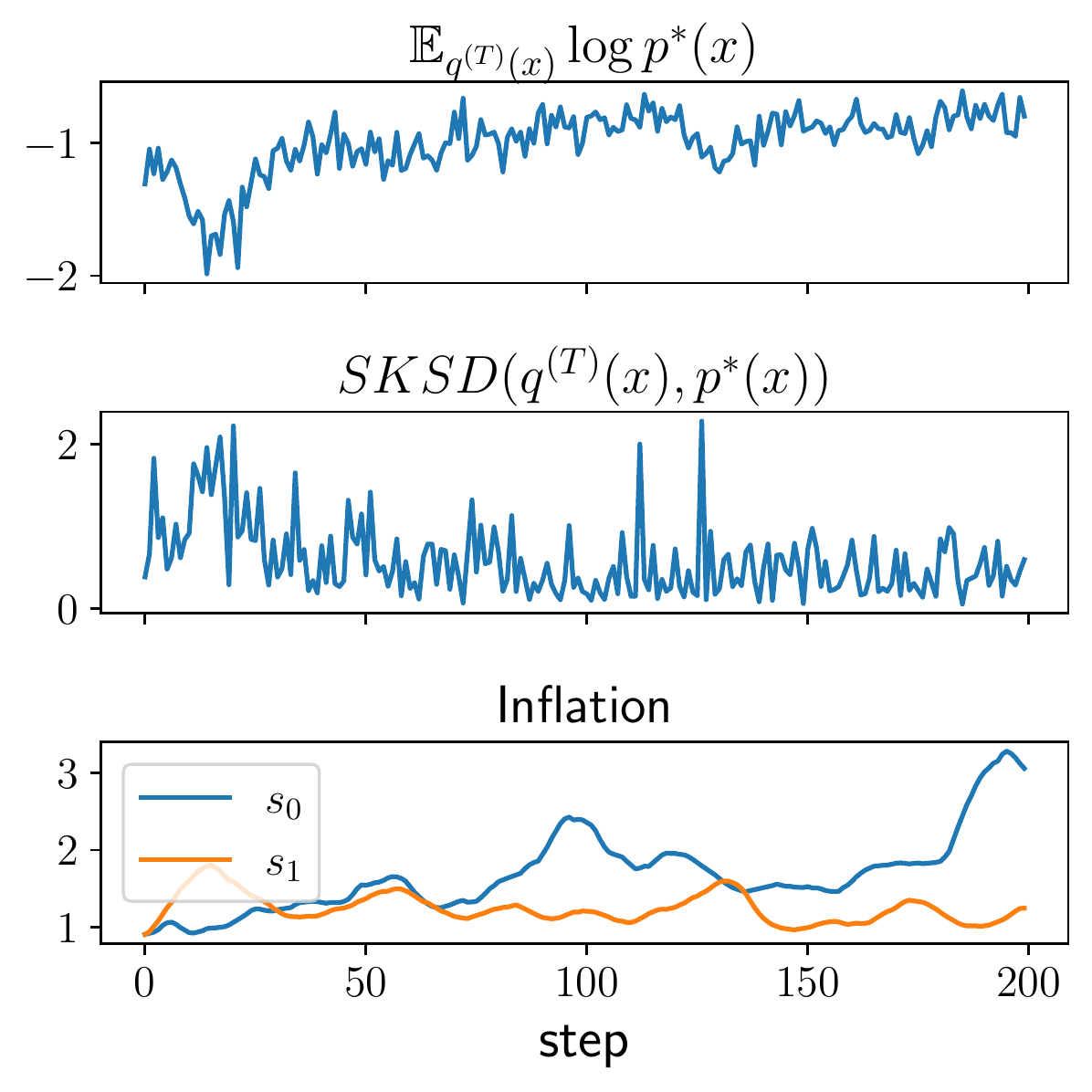}}
	\caption{Toy example showing the efficacy of training HMC hyperpameters using our method on two densities: wave (top row) with $T=5$ and dual moon (bottom row) with $T=10$. Left column illustrates the non desired behavior when chains hardly explore the density and stuck in a small region close to the mass of the tight Gaussian initial proposal (light blue contour ellipses). Right column shows how optimizing the step sizes and the inflation parameter leads to a vast improvement of the exploration. More specifically, (b) justifies scaling each dimension of the target, since the inflation is bigger on the horizontal axis.} \label{fig:hmc_efficacy}
\end{figure*}

We include in this experiment results that demonstrate the efficacy of training the HMC hyperparameters using our proposed gradient-based strategy on 2D densities. In Figure \ref{fig:hmc_efficacy}, each row correspond to a different example: first row is a \textit{wave} density,  second row is a \textit{dual moon} density. In the first column, the initial set up is showed, including the density contour (dark blue), the initial Gaussian proposal contour (light blue) and samples from HMC (green). In both cases, due to the tightness of the proposal, chains do not properly explore the density and get stuck close the initial state. In the second column, results after training HMC hyperparameters and the inflation parameter are included. Again, in both cases, and more specially in the first one, the inflation of the horizontal variance of the proposal successfully increases to better cover the surface, and final HMC samples vastly improve the exploration. In the third column, the approximations of the HMC objective, the SKSD discrepancy and the inflation parameters over the optimization steps are included.

\subsection{HMC optimization} \label{sec:app_hmc_convergence}

\begin{figure*}[htbp]
	\centering
	\subfigure[$\log p(\xu | \xo) $]{\includegraphics[width=0.32\linewidth]{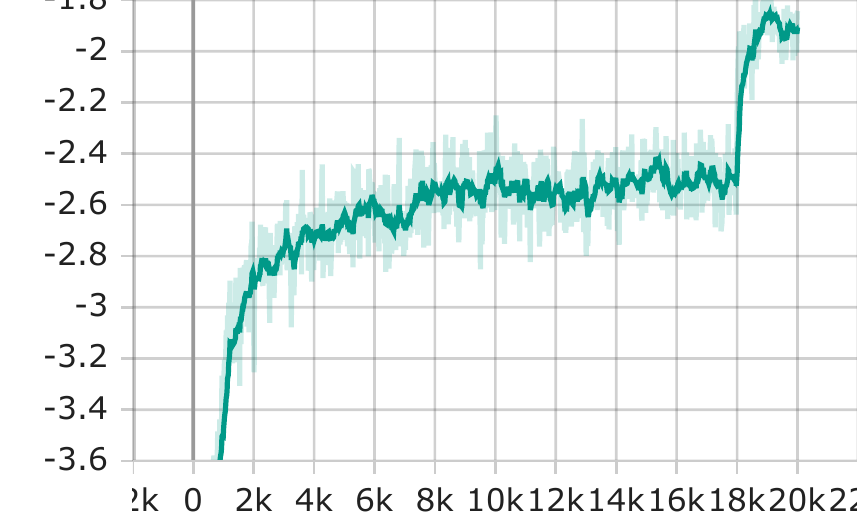}}
	\subfigure[$\log p(\y | \xo) $]{\includegraphics[width=0.32\linewidth]{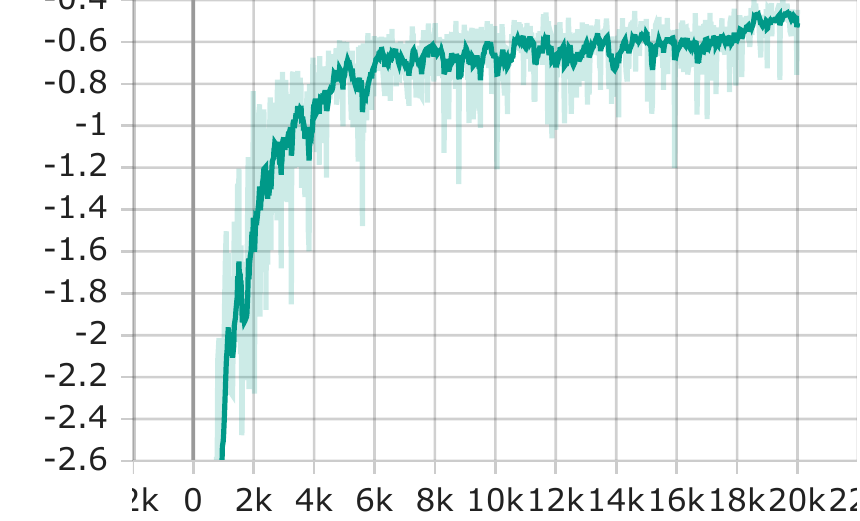}}
	\subfigure[RMSE]{\includegraphics[width=0.32\linewidth]{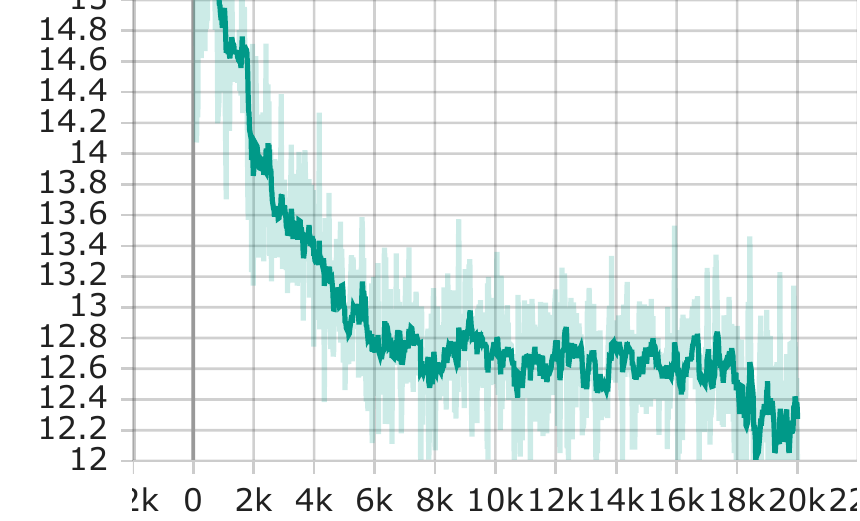}}
	\caption{Validation curves during optimization for Naval dataset. } \label{fig:val_curves}
\end{figure*}

We face the computational cost of running HMC by defining a small percentage of training steps for the last stage in Algorithm \ref{alg:training}. A  10\% of the total training steps for $T_{HMC}$ is sufficient for obtaining the convergence. In Figure \ref{fig:val_curves} we include the validation metrics obtained during the optimization of Yatch dataset, where $T_{HMC}=2\times 10 ^3$. 

\begin{figure*}[htbp]
	\centering
	\subfigure[Mean acceptance rate $\bar{p}_a$ ]{\includegraphics[width=0.45\linewidth]{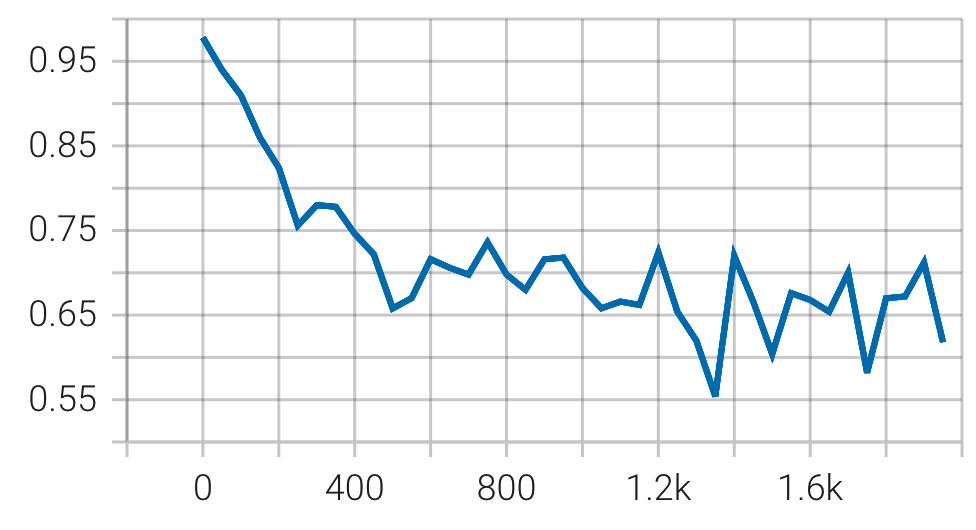}}
	\subfigure[Step sizes]{\includegraphics[width=0.45\linewidth]{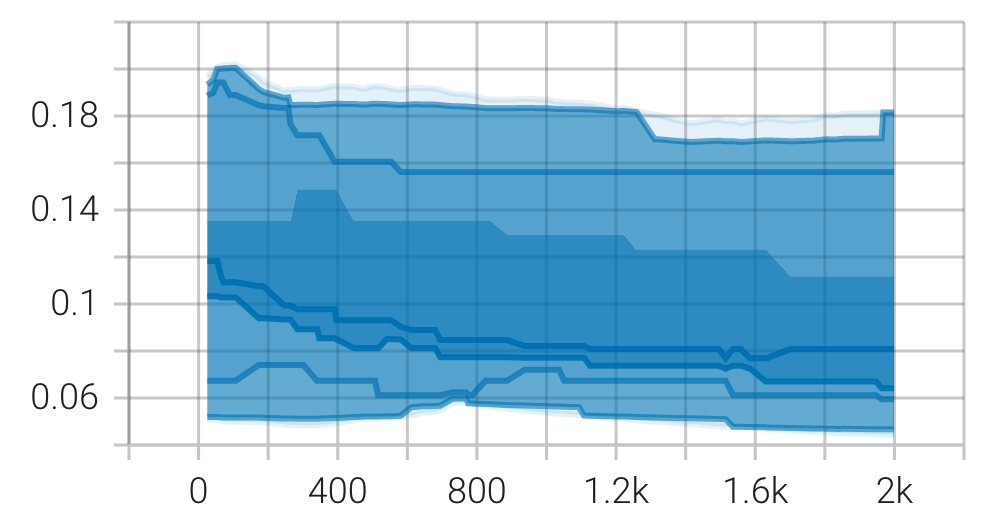}}
	\caption{Evolution of mean acceptance rate $\bar{p}_a$ (a) and step sizes (b) over the training steps. } \label{fig:hmc_accpt}
\end{figure*}

In figure \ref{fig:hmc_accpt} (a), the mean acceptance rate of the HMC sampler over the training steps and the step sizes (b) are included. The steps are initialized from $U(0.05, 0.2)$. After $2\times10^3$ steps, the mean acceptance rate converges to a value closer to $\bar{p}_a = 0.65$, which is defined as the optimal desired acceptance probability \citep{neal2011mcmc}. This empirical result provides evidence that, apart from reducing the computational cost, reducing the HMC training step to this value is sufficient for achieving convergence.

\subsection{Reparameterization trick for solving ill-posed HMC with hierarchical densities} \label{sec:app_reparam}

We provide in this section strong empirical demonstration on the efficacy of our proposed reparameterization trick. As stated in Section \ref{sec:hier_reparam}, naïve implementations of HMC are ill-posed when combined with hierarchical densities. The autoregressive correlations lead to non-smooth densities with huge peaks. The large gradients $\nabla_{\bm{h}_{1:L}} \log p^*(\bm{h}_{1:L})$ evaluated on these regions inside the Leapfrog steps of Equations \ref{eq:leapfrog} make huge modifications of the proposed states. If we denote these diverged states by $\bm{h}_{1:L}^{(ill)}$, evaluating the objective $\log p^*(\bm{h}_{1:L}^{(ill)})$ lead to overflow issues. This undesired behavior is what we call \textit{divergence of the Leapfrog integrator}, and is also illustrated by \cite{betancourt2017conceptual} in Figure 35.

\begin{figure*}[h]
	\centering
	\subfigure[Mean acceptance rate $\bar{p}_a$]{\includegraphics[width=0.49\linewidth]{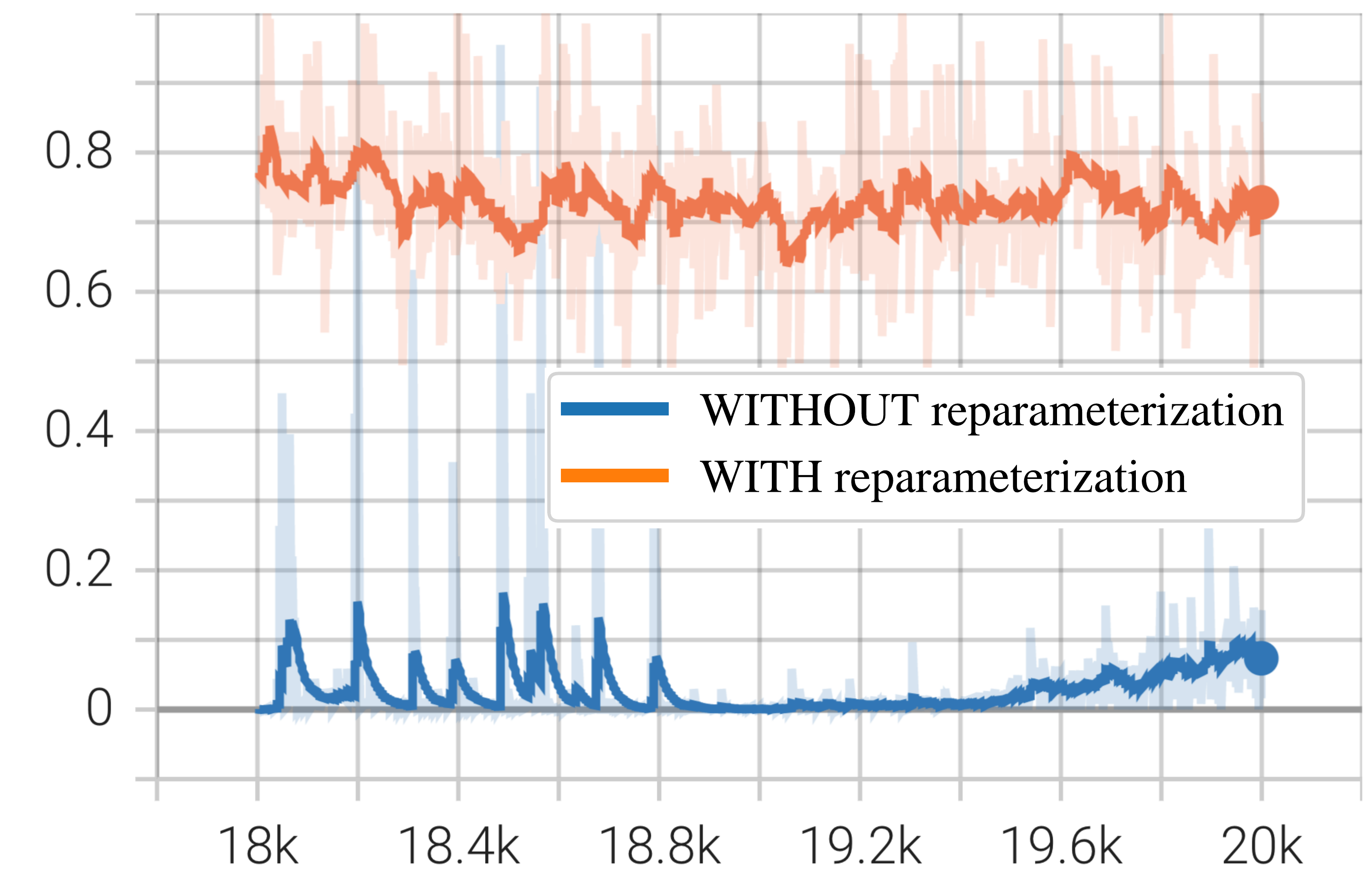}}
	\subfigure[ $\log p(\xu | \xo)$]{\includegraphics[width=0.49\linewidth]{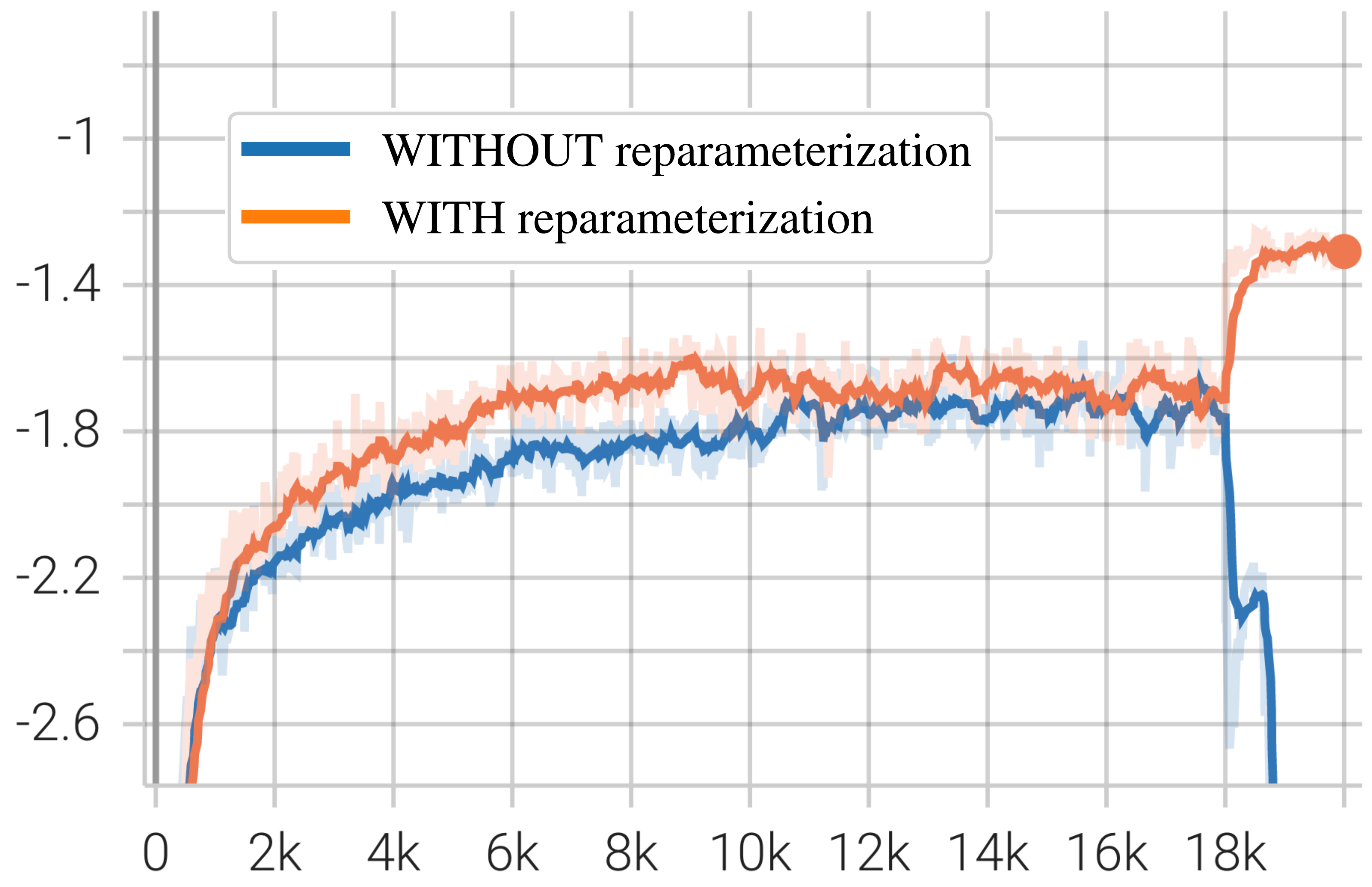}}
	\caption{Demonstration of the efficacy of our reparameterization trick. Our model is showed with orange lines, and the same model without the reparameterization with blue lines. In (a), the mean acceptance rate of the HMC proposals over the optimization steps is included. We demonstrate that without the reparameterization, HMC is ill-posed and by rejecting the proposals that make the integrator diverge, the acceptance rate is extremely low, thus not properly exploring the density. In (b), the imputation log likelihood metric is showed for the whole optimization. With the reparameterization trick, we successfully solve the pathological issues, leading to a considerable increase of the metric. }
	\label{fig:hmc_reparam}
\end{figure*}

In order to avoid the aforementioned overflow issues, we can directly reject these problematic states. Nevertheless, by doing this, HMC will not properly explore the density by the time the states fall into the problematic regions, leading to extremely low acceptance rates. To demonstrate this, we include in Figure \ref{fig:hmc_reparam} (a) with blue line the evolution of the acceptance rates when optimizing HMC with the HH-VAEM variant without reparameterization (as illustrated in Figure \ref{fig:reparam} (a)). The acceptance rate is extremely low as expected, which is a clear indicator of a poorly mixing sampler.

On the contrary, by using our proposed reparameterization trick (Figure \ref{fig:reparam} (b)), we are able to make HMC work properly and tune the step sizes, getting closer to the ideal acceptance rate $\bar{p}_a=0.65$ \citep{neal2011mcmc}. Evaluations are applied to the same validation split of the Boston dataset. The step sizes of HMC are initialized from $U(0.05, 0.2)$ in both cases. 

Further, we also include in Figure \ref{fig:hmc_reparam} (b) the evolution of the imputation log likelihood metric of Equation \eqref{eq:imputation_likelihood} during the whole optimization for both approaches. First $1,8\times 10^4$ steps correspond to the pretraining stage using only the ELBO. When introducing HMC without the reparameterization, due to the low acceptance rate, the states hardly move from the initial proposal, or move away from the density, and the joint optimization fail. We demonstrate again the effectiveness of the reparameterization trick by observing an increase in this metric. 

\section{Extended experiments}
\subsection{Deterministic imputation metrics}\label{sec:app_error}
We include in Table \ref{tab:rmse_xu} results on the RMSE obtained with other discriminative validated predictors, using mean imputation under the same missing rates. Additionally, we include here the missForest in the baselines, a wide-spread method for missing data imputation using a Random Forest approach \citep{stekhoven2012missforest}. For classification tasks, the error rate is considered. In almost all cases, HH-VAEM outperforms the baselines.

\begin{table}[h] 

\setlength{\tabcolsep}{3pt}

    \centering
    \resizebox{\linewidth}{!}{
    \begin{tabular}{@{}rrrrrrrrrrr@{}}
        \toprule
         &
          Bank &
          Insurance &
          Avocado &
          Naval &
          Yatch &
          Diabetes &
          Concrete &
          Wine &
          Energy &
          Boston \\ \midrule
        missForest &
          $0.64 \pm 0.01$ &
          $0.61 \pm 0.06$ &
          $0.59 \pm 0.02$ &
          $0.30 \pm 0.01$ &
          $\bm{0.86 \pm 0.12}$ &
          $0.76 \pm 0.08$ &
          $0.76 \pm 0.07$ &
          $0.77 \pm 0.11$ &
          $0.64 \pm 0.08$ &
          $0.59 \pm 0.06$ \\
        VAEM &
          $0.57 \pm 0.01$ &
          $0.40 \pm 0.01$ &
          $0.59 \pm 0.00$ &
          $0.33 \pm 0.01$ &
          $0.93 \pm 0.04$ &
          $0.79 \pm 0.02$ &
          $0.74 \pm 0.04$ &
          $0.64 \pm 0.03$ &
          $0.75 \pm 0.02$ &
          $0.65\pm 0.02$ \\
        MIWAEM &
          $0.56 \pm 0.00$ &
          $0.39 \pm 0.00$ &
          $0.59 \pm 0.01$ &
          $0.34 \pm 0.01$ &
          $0.94 \pm 0.04$ &
          $0.75\pm 0.01$ &
          $0.71 \pm 0.03$ &
          $0.63 \pm 0.02$ &
          $0.75\pm 0.02$ &
          $0.63 \pm 0.01$ \\
        H-VAEM &
          $0.56 \pm 0.01$ &
          $0.39 \pm 0.00$ &
          $0.58 \pm 0.01$ &
          $0.32 \pm 0.01$ &
          $0.92 \pm 0.05$ &
          $0.77\pm 0.01$ &
          $0.71 \pm 0.02$ &
          $0.60 \pm 0.04$ &
          $0.55 \pm 0.03$ &
          $0.59 \pm 0.01$ \\
        HMC-VAE &
          $0.55 \pm 0.01$ &
          $\bm{0.38 \pm 0.00}$ &
          $0.58 \pm 0.01$ &
          $0.30 \pm 0.02$ &
          $0.91\pm 0.05$ &
          $0.76 \pm 0.03$ &
          $0.71 \pm 0.02$ &
          $0.60 \pm 0.02$ &
          $0.55 \pm 0.02$ &
          $0.57 \pm 0.02$ \\
        HH-VAEM &
          $\bm{0.54 \pm 0.01}$ &
          $\bm{0.38 \pm 0.00}$ &
          $\bm{0.57 \pm 0.00}$ &
          $\bm{0.29 \pm 0.02}$ &
          $0.90 \pm 0.00$ &
          $\bm{0.75 \pm 0.01}$ &
          $\bm{0.70 \pm 0.01}$ &
          $\bm{0.59 \pm 0.04}$ &
          $\bm{0.54\pm 0.01}$ &
          $\bm{0.56 \pm 0.03}$ \\ \bottomrule
        \end{tabular}
    }
    \vspace{0.1cm}
    \caption{Test RMSE of the unobserved features for our model and baselines.}
    \label{tab:rmse_xu}
\end{table}

\subsection{Likelihood of the observed features}

We include in this section results on the negative log likelihood of the observed features
\begin{equation}
    \log p(\xo) \approx  \log \mathbb{E}_{\beps \sim q^{(T)}(\beps | \xo) } \left[  p(\xo | \beps) \right] \approx   \log \frac{1}{k} \sum_i^k p(\xo | \beps_i)  .
\end{equation}
Results are included in Table \ref{tab:nll_xo}. In almost all the cases, we confirm the incremental superiority when adding each part of our proposed design.

\begin{table}[h] 

    \setlength{\tabcolsep}{3pt}

    \centering
    \resizebox{\linewidth}{!}{
    \begin{tabular}{@{}rrrrrrrrrrr@{}}
        \toprule
         &
          bank &
          insurance &
          avocado &
          naval &
          yatch &
          diabetes &
          concrete &
          wine &
          energy &
          boston \\ \midrule
            VAEM &
              $0.51 \pm 0.05$ &
              $0.99 \pm 0.05$ &
              $0.44 \pm 0.01$ &
              $0.21 \pm 0.01$ &
              $0.62 \pm 0.13$ &
              $0.92 \pm 0.12$ &
              $0.63 \pm 0.18$ &
              $0.73 \pm 0.18$ &
              $1.86 \pm 0.09$ &
              $0.56 \pm 0.11$ \\
            MIWAEM &
              $0.63 \pm 0.02$ &
              $1.06 \pm 0.03$ &
              $0.60 \pm 0.03$ &
              $0.33 \pm 0.01$ &
              $0.75 \pm 0.07$ &
              $1.05 \pm 0.06$ &
              $0.76 \pm 0.09$ &
              $0.80 \pm 0.06$ &
              $1.77 \pm 0.15$ &
              $0.67 \pm 0.03$ \\
            H-VAEM &
              $0.40 \pm 0.04$ &
              $0.93 \pm 0.04$ &
              $0.42 \pm 0.05$ &
              $0.19 \pm 0.07$ &
              $0.58 \pm 0.09$ &
              $0.70 \pm 0.13$ &
              $0.53 \pm 0.18$ &
              $0.71 \pm 0.15$ &
              $0.38 \pm 0.02$ &
              $0.49 \pm 0.07$ \\
            HMC-VAE &
              $0.37 \pm 0.07$ &
              $\bm{0.92 \pm 0.04}$ &
              $0.39 \pm 0.06$ &
              $0.18 \pm 0.05$ &
              $0.54 \pm 0.10$ &
              $\bm{0.68 \pm 0.07}$ &
              $0.49 \pm 0.22$ &
              $\bm{0.55 \pm 0.07}$ &
              $0.40 \pm 0.06$ &
              $0.41 \pm 0.04$ \\
            HH-VAEM &
              $\bm{0.33 \pm 0.03}$ &
              $0.95 \pm 0.05$ &
              $\bm{0.36 \pm 0.01}$ &
              $\bm{0.17 \pm 0.04}$ &
              $\bm{0.45\pm 0.04}$ &
              $0.68 \pm 0.16$ &
              $\bm{0.40 \pm 0.16}$ &
              $0.64 \pm 0.17$ &
              $\bm{0.37 \pm 0.06}$ &
              $\bm{0.41 \pm 0.04}$ \\ \bottomrule
        \end{tabular}
    }
    \vspace{0.1cm}
    \caption{Test NLL of the observed features for our model and baselines.}
    \label{tab:nll_xo}
\end{table}

\subsection{Heterogeneous likelihoods}\label{sec:app_heterog}

In experiment \ref{sec:exp11} we reported an average likelihood across heterogeneous variables. Although this quantity is not a valid joint likelihood probability, we employed this average to provide a fair comparison on models that have been trained on the same heterogeneous likelihoods. In this section, we show the comparison on averaging separately the three considered marginal likelihoods (Gaussian, Bernoulli and Categorical) for two of the biggest datasets considered on Tables \ref{tab:ll_gaussian_xu_d}-\ref{tab:ll_categorical_xu_d}. Again, we show the incremental superiority when adding the different design choices of our model.

\begin{minipage}{\textwidth}
		\begin{minipage}{0.32\textwidth}
			
			\centering
			\resizebox{\linewidth}{!}{
				\begin{tabular}{@{}rrr@{}}
                    \toprule
                             & Bank            & Avocado          \\ \midrule
                    VAEM     & $0.36 \pm 0.29$ & $0.26 \pm 0.08$  \\
                    MIWAEM   & $0.33 \pm 0.27$ & $0.32 \pm 0.06$ \\
                    H-VAEM   & $0.26 \pm 0.22$ & $0.29 \pm 0.07$  \\
                    HMC-VAEM & $0.25 \pm 0.21$ & $0.25 \pm 0.08$  \\
                    \textbf{HH-VAEM}  & $\bm{0.20 \pm 0.22}$ & $\bm{0.22 \pm 0.07}$ \\ \bottomrule
                \end{tabular}
			}
			\captionof{table}{Average test Gaussian NLL of the observed features.}
			\label{tab:ll_gaussian_xu_d}
		\end{minipage}
		\hfill
		\begin{minipage}{0.32\textwidth}
			
			\centering
			\resizebox{\linewidth}{!}{
				\begin{tabular}{@{}lll@{}}
                    \toprule
                            & Bank                 & Avocado              \\ \midrule
                    VAEM    & $0.13 \pm 0.00$ & $0.06 \pm 0.00$ \\
                    MIWAEM  & $0.15 \pm 0.00$ & $0.09 \pm 0.00$ \\
                    H-VAEM & $0.11 \pm 0.00$ & $0.07 \pm 0.00$ \\
                    HMC-VAEM & $0.08 \pm 0.00$ & $0.05 \pm 0.00$ \\
                    \textbf{HH-VAEM}  & $\bm{0.07 \pm 0.00}$ & $\bm{0.04 \pm 0.00}$ \\ \bottomrule
                    \end{tabular}
			}
			\captionof{table}{Average test Bernoulli NLL of the observed features.}
			\label{tab:ll_bernoulli_xu_d}
		\end{minipage}
		\hfill
		\begin{minipage}{0.32\textwidth}
			
			\centering
			\resizebox{\linewidth}{!}{
				\begin{tabular}{@{}rrr@{}}
                    \toprule
                     & Bank             & Avocado          \\ \midrule
                    VAEM    & $0.24 \pm 0.16$ & $0.33 \pm 0.00$ \\
                    MIWAEM  & $0.26 \pm 0.17$ & $0.36 \pm 0.00$ \\
                    H-VAEM & $0.23 \pm 0.16$ & $0.32 \pm 0.00$ \\
                    HMC-VAEM & $0.22 \pm 0.15$ & $\bm{0.30 \pm 0.00}$ \\
                    \textbf{HH-VAEM}  & $\bm{0.21 \pm 0.15}$ & $\bm{0.30 \pm 0.00}$ \\ \bottomrule
                \end{tabular}
			}
			\captionof{table}{Average test Cat. NLL of the observed features.}
			\label{tab:ll_categorical_xu_d}
		\end{minipage}
		\hfill
\end{minipage}

\subsection{SAIA log-likelihoods}

\begin{figure}[t]
	\centering
	\includegraphics[width=0.8\linewidth]{figs/exp2/legend} \\
	\subfigure[Avocado]{\includegraphics[height=0.19\linewidth]{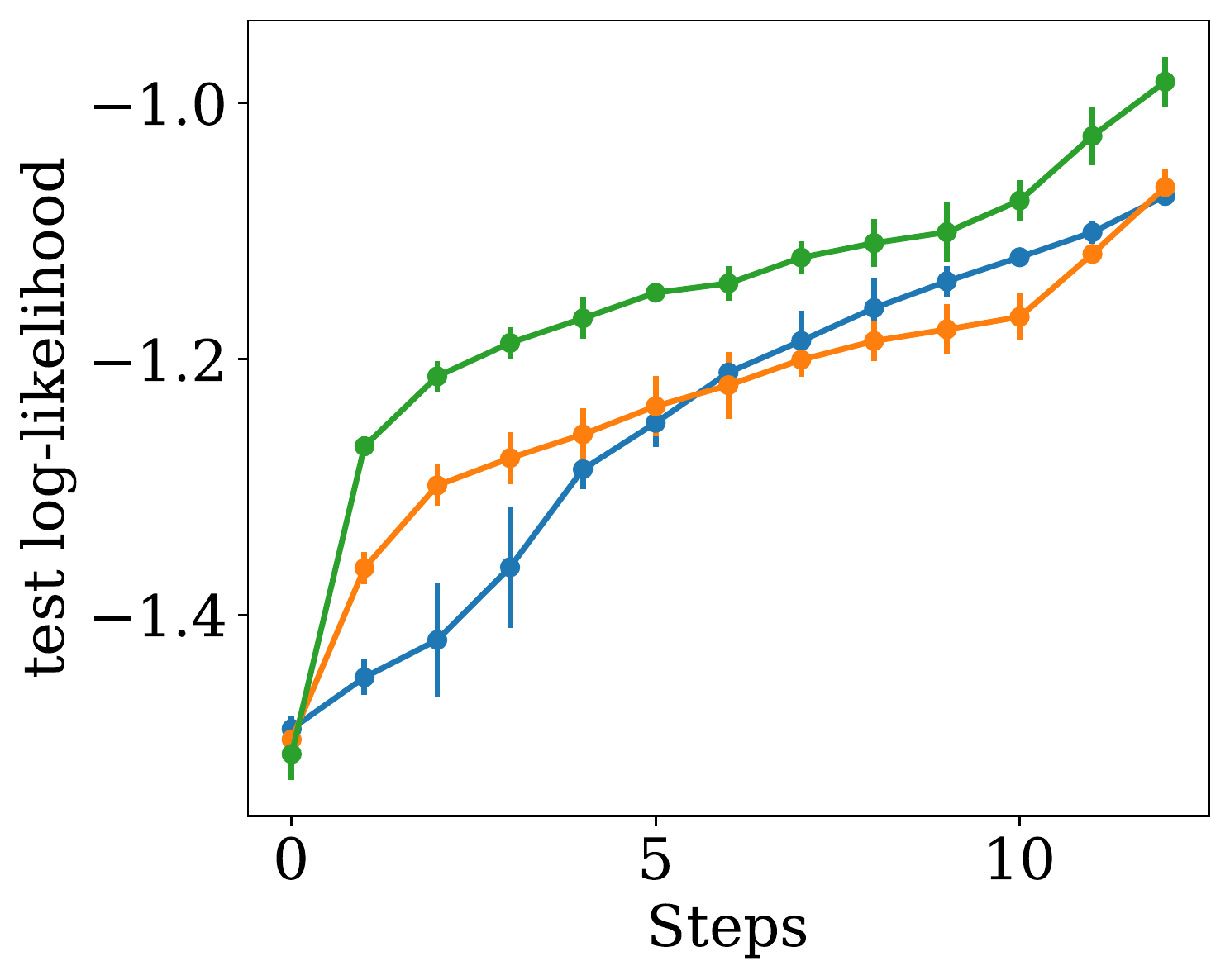}}
	\subfigure[Yatch]{\includegraphics[height=0.19\linewidth]{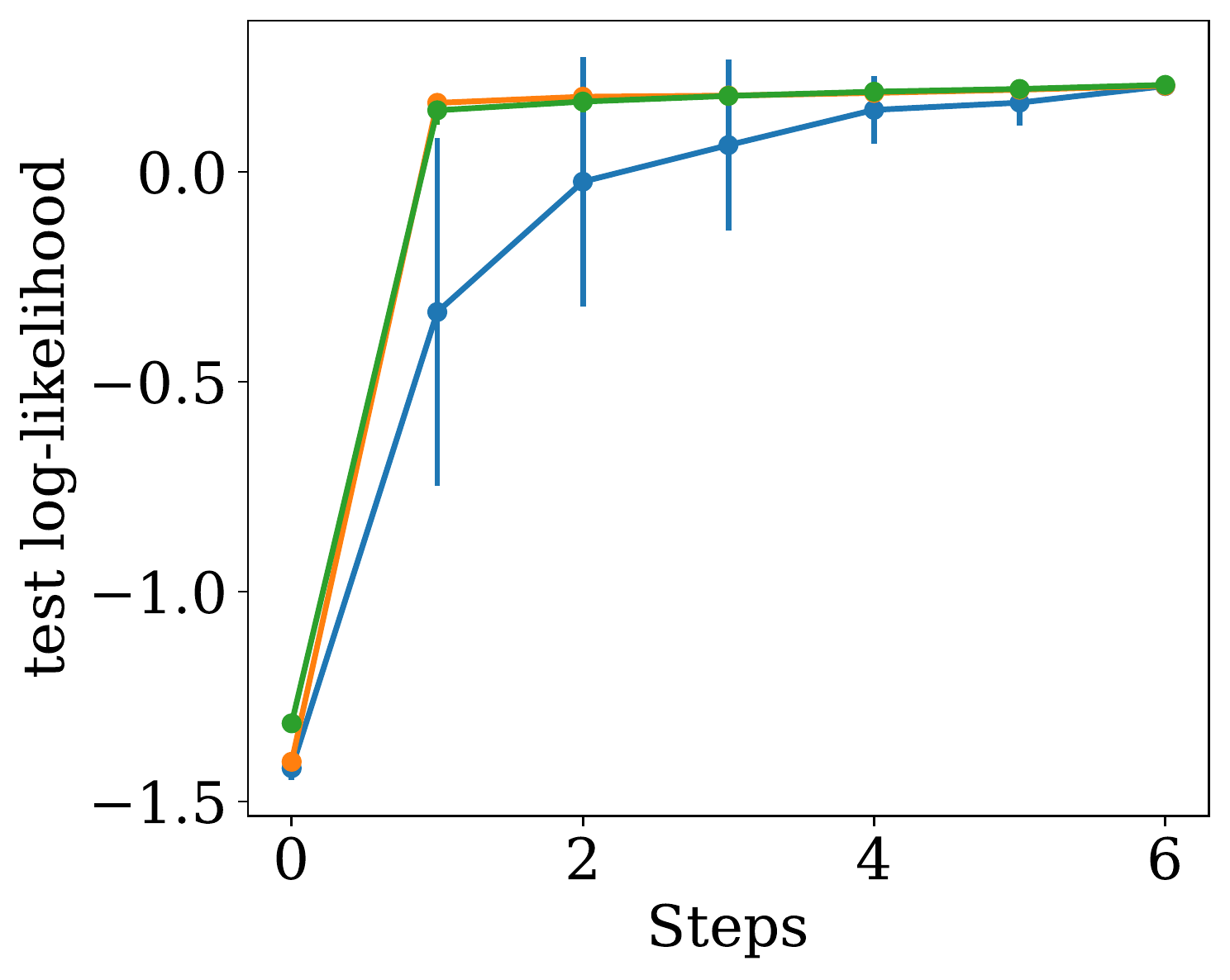}}
	\subfigure[Boston]{\includegraphics[height=0.19\linewidth]{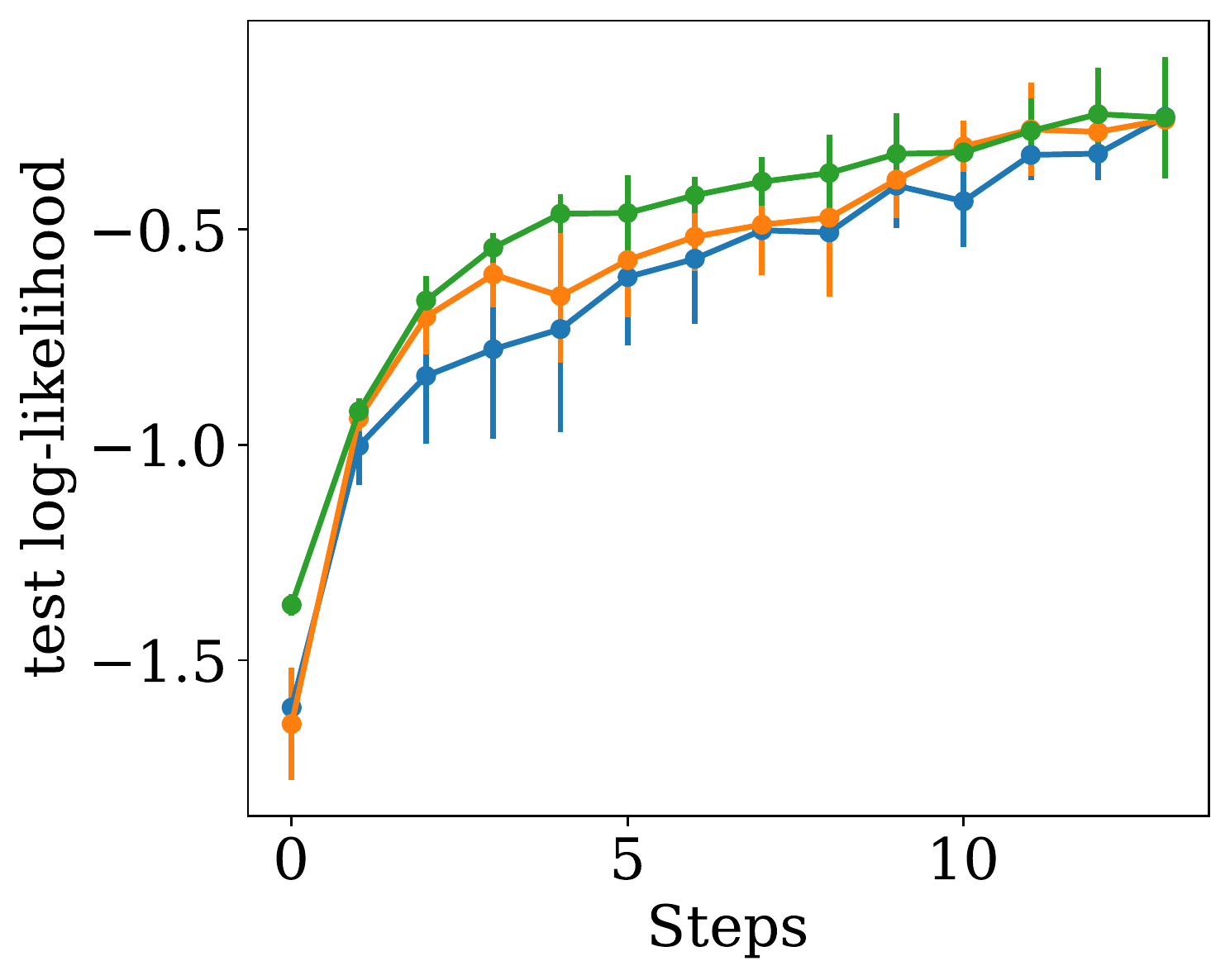}}
	\subfigure[Energy]{\includegraphics[height=0.19\linewidth]{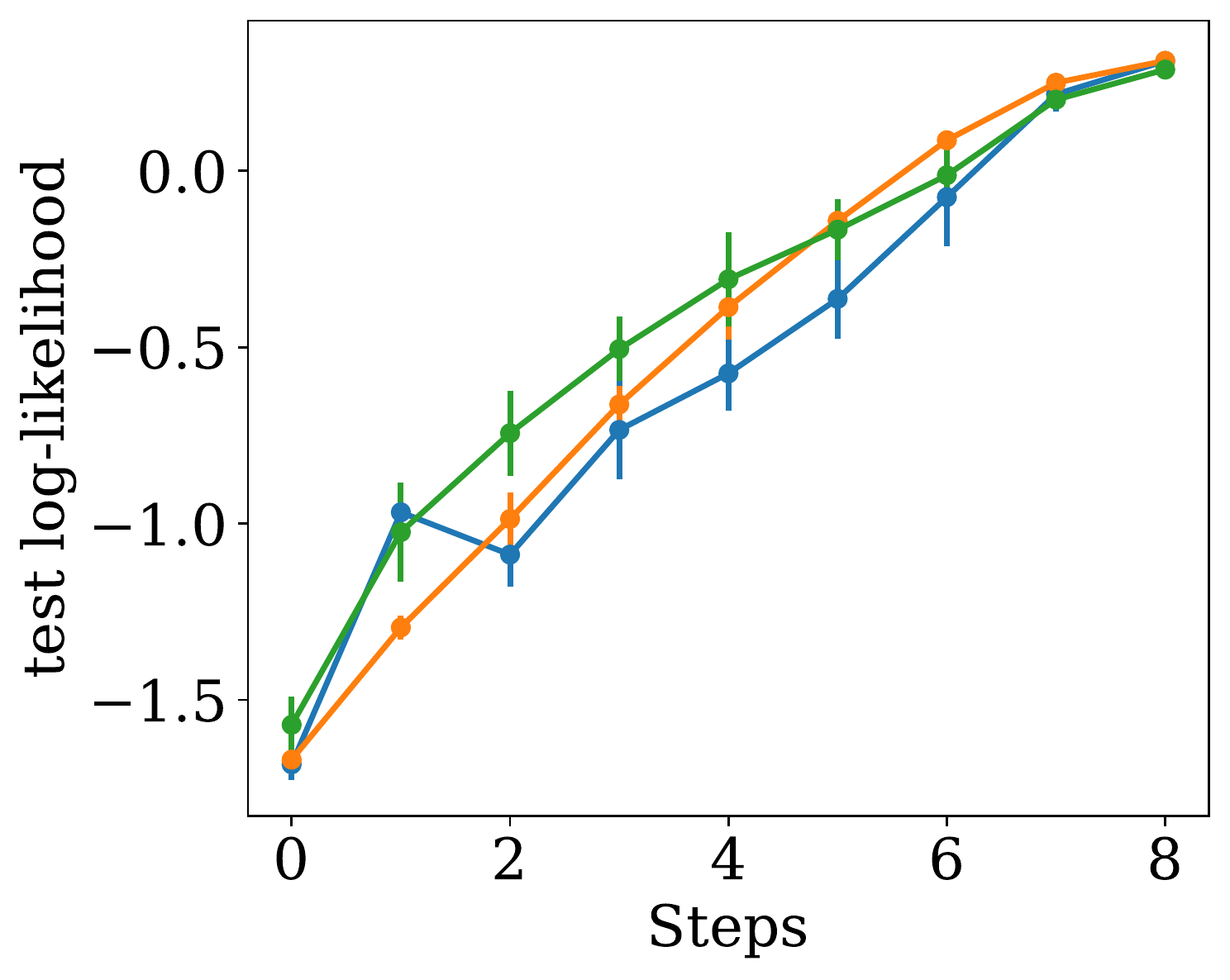}} 
	
	\subfigure[Wine]{\includegraphics[height=0.19\linewidth]{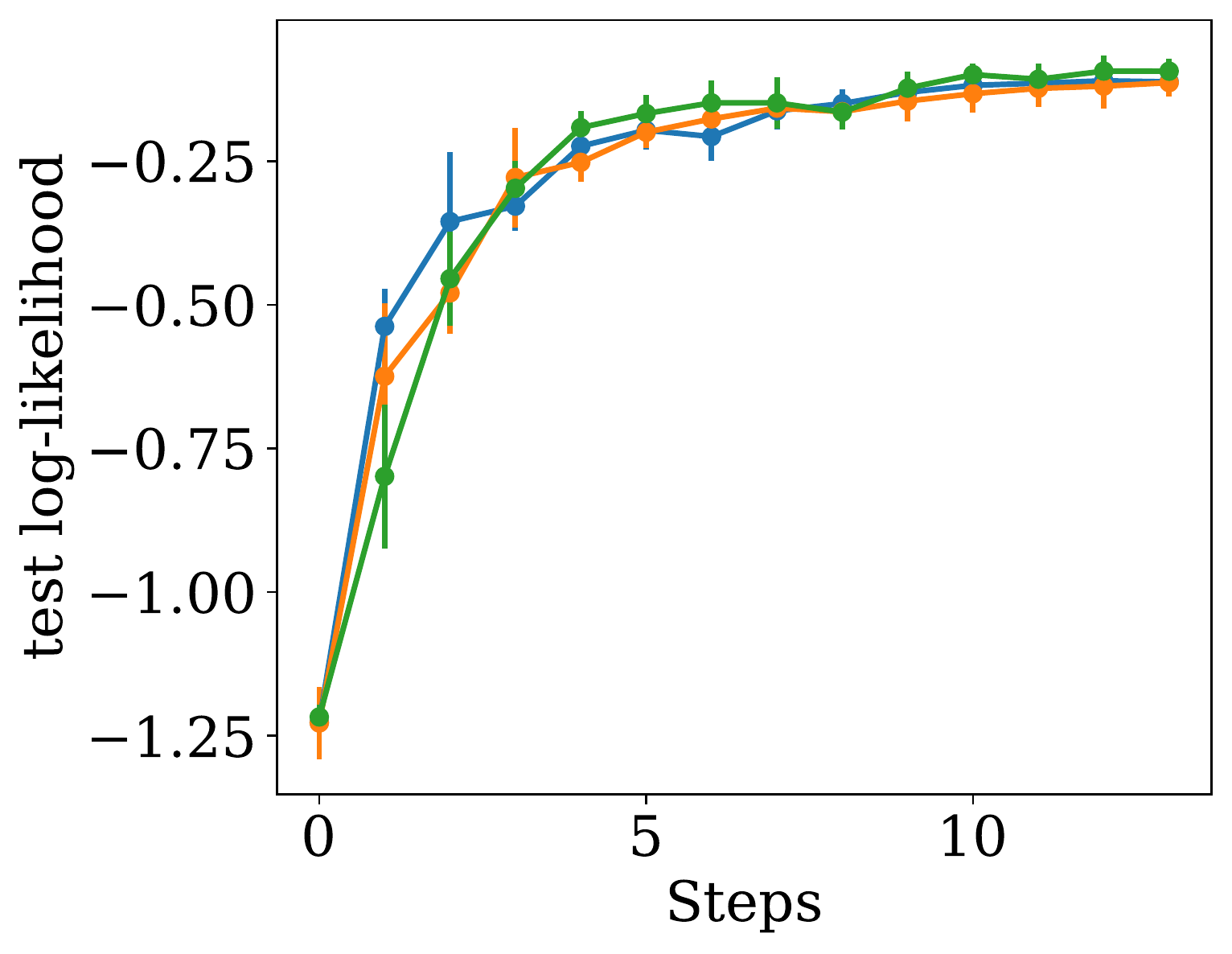}}
	\subfigure[Naval]{\includegraphics[height=0.19\linewidth]{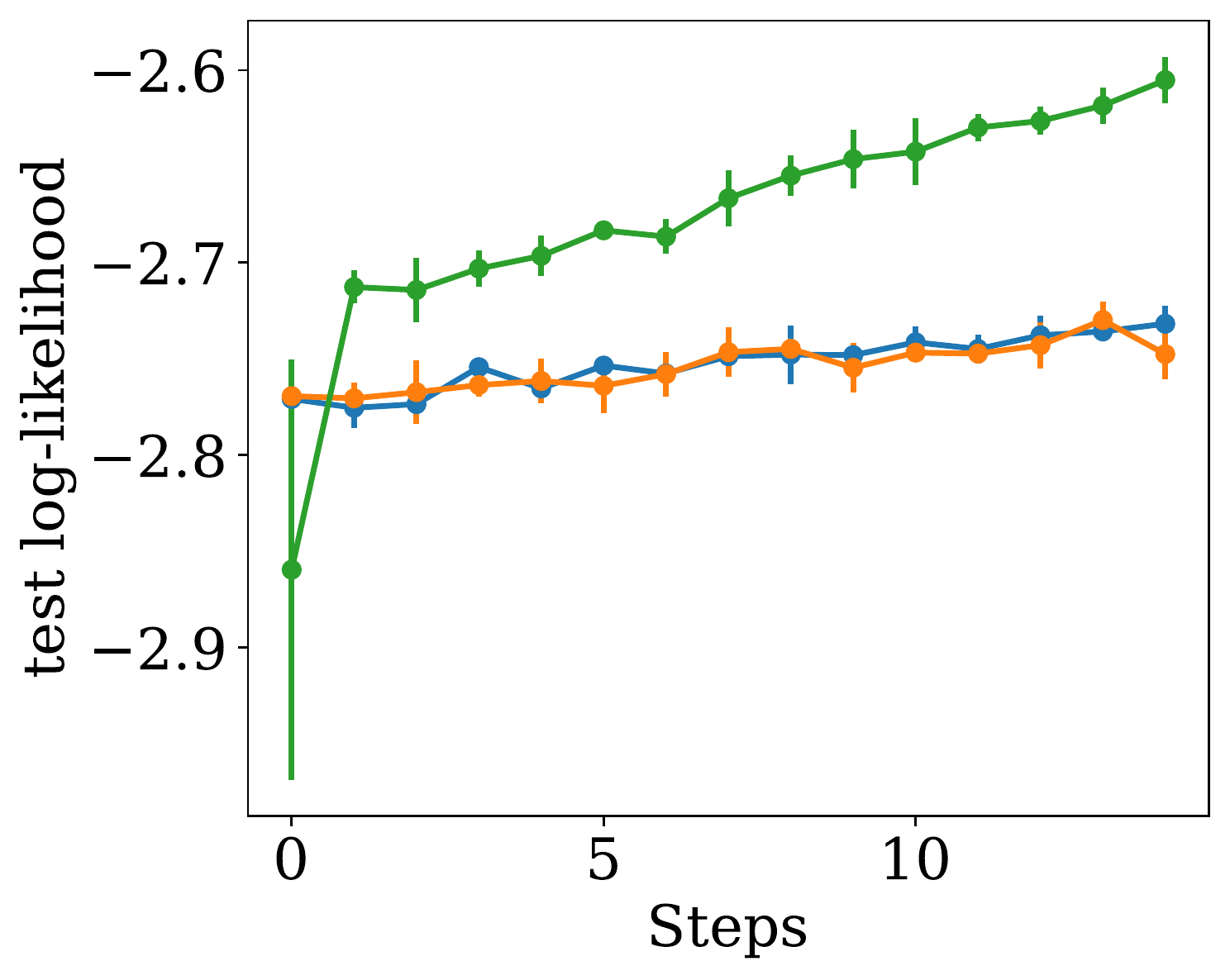}}
	\subfigure[Concrete]{\includegraphics[height=0.19\linewidth]{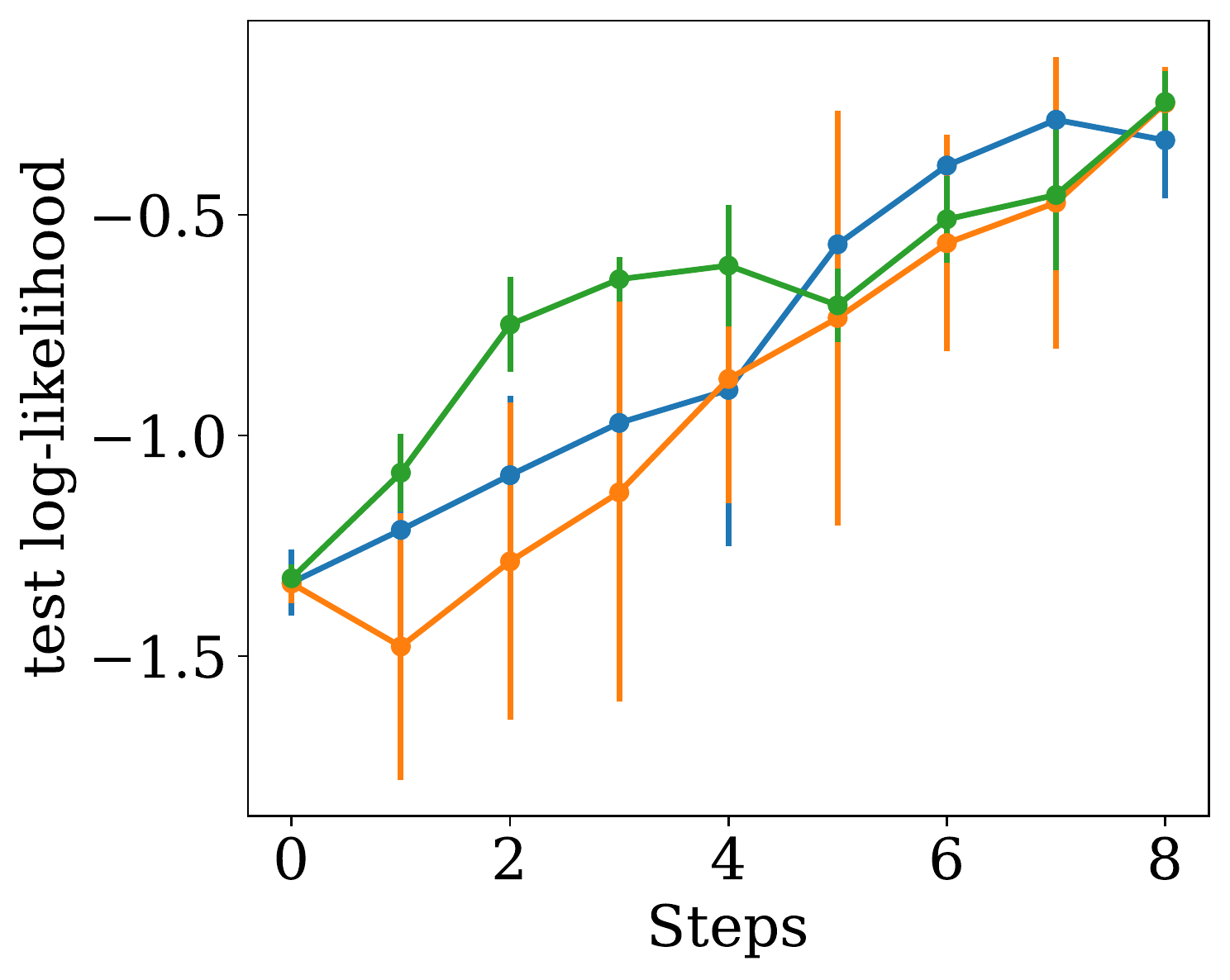}}
	\subfigure[Diabetes]{\includegraphics[height=0.19\linewidth]{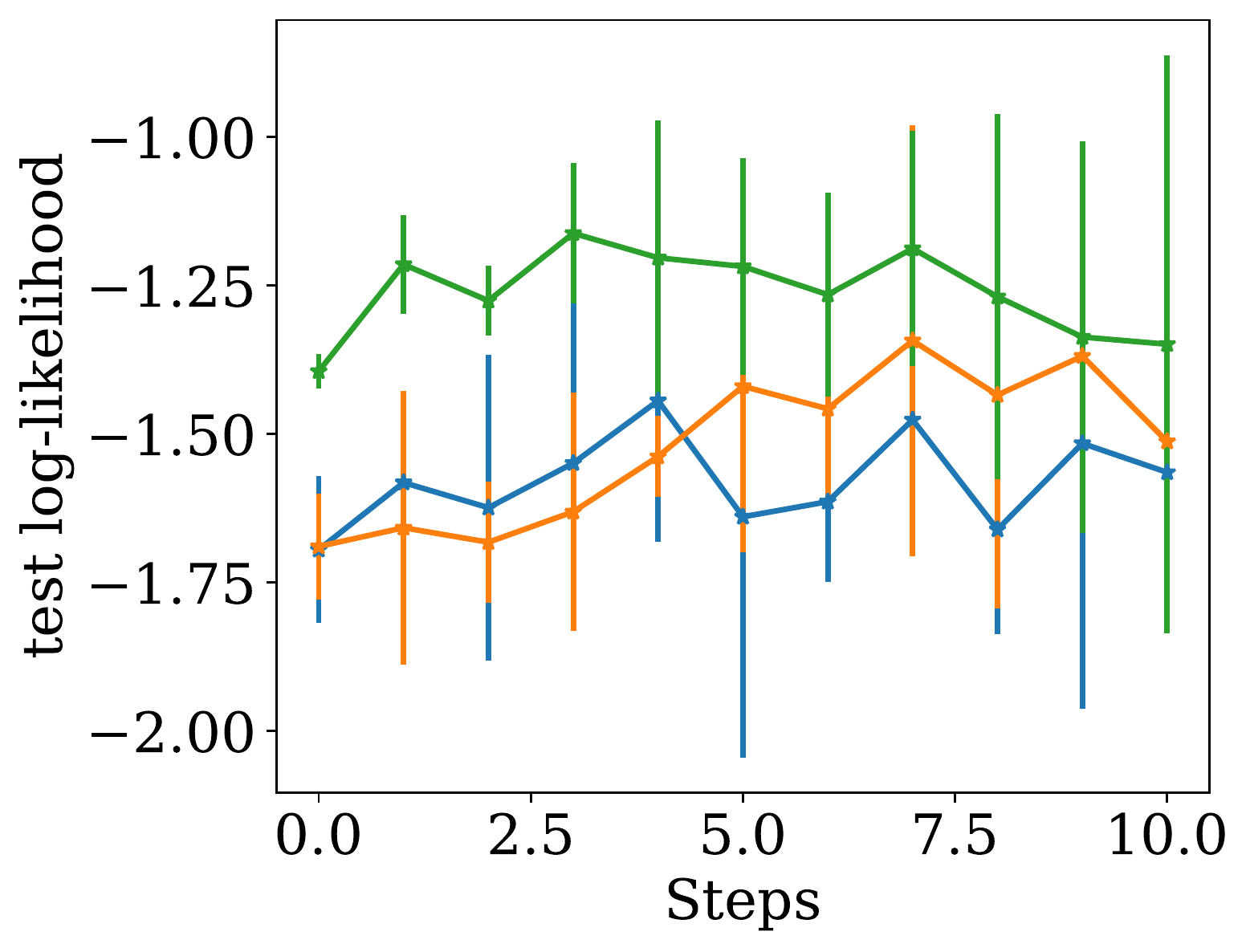}}
	\caption{SAIA log-likelihood curves. Horizontal axis shows 
	acquisition steps (number of discovered features). Vertical axis is the log-likelihood of the target $\log p(\y | \xo)$.} \label{fig:al_curves_ll}
\end{figure}

In order to extend the results provided in Section \ref{sec:saia}, we include here the log-likelihoods curves when dynamically selecting features using the same procedure (Figure \ref{fig:al_curves_ll}).

\subsection{Training times}\label{sec:app_times}

Table \ref{tab:training_times} shows the average training time in minutes for each model in the experiments for Tables \ref{tab:ll_xu} and \ref{tab:ll_y}. 
The ratio between training times for our method and the Gaussian baselines is approximately between 5 and 10.

\begin{table}[h] 

    \setlength{\tabcolsep}{3pt}

    \centering
    \resizebox{\linewidth}{!}{
	\begin{tabular}{@{}lrrrrrrrrrr@{}}
        \toprule
        \multicolumn{1}{r}{} &
          Bank &
          Avocado &
          Yatch &
          Diabetes &
          Concrete &
          Wine &
          Energy &
          Boston \\ \midrule
        VAEM &
          $29.92 \pm 0.39$ &
          $21.49 \pm 1.64$ &
          $5.89\pm 0.01$ &
          $8.79\pm 0.30$ &
          $7.70\pm 0.58$ &
          $7.92\pm 0.11$ &
          $8.18\pm 0.09$ &
          $10.01\pm 0.34$ \\
        MIWAEM & $63.33 \pm 6.20$
           & $37.17 \pm 2.28$
           & $8.21 \pm 0.24$
           & $13.29 \pm 0.33$
           & $11.51 \pm 0.52$
           & $11.81 \pm 0.13$
           & $16.71 \pm 0.16$
           & $15.01 \pm 0.13$
           \\
        H-VAEM &
          $41.44 \pm 0.38$ &
          $33.83 \pm 0.81$ &
          $15.84 \pm 0.11$ &
          $13.38\pm 0.47$ &
          $11.80\pm 0.38$ &
          $10.61\pm 0.54$ &
          $12.71\pm 1.20$ &
          $13.74 \pm 1.43$ \\
        HMC-VAEM &
          $281.88 \pm 9.14$ &
          $356.22 \pm 5.94$ &
          $23.50 \pm 1.60$ &
          $50.42\pm 1.18$ &
          $42.70\pm 3.14$ &
          $63.46\pm 1.97$ &
          $90.87\pm 7.05$ &
          $103.72 \pm 7.24$ \\
        \textbf{HH-VAEM} &
          $316.81 \pm 9.49$ &
          $388.28 \pm 6.47$ &
          $27.08 \pm 0.12$ &
          $68.29\pm 4.06$ &
          $65.78\pm 0.43$ &
          $79.97\pm 5.53$ &
          $140.33\pm 7.36$ &
          $129.30 \pm 4.69$ \\ \bottomrule
    \end{tabular}
    }
    \vspace{0.2cm}
    \caption{Training times (in minutes) of our model and baselines.}
    \label{tab:training_times}

\end{table}

\subsection{SAIA times}

\begin{figure*}[t]
	\centering
	\includegraphics[width=\linewidth]{figs/exp2/legend} \\
	\subfigure[Avocado]{\includegraphics[height=0.2\linewidth]{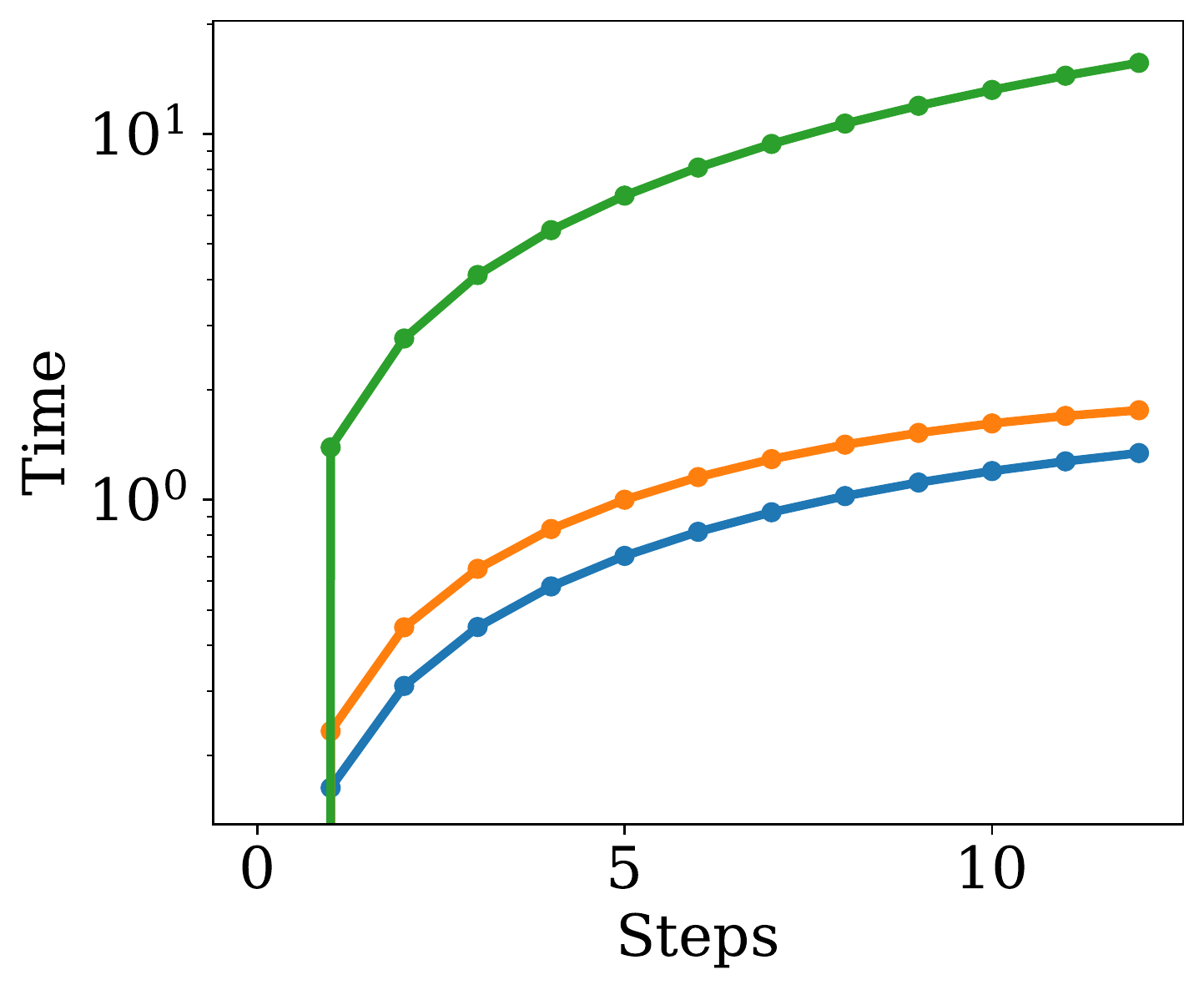}}
	\subfigure[Yatch]{\includegraphics[height=0.2\linewidth]{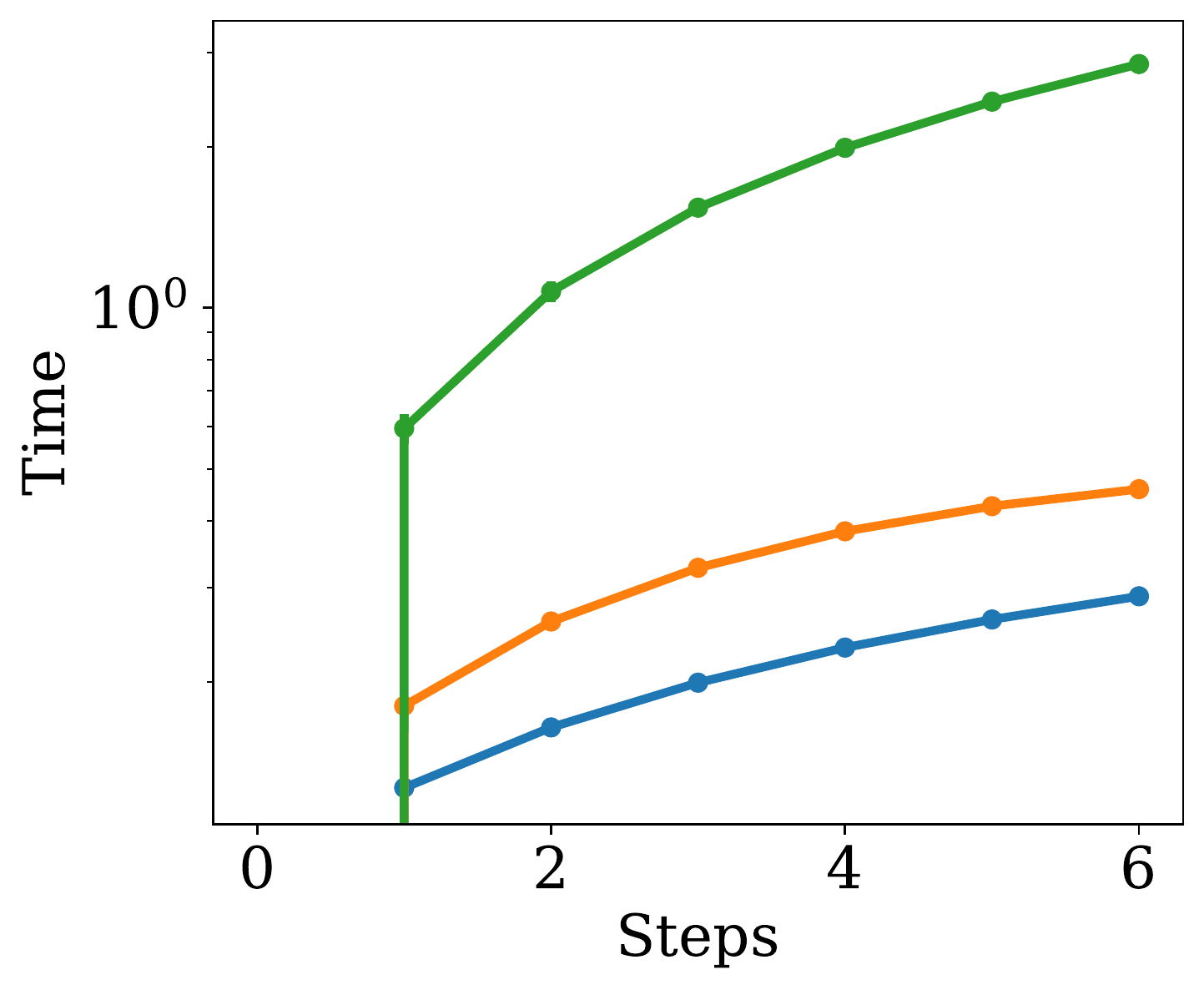}}
	\subfigure[Boston]{\includegraphics[height=0.2\linewidth]{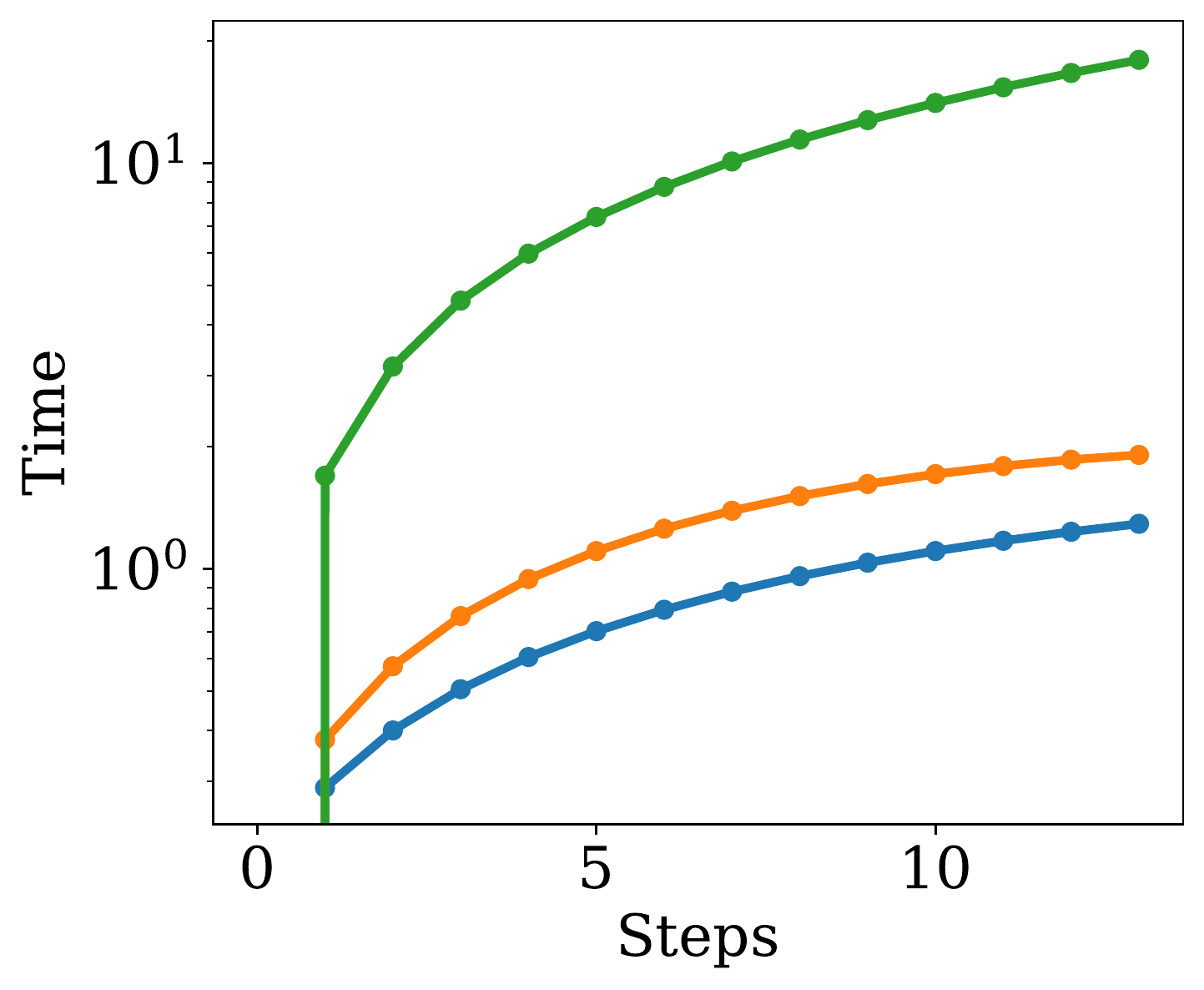}}
	\subfigure[Energy]{\includegraphics[height=0.2\linewidth]{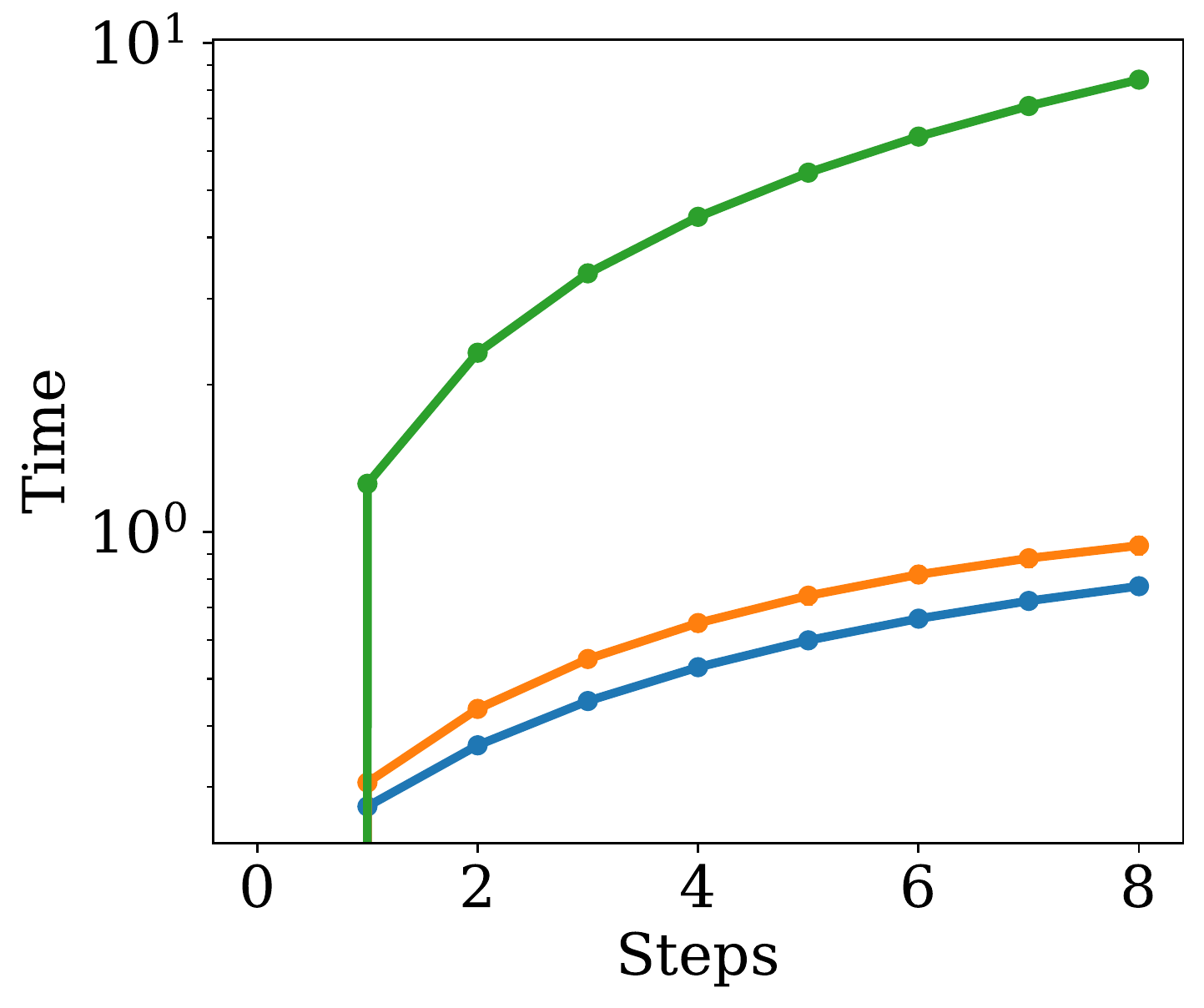}} 
	\subfigure[Wine]{\includegraphics[height=0.2\linewidth]{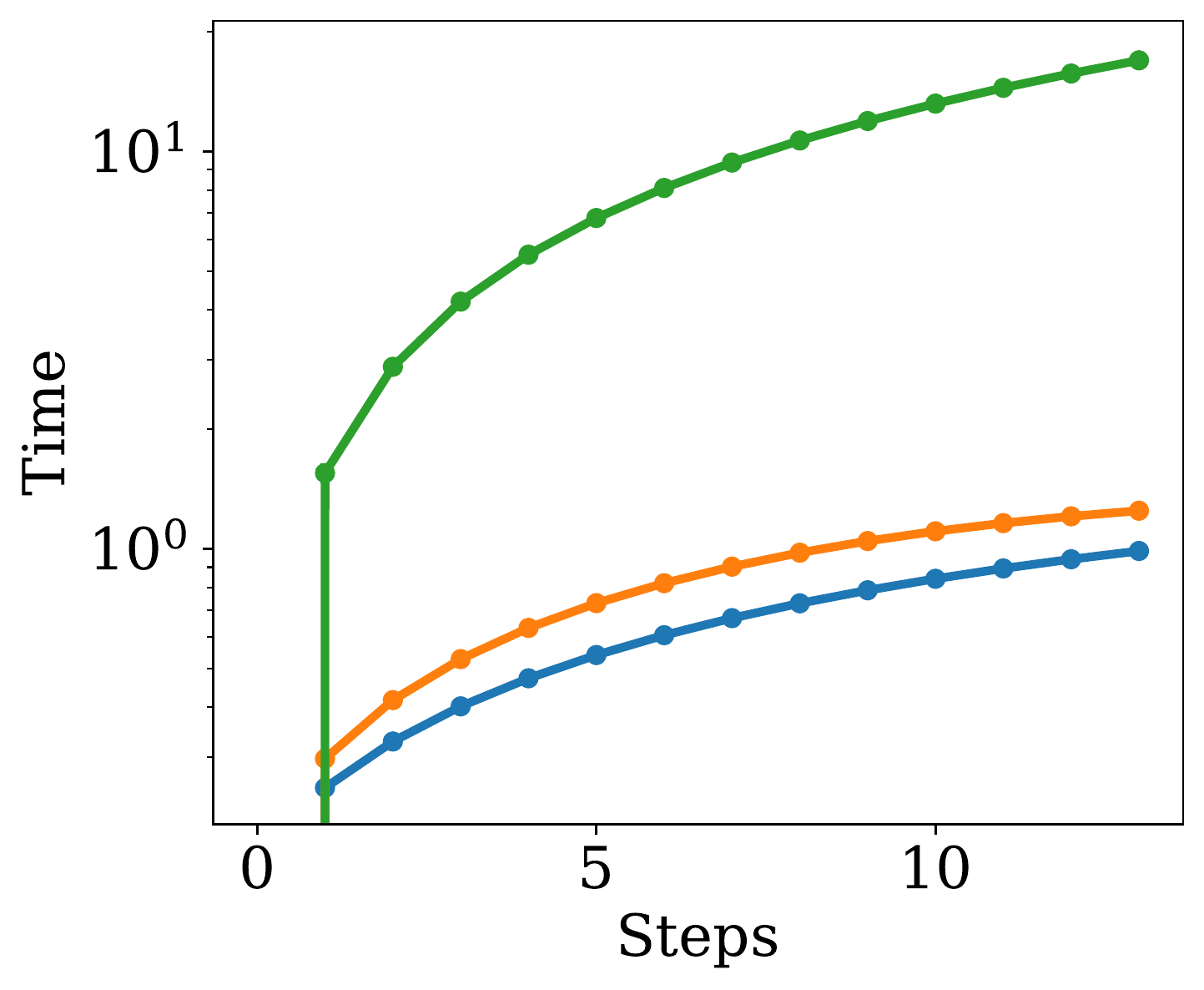}}
	\subfigure[Naval]{\includegraphics[height=0.2\linewidth]{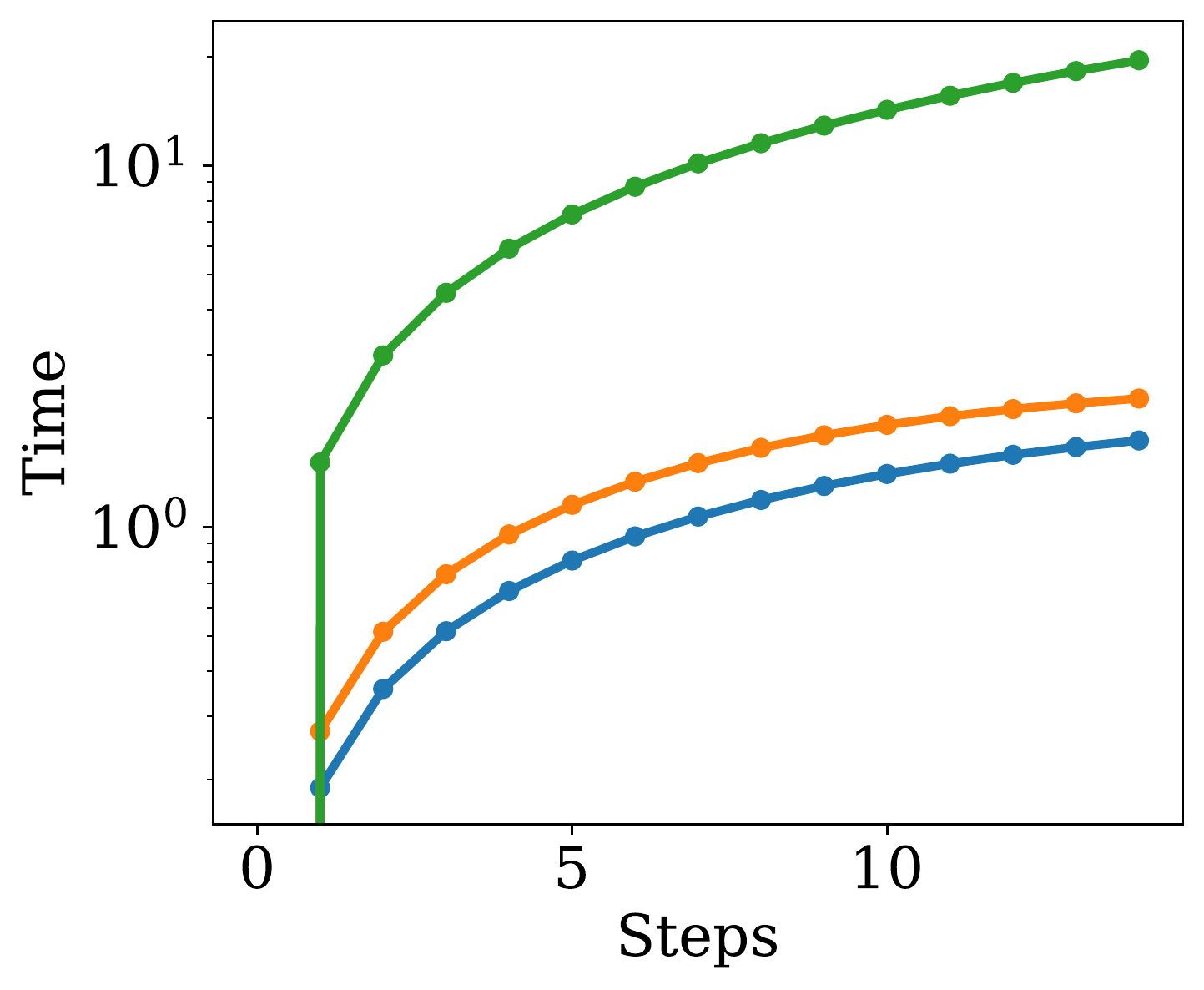}}
	\subfigure[Concrete]{\includegraphics[height=0.2\linewidth]{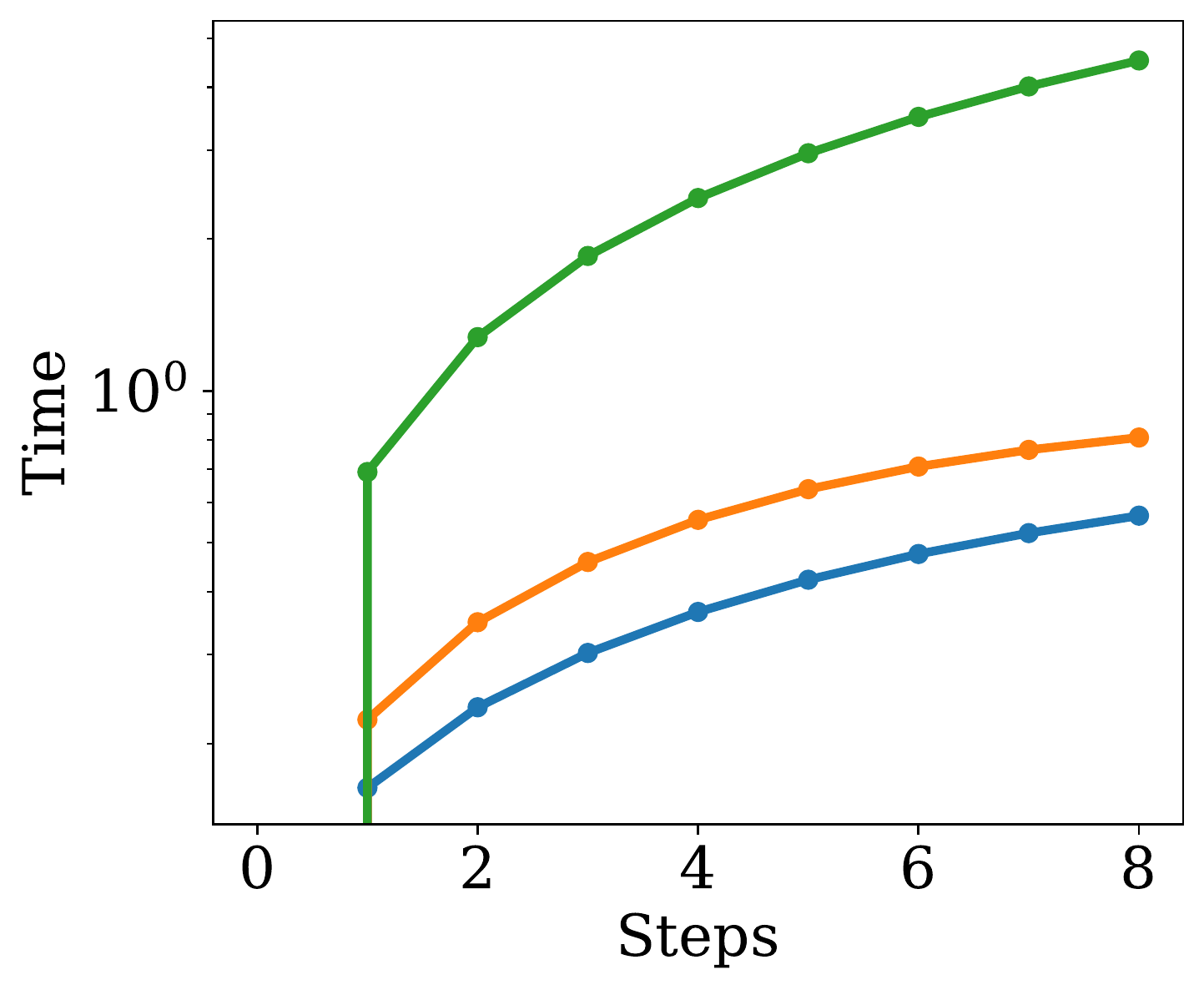}}
	\subfigure[Diabetes]{\includegraphics[height=0.2\linewidth]{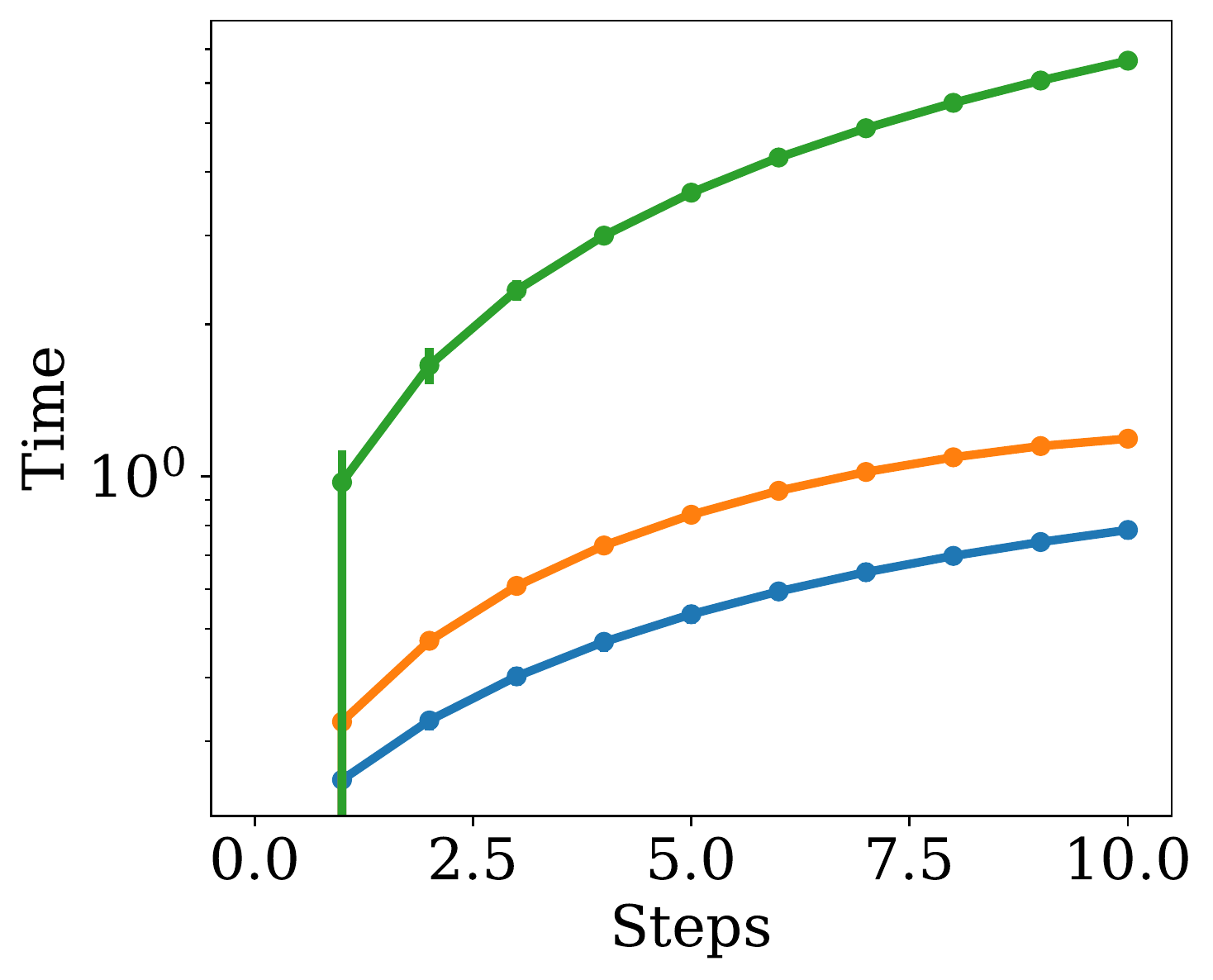}}
	\caption{SAIA time curves. Horizontal axis shows 
	acquisition steps (number of discovered features). Vertical axis is the elapsed time.} \label{fig:al_times}
\end{figure*}

The times for obtaining the SAIA metric curves in Section \ref{sec:saia} are included in Figure \ref{fig:al_times}. Although the performance is improved with HH-VAEM, it requires considerably longer time than the baselines to evaluate the reward, due to the HMC algorithm for sampling from the better approximated posterior. Future work might be oriented in proposing ways to measure and reduce this gap.

%% file: main.bbl
\begin{thebibliography}{10}

\bibitem{barrejon2021medical}
D.~Barrej{\'o}n, P.~M. Olmos, and A.~Art{\'e}s-Rodr{\'\i}guez.
\newblock Medical data wrangling with sequential {V}ariational {A}utoencoders.
\newblock {\em arXiv preprint arXiv:2103.07206}, 2021.

\bibitem{bengio2009learning}
Y.~Bengio.
\newblock {\em Learning deep architectures for {AI}}.
\newblock Now Publishers Inc, 2009.

\bibitem{bernardo1979expected}
J.~M. Bernardo.
\newblock Expected information as expected utility.
\newblock {\em the Annals of Statistics}, pages 686--690, 1979.

\bibitem{betancourt2017conceptual}
M.~Betancourt.
\newblock A conceptual introduction to {H}amiltonian {M}onte {C}arlo.
\newblock {\em arXiv preprint arXiv:1701.02434}, 2017.

\bibitem{betancourt2015hamiltonian}
M.~Betancourt and M.~Girolami.
\newblock Hamiltonian {M}onte {C}arlo for hierarchical models.
\newblock {\em Current trends in Bayesian methodology with applications},
  79(30):2--4, 2015.

\bibitem{campbell2021gradient}
A.~Campbell, W.~Chen, V.~Stimper, J.~M. Hernandez-Lobato, and Y.~Zhang.
\newblock A gradient {B}ased {S}trategy for {H}amiltonian {M}onte {C}arlo
  {H}yperparameter optimization.
\newblock In {\em International Conference on Machine Learning}, pages
  1238--1248. PMLR, 2021.

\bibitem{caterini2018hamiltonian}
A.~L. Caterini, A.~Doucet, and D.~Sejdinovic.
\newblock Hamiltonian {V}ariational {A}uto-{E}ncoder.
\newblock {\em arXiv preprint arXiv:1805.11328}, 2018.

\bibitem{child2020very}
R.~Child.
\newblock Very deep {VAE}s generalize autoregressive models and can outperform
  them on images.
\newblock {\em arXiv preprint arXiv:2011.10650}, 2020.

\bibitem{collier2020vaes}
M.~Collier, A.~Nazabal, and C.~K. Williams.
\newblock {VAE}s in the presence of missing data.
\newblock {\em arXiv preprint arXiv:2006.05301}, 2020.

\bibitem{cremer2018inference}
C.~Cremer, X.~Li, and D.~Duvenaud.
\newblock Inference {S}uboptimality in {V}ariational {A}utoencoders.
\newblock In {\em International Conference on Machine Learning}, pages
  1078--1086. PMLR, 2018.

\bibitem{dua2017uci}
D.~Dua, C.~Graff, et~al.
\newblock {UCI} machine learning repository.
\newblock 2017.

\bibitem{duane1987hybrid}
S.~Duane, A.~D. Kennedy, B.~J. Pendleton, and D.~Roweth.
\newblock Hybrid {M}onte {C}arlo.
\newblock {\em Physics letters B}, 195(2):216--222, 1987.

\bibitem{eduardo2020robust}
S.~Eduardo, A.~Naz{\'a}bal, C.~K. Williams, and C.~Sutton.
\newblock Robust {V}ariational {A}utoencoders for outlier detection and repair
  of mixed-type data.
\newblock In {\em International Conference on Artificial Intelligence and
  Statistics}, pages 4056--4066. PMLR, 2020.

\bibitem{garnelo2018conditional}
M.~Garnelo, D.~Rosenbaum, C.~Maddison, T.~Ramalho, D.~Saxton, M.~Shanahan,
  Y.~W. Teh, D.~Rezende, and S.~A. Eslami.
\newblock Conditional {N}eural {P}rocesses.
\newblock In {\em International Conference on Machine Learning}, pages
  1704--1713. PMLR, 2018.

\bibitem{ghahramani1995learning}
Z.~Ghahramani and M.~I. Jordan.
\newblock Learning from incomplete data.
\newblock 1995.

\bibitem{girolami2011riemann}
M.~Girolami and B.~Calderhead.
\newblock Riemann manifold langevin and hamiltonian monte carlo methods.
\newblock {\em Journal of the Royal Statistical Society: Series B (Statistical
  Methodology)}, 73(2):123--214, 2011.

\bibitem{gong2020sliced}
W.~Gong, Y.~Li, and J.~M. Hern{\'a}ndez-Lobato.
\newblock Sliced {K}ernelized {S}tein {D}iscrepancy.
\newblock {\em arXiv preprint arXiv:2006.16531}, 2020.

\bibitem{gong2021variational}
Y.~Gong, H.~Hajimirsadeghi, J.~He, T.~Durand, and G.~Mori.
\newblock Variational {S}elective {A}utoencoder: Learning from
  partially-observed heterogeneous data.
\newblock In {\em International Conference on Artificial Intelligence and
  Statistics}, pages 2377--2385. PMLR, 2021.

\bibitem{hoffman2017learning}
M.~D. Hoffman.
\newblock Learning deep latent {G}aussian models with {M}arkov {C}hain {M}onte
  {C}arlo.
\newblock In {\em International conference on machine learning}, pages
  1510--1519. PMLR, 2017.

\bibitem{huang2018active}
S.-J. Huang, M.~Xu, M.-K. Xie, M.~Sugiyama, G.~Niu, and S.~Chen.
\newblock Active feature acquisition with supervised matrix completion.
\newblock In {\em Proceedings of the 24th ACM SIGKDD International Conference
  on Knowledge Discovery \& Data Mining}, pages 1571--1579, 2018.

\bibitem{ipsen2020deal}
N.~Ipsen, P.-A. Mattei, and J.~Frellsen.
\newblock How to deal with missing data in supervised deep learning?
\newblock In {\em Artemiss-ICML Workshop on the Art of Learning with Missing
  Values}, 2020.

\bibitem{joy2021capturing}
T.~Joy, S.~Schmon, P.~Torr, S.~Narayanaswamy, and T.~Rainforth.
\newblock Capturing label characteristics in vaes.
\newblock In {\em Proceedings of the ICLR Conference 2021}. OpenReview, 2021.

\bibitem{kingma2016improved}
D.~P. Kingma, T.~Salimans, R.~Jozefowicz, X.~Chen, I.~Sutskever, and
  M.~Welling.
\newblock Improved {V}ariational {I}nference with {I}nverse {A}utoregressive
  {F}low.
\newblock {\em Advances in neural information processing systems},
  29:4743--4751, 2016.

\bibitem{kingma2013auto}
D.~P. Kingma and M.~Welling.
\newblock Auto-encoding variational bayes.
\newblock {\em arXiv preprint arXiv:1312.6114}, 2013.

\bibitem{kingma2019introduction}
D.~P. Kingma and M.~Welling.
\newblock An introduction to {V}ariational {A}utoencoders.
\newblock {\em arXiv preprint arXiv:1906.02691}, 2019.

\bibitem{kraskov2004estimating}
A.~Kraskov, H.~St{\"o}gbauer, and P.~Grassberger.
\newblock Estimating mutual information.
\newblock {\em Physical review E}, 69(6):066138, 2004.

\bibitem{lecun1998mnist}
Y.~LeCun.
\newblock The {MNIST} database of handwritten digits.
\newblock {\em http://yann. lecun. com/exdb/mnist/}, 1998.

\bibitem{li2019generative}
Y.~Li, J.~Bradshaw, and Y.~Sharma.
\newblock Are generative classifiers more robust to adversarial attacks?
\newblock In {\em International Conference on Machine Learning}, pages
  3804--3814. PMLR, 2019.

\bibitem{lindley1956measure}
D.~V. Lindley.
\newblock On a measure of the information provided by an experiment.
\newblock {\em The Annals of Mathematical Statistics}, pages 986--1005, 1956.

\bibitem{little2019statistical}
R.~J. Little and D.~B. Rubin.
\newblock {\em Statistical analysis with missing data}, volume 793.
\newblock John Wiley \& Sons, 2019.

\bibitem{liu2016kernelized}
Q.~Liu, J.~Lee, and M.~Jordan.
\newblock A {K}ernelized {S}tein {D}iscrepancy for goodness-of-fit tests.
\newblock In {\em International conference on machine learning}, pages
  276--284. PMLR, 2016.

\bibitem{liu2015faceattributes}
Z.~Liu, P.~Luo, X.~Wang, and X.~Tang.
\newblock Deep learning face attributes in the wild.
\newblock In {\em Proceedings of International Conference on Computer Vision
  (ICCV)}, December 2015.

\bibitem{ma2020vaem}
C.~Ma, S.~Tschiatschek, J.~M. Hern{\'a}ndez-Lobato, R.~Turner, and C.~Zhang.
\newblock {VAEM}: a {D}eep {G}enerative {M}odel for {H}eterogeneous {M}ixed
  {T}ype {D}ata.
\newblock {\em arXiv preprint arXiv:2006.11941}, 2020.

\bibitem{ma2018eddi}
C.~Ma, S.~Tschiatschek, K.~Palla, J.~M. Hern{\'a}ndez-Lobato, S.~Nowozin, and
  C.~Zhang.
\newblock Ed{DI}: {E}fficient {D}ynamic {D}iscovery of {H}igh-{V}alue
  {I}nformation with {P}artial {VAE}.
\newblock {\em arXiv preprint arXiv:1809.11142}, 2018.

\bibitem{maaloe2019biva}
L.~Maal{\o}e, M.~Fraccaro, V.~Li{\'e}vin, and O.~Winther.
\newblock {BIVA}: A very deep hierarchy of latent variables for generative
  modeling.
\newblock {\em arXiv preprint arXiv:1902.02102}, 2019.

\bibitem{mattei2019miwae}
P.-A. Mattei and J.~Frellsen.
\newblock {MIWAE}: Deep generative modelling and imputation of incomplete data
  sets.
\newblock In {\em International Conference on Machine Learning}, pages
  4413--4423. PMLR, 2019.

\bibitem{melville2004active}
P.~Melville, M.~Saar-Tsechansky, F.~Provost, and R.~Mooney.
\newblock Active feature-value acquisition for classifier induction.
\newblock In {\em Fourth IEEE International Conference on Data Mining
  (ICDM'04)}, pages 483--486. IEEE, 2004.

\bibitem{nazabal2020handling}
A.~Nazabal, P.~M. Olmos, Z.~Ghahramani, and I.~Valera.
\newblock Handling {I}ncomplete {H}eterogeneous {D}ata using {VAE}s.
\newblock {\em Pattern Recognition}, 107:107501, 2020.

\bibitem{neal2011mcmc}
R.~M. Neal et~al.
\newblock {MCMC} using {H}amiltonian dynamics.
\newblock {\em Handbook of markov chain monte carlo}, 2(11):2, 2011.

\bibitem{razavi2019preventing}
A.~Razavi, A.~v.~d. Oord, B.~Poole, and O.~Vinyals.
\newblock Preventing posterior collapse with delta-{VAE}s.
\newblock {\em arXiv preprint arXiv:1901.03416}, 2019.

\bibitem{rezende2014stochastic}
D.~J. Rezende, S.~Mohamed, and D.~Wierstra.
\newblock Stochastic backpropagation and approximate inference in deep
  generative models.
\newblock In {\em International conference on machine learning}, pages
  1278--1286. PMLR, 2014.

\bibitem{ross2014mutual}
B.~C. Ross.
\newblock Mutual information between discrete and continuous data sets.
\newblock {\em PloS one}, 9(2):e87357, 2014.

\bibitem{saar2009active}
M.~Saar-Tsechansky, P.~Melville, and F.~Provost.
\newblock Active feature-value acquisition.
\newblock {\em Management Science}, 55(4):664--684, 2009.

\bibitem{salakhutdinov2015learning}
R.~Salakhutdinov.
\newblock Learning deep generative models.
\newblock {\em Annual Review of Statistics and Its Application}, 2:361--385,
  2015.

\bibitem{salakhutdinov2009deep}
R.~Salakhutdinov and G.~Hinton.
\newblock Deep {B}oltzmann machines.
\newblock In {\em Artificial intelligence and statistics}, pages 448--455.
  PMLR, 2009.

\bibitem{salimans2015markov}
T.~Salimans, D.~Kingma, and M.~Welling.
\newblock Markov {C}hain {M}onte {C}arlo and {V}ariational {I}nference:
  Bridging the gap.
\newblock In {\em International Conference on Machine Learning}, pages
  1218--1226. PMLR, 2015.

\bibitem{smieja2018processing}
M.~{\'S}mieja, {\L}.~Struski, J.~Tabor, B.~Zieli{\'n}ski, and P.~Spurek.
\newblock Processing of missing data by neural networks.
\newblock {\em Advances in neural information processing systems}, 31, 2018.

\bibitem{sonderby2016ladder}
C.~K. S{\o}nderby, T.~Raiko, L.~Maal{\o}e, S.~K. S{\o}nderby, and O.~Winther.
\newblock Ladder {V}ariational {A}utoencoders.
\newblock {\em Advances in neural information processing systems},
  29:3738--3746, 2016.

\bibitem{stekhoven2012missforest}
D.~J. Stekhoven and P.~B{\"u}hlmann.
\newblock Missforest—non-parametric missing value imputation for mixed-type
  data.
\newblock {\em Bioinformatics}, 28(1):112--118, 2012.

\bibitem{thahir2012efficient}
M.~Thahir, T.~Sharma, and M.~K. Ganapathiraju.
\newblock An efficient heuristic method for active feature acquisition and its
  application to protein-protein interaction prediction.
\newblock In {\em BMC proceedings}, volume~6, pages 1--9. BioMed Central, 2012.

\bibitem{thin2021monte}
A.~Thin, N.~Kotelevskii, A.~Doucet, A.~Durmus, E.~Moulines, and M.~Panov.
\newblock Monte carlo variational auto-encoders.
\newblock In {\em International Conference on Machine Learning}, pages
  10247--10257. PMLR, 2021.

\bibitem{tresp1993training}
V.~Tresp, S.~Ahmad, and R.~Neuneier.
\newblock Training neural networks with deficient data.
\newblock {\em Advances in neural information processing systems}, 6, 1993.

\bibitem{vahdat2020nvae}
A.~Vahdat and J.~Kautz.
\newblock {NVAE}: A {D}eep {H}ierarchical {V}ariational {A}utoencoder.
\newblock {\em arXiv preprint arXiv:2007.03898}, 2020.

\bibitem{wang2020posterior}
Y.~Wang and J.~P. Cunningham.
\newblock Posterior collapse and latent variable non-identifiability.
\newblock In {\em Third Symposium on Advances in Approximate Bayesian
  Inference}, 2020.

\bibitem{xiao2017fashion}
H.~Xiao, K.~Rasul, and R.~Vollgraf.
\newblock Fashion-{MNIST}: a novel image dataset for benchmarking machine
  learning algorithms.
\newblock {\em arXiv preprint arXiv:1708.07747}, 2017.

\bibitem{zhang2018advances}
C.~Zhang, J.~B{\"u}tepage, H.~Kjellstr{\"o}m, and S.~Mandt.
\newblock Advances in {V}ariational {I}nference.
\newblock {\em IEEE transactions on pattern analysis and machine intelligence},
  41(8):2008--2026, 2018.

\end{thebibliography}
